%% file: main.tex
\documentclass[11pt]{article}

\usepackage{acl}

\usepackage{times}
\usepackage{latexsym}

\usepackage[T1]{fontenc}

\usepackage[utf8]{inputenc}

\usepackage{microtype}

\usepackage{inconsolata}

\usepackage{graphicx}
\usepackage[export]{adjustbox} 
\input{math_commands.tex}

\usepackage{multirow}
\usepackage{pifont}
\usepackage{booktabs}
\usepackage{xspace} 
\usepackage[table]{xcolor}
\usepackage[most]{tcolorbox} 
\tcbuselibrary{skins} 
\usepackage{threeparttable}
\usepackage{enumitem}
\usepackage{hyperref}
\usepackage{etoc}

\newtcolorbox{takeawaybox}[1]{
  colback=white!98!black,
  colframe=white!86!black,
  title={\textcolor{black}{#1}},
  boxrule=0.8pt,
  arc=4pt,
  left=6pt,
  right=6pt,
  top=3pt,
  bottom=3pt,
}

\definecolor{myGreen}{RGB}{34, 139, 34}
\definecolor{myRed}{HTML}{CC3333}
\definecolor{myYellow}{HTML}{DAA520} 
\definecolor{myBlue}{HTML}{4682B4}

\newcommand{\model}{\texttt{Mem-Gallery}\xspace}
\newcommand{\cmark}{\textcolor{myGreen}{\ding{51}}}
\newcommand{\xmark}{\textcolor{myRed}{\ding{55}}}
\newcommand{\pmark}{\textcolor{myBlue}{\ding{51}\rotatebox[origin=c]{-6.2}{\kern-0.7em\ding{55}}}}

\newcommand{\qwenlogo}{\includegraphics[height=1.2em]{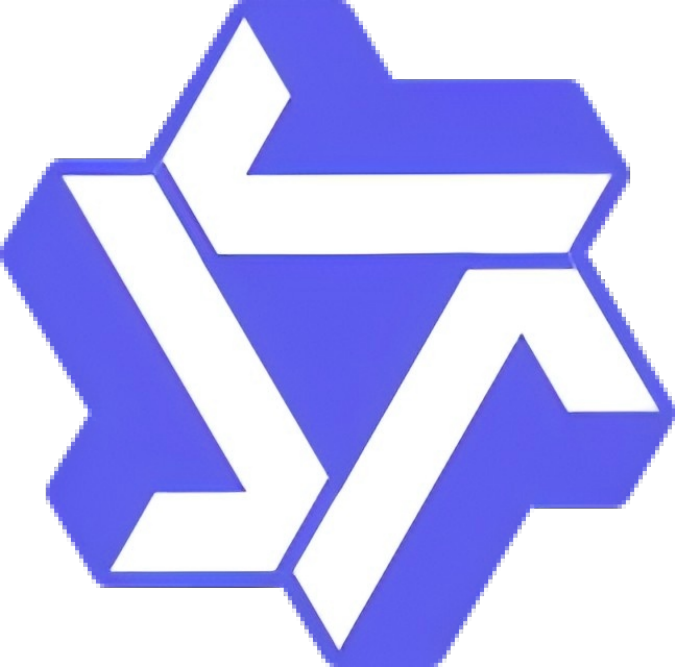}}
\newcommand{\geminilogo}{\includegraphics[height=1.2em]{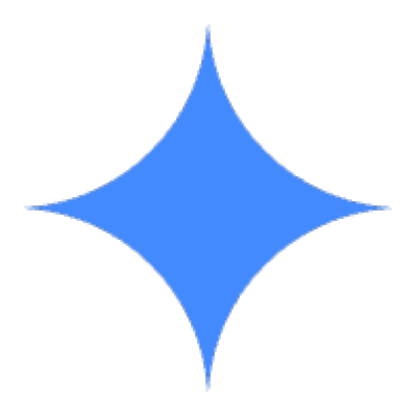}}
\newcommand{\openailogo}
{\includegraphics[height=1.2em]{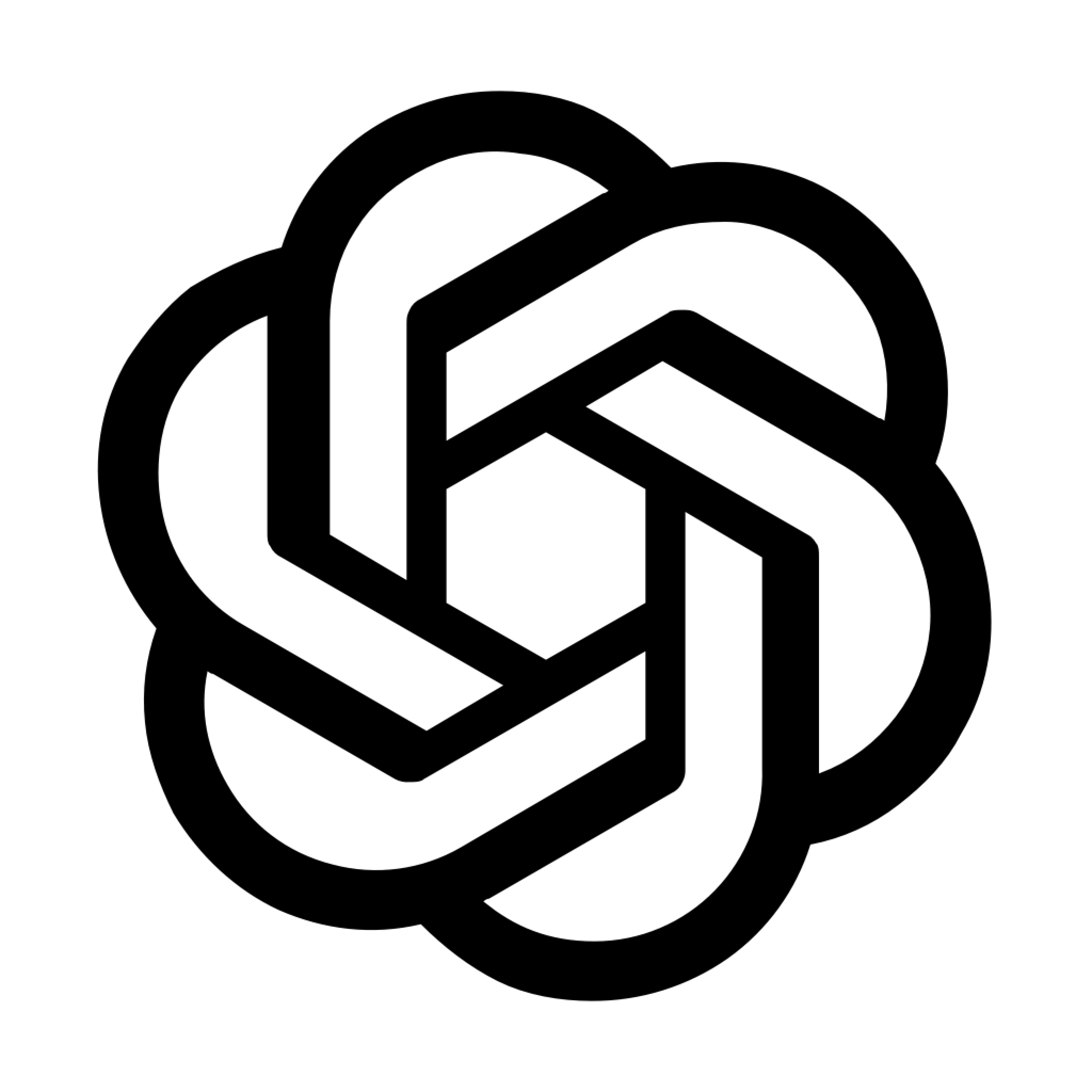}}
\newcommand{\ourlogo}{\raisebox{-0.42em}{\includegraphics[height=1.6em]{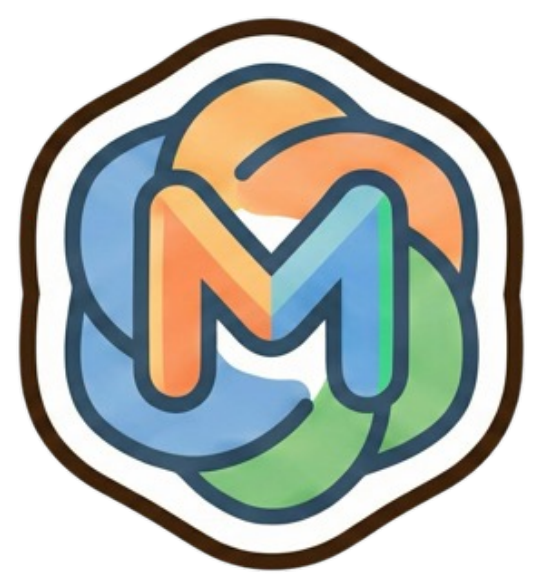}}}

\newcommand{\gitlogo}{\raisebox{-0.24em}{\includegraphics[height=1.2em]{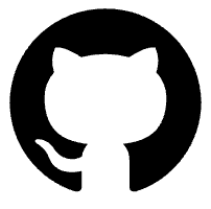}}}

\definecolor{TopOne}{RGB}{254,224,210}
\definecolor{TopTwo}{RGB}{222,235,247}
\definecolor{TopThree}{RGB}{254,237,222}

\definecolor{qwenbar}{RGB}{228,220,246}
\definecolor{geminibar}{RGB}{220,240,238}
\definecolor{gptbar}{RGB}{230,232,236}

\newcommand{\name}{\texttt{Mem-Gallery}\xspace}

\title{\ourlogo\ \model: Benchmarking Multimodal Long-Term Conversational Memory for MLLM Agents}

\author{Yuanchen Bei$^1$,\ Tianxin Wei$^1$,\ Xuying Ning$^1$,\ Yanjun Zhao$^1$,\ Zhining Liu$^1$,\\ {\bf Xiao Lin}$^1${\bf,\ Yada Zhu}$^2${\bf,\ Hendrik Hamann}$^{3,4}${\bf,\ Jingrui He}$^1${\bf,\ Hanghang Tong}$^1$ \\
$^1$University of Illinois Urbana-Champaign\quad $^2$MIT-IBM Watson AI Lab, IBM Research\\
$^3$Stony Brook University\quad $^4$Brookhaven National Laboratory}

\begin{document}
\maketitle
\input{0_abstract}

\input{1_intro}

\input{2_related_work}

\input{3_method}

\input{4_experiment}

\input{5_conclusion}

\clearpage
\section*{Limitations}
While \name provides a comprehensive evaluation of multimodal long-term conversational memory, the following limitations remain. First, the benchmark focuses on vision–language conversational settings and does not explicitly cover other modalities such as audio or embodied signals, which may be relevant in broader agentic scenarios. Second, the evaluation primarily focuses on memory-centric capabilities in long-horizon conversations and does not aim to exhaustively assess other agent behaviors, such as planning or tool use. We leave these extensions to future work.

\bibliography{main}

\clearpage
\appendix

\input{X_suppl}

\end{document}

%% file: math_commands.tex
\usepackage{amsmath,amsfonts,bm}

\def\eqref#1{equation~\ref{#1}}

\def\1{\bm{1}}

\DeclareMathAlphabet{\mathsfit}{\encodingdefault}{\sfdefault}{m}{sl}
\SetMathAlphabet{\mathsfit}{bold}{\encodingdefault}{\sfdefault}{bx}{n}



%% file: 0_abstract.tex
\begin{abstract}
Long-term memory is a critical capability for multimodal large language model (MLLM) agents, particularly in conversational settings where information accumulates and evolves over time.
However, existing benchmarks either evaluate multi-session memory in text-only conversations or assess multimodal understanding within localized contexts, failing to evaluate how multimodal memory is preserved, organized, and evolved across long-term conversational trajectories. 
Thus, we introduce \model\footnote{\gitlogo\  \url{https://github.com/YuanchenBei/Mem-Gallery}}, a new benchmark for evaluating multimodal long-term conversational memory in MLLM agents.
\model features high-quality multi-session conversations grounded in both visual and textual information, with long interaction horizons and rich multimodal dependencies. 
Building on this dataset, we propose a systematic evaluation framework that assesses key memory capabilities along three functional dimensions: memory extraction and test-time adaptation, memory reasoning, and memory knowledge management. 
Extensive benchmarking across thirteen memory systems reveals several key findings, highlighting the necessity of explicit multimodal information retention and memory organization, the persistent limitations in memory reasoning and knowledge management, as well as the efficiency bottleneck of current models. 
\end{abstract}

%% file: 1_intro.tex
\section{Introduction}
\label{sec:intro}

The rapid progress of Multimodal Large Language Models (MLLMs) has enabled the development of agents that can perceive, reason, and interact with the world through both language and vision~\cite{wu2023multimodal,zhang2024mm}.
A fundamental capability for such agents is long-term memory: store, retrieve, and update information accumulated over extended interactions~\cite{zhang2025survey}. In particular, multi-session conversations constitute a primary medium through which agents acquire, refine, and utilize memory, making conversational settings a natural and critical testbed for evaluating long-term memory capabilities~\cite{wu2025longmemeval}.

\begin{figure}[tbp]
\vspace{-1em}
\centering
\includegraphics[width=1.01\linewidth]{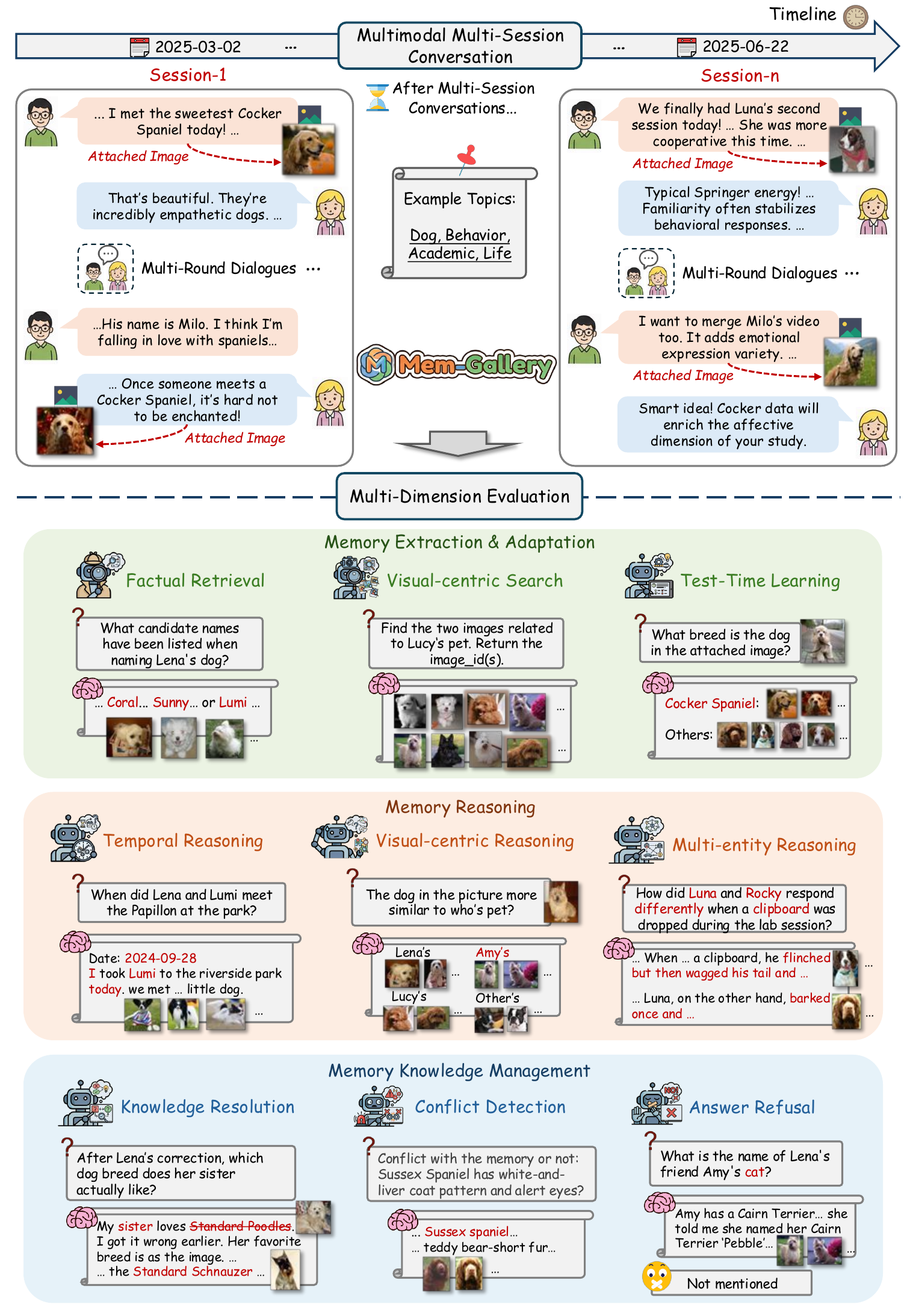}
\caption{The conceptual illustration of \name.}
\vspace{-1em}
\label{fig:toy_example}
\end{figure}

Human memory in conversation is inherently \textit{multimodal and evolving}~\cite{ardesch2019evolutionary,luppi2022synergistic}.
An effective agent therefore requires the ability to not only recall past information but also integrate visual and textual cues, reason across events, and update outdated knowledge as conversations progress~\cite{hu2025memory,bo2025agentic}. However, endowing MLLM agents with such long-term multimodal conversational memory largely remains an open challenge, and its systematic evaluation is still underexplored.

Despite growing interest in agentic memory, existing benchmarks reflect a fundamental mismatch with the real-world conversational memory for MLLM agents. 
Current conversation benchmarks tend to fall into two disjoint categories. On one hand, \textbf{text-only conversational memory benchmarks} evaluate memory over multi-session dialogues but discard visual modality~\cite{wu2025longmemeval,hu2025evaluating}. On the other hand, \textbf{localized multimodal context understanding benchmarks} 
introduce the visual modality but typically focus on short-horizon understanding within only one session, failing to assess cross-session information evolution and management~\cite{liu2024mmdu,xue2025mmrc}. This limitation makes them unsuitable for memory evaluation.  
Consequently, current conversational benchmarks still fall short in {\em evaluating how agents organize, maintain, and retrieve multimodal memory over extended conversational timelines, where visual and linguistic information interact dynamically}.

To address this gap, we propose \model, a benchmark designed to evaluate multimodal long-term conversational memory in MLLM agents systematically, as described in Figure~\ref{fig:toy_example}. \name firstly introduces a new dataset of multi-session conversations grounded in both images and text, reflecting daily and domain-specific knowledge. Each conversation spans a long interaction horizon, where information is incrementally introduced, referenced, and updated. 
Based on this dataset, we structure evaluation tasks into three key functional dimensions of memory. These three dimensions correspond to the core stages of long-term conversational memory in real-world agents: acquiring usable memory, reasoning over evolving multimodal evidence, and regulating memory under dynamic and potentially inconsistent states.
Specifically, \name evaluates:
(1) \textbf{Memory Extraction and Adaptation}, which includes multimodal factual retrieval, visual-centric search, and test-time learning over long multimodal histories. (2) \textbf{Memory Reasoning}, which evaluates how agents conduct reasoning over multimodal memory clues, including temporal reasoning, visual-centric reasoning, and multi-entity reasoning. (3) \textbf{Memory Knowledge Management}, which examines the ability to resolve knowledge contradictions, detect conflicts, and appropriately refuse to answer when information is missing, outdated, or inconsistent.

Benchmarking across thirteen memory mechanisms reveals several findings.
(1) \textbf{Multimodal Information Effectiveness}: explicitly preserving visual information in memory is beneficial.
(2) \textbf{Memory Organization Importance}:  
highlighting the necessity of principled multimodal memory organization and maintenance. 
(3) \textbf{Memory Reasoning and Knowledge Management Limitations}: 
existing multimodal memory models struggle in reasoning-intensive settings, as well as in handling knowledge updates and conflicts.
(4) \textbf{Efficiency Bottleneck}: 
multimodal memory overall introduces larger storage and retrieval overhead that may hinder practical deployment.

Our contributions are summarized as follows: 
\vspace{-0.25em}
\begin{itemize}[leftmargin=*, itemsep=0pt]
    \item \textbf{New Scenario \& Dataset}: We formulate multimodal long-term conversational memory as an evolving system that spans multiple sessions, modalities, and memory functions, and build a customized conversational dataset. 
    \item \textbf{Evaluation Framework}: We propose a new evaluation framework that systematically assesses multimodal long-term conversational memory across memory extraction \& adaptation, reasoning, and knowledge management.
    \item \textbf{Benchmark Takeaways}: Through extensive benchmarking, we reveal key advantages and limitations of existing memory designs in multimodal long-term conversations, providing actionable insights for future research.
\end{itemize}

%% file: 2_related_work.tex
\section{Related Works}
\label{sec:related}

\begin{table*}[t!]
  \centering
  \caption{Comparison between \model with representative conversational benchmarks. \cmark: Satisfies; \xmark: Does not satisfy; \pmark: Text modality only.}
  \vspace{-0.5em}
  \resizebox{\linewidth}{!}{%
     \begin{tabular}{l|cccc|ccc|ccc|ccc}
    \toprule
\multirow{2}[4]{*}{Benchmark} & \multicolumn{4}{c|}{Conversational Characteristics} & \multicolumn{3}{c|}{Extract.\&Adapt.} & \multicolumn{3}{c|}{Reasoning} & \multicolumn{3}{c}{Management} \\
\cmidrule{2-14}          & A. Round & A. Img. & Multi-Sess. & MM Info. &   FR  & VS    & TTL    & TR    & VR    & MR    & KR  & CD   & AR \\
    \midrule
    DuLeMon~\cite{xu2022long} & 8.16 & --- & \cmark & \xmark & \pmark & \xmark & \xmark & \xmark & \xmark & \xmark & \xmark & \xmark & \xmark \\
    DialogBench~\cite{ou2024dialogbench} & 7.48 & --- & \cmark & \xmark & \pmark & \xmark & \xmark & \xmark & \xmark & \pmark & \pmark & \xmark & \xmark \\
    MemoryBank~\cite{zhong2024memorybank} & 3.77 & --- & \cmark & \xmark & \pmark & \xmark & \xmark & \pmark & \xmark & \xmark & \xmark & \xmark & \xmark \\
    MMDU~\cite{liu2024mmdu} & 14.95 & 3.83 & \xmark & \cmark & \cmark & \xmark & \xmark & \xmark & \cmark & \cmark & \xmark & \xmark & \xmark\\
    LoCoMo~\cite{maharana2024evaluating} & 10.81 & 3.35 & \cmark & \cmark & \cmark & \xmark & \xmark & \pmark & \xmark & \cmark & \xmark & \xmark & \pmark \\
    LOCCO~\cite{jia2025evaluating} & 4.77 & --- & \cmark & \xmark & \pmark & \xmark & \xmark & \xmark & \xmark & \xmark & \xmark & \xmark & \xmark \\
    LongMemEval~\cite{wu2025longmemeval} & 5.19 & --- & \cmark & \xmark & \pmark & \xmark & \xmark & \pmark & \xmark & \pmark & \pmark & \xmark & \pmark \\
    MemoryAgentBench~\cite{hu2025evaluating} & 9.55 & --- & \cmark & \xmark & \pmark & \xmark & \pmark & \pmark & \xmark & \pmark & \pmark & \xmark & \xmark \\
    MMRC~\cite{xue2025mmrc}  & 12.90 & 2.90 & \xmark & \cmark & \cmark & \xmark & \xmark & \cmark & \cmark & \cmark & \pmark & \xmark & \cmark \\
    \midrule
     \rowcolor{TopThree} \textbf{\name} (Ours) & \textbf{16.51} & \textbf{4.18} & \cmark & \cmark & \cmark & \cmark & \cmark & \cmark & \cmark & \cmark & \cmark & \cmark & \cmark \\
    \bottomrule
    \end{tabular}%
    }
     \label{tab:bench_comparison}%
     \\
     \vspace{1.5pt} 
     \parbox{\textwidth}{ 
\fontsize{7}{7}\selectfont
\textbf{*} \textbf{A. Round}: average user-assistant dialogue round numbers per session. \textbf{A. Img.}: average image numbers per session. \textbf{Multi-Sess. \& MM Info.}: whether the dataset has basic characteristics of multi-session and multimodal information. This property serves as a fundamental prerequisite for multimodal long-term conversational memory. For the three evaluation dimensions: (i) \textbf{Memory extraction \& adaptation} includes factual retrieval (FR), visual-centric search (VS), and test-time learning (TTL) subtasks. (ii) \textbf{Memory reasoning} includes temporal reasoning (TR), visual-centric reasoning (VR), and multi-entity reasoning (MR) subtasks. (iii) \textbf{Memory knowledge management} includes knowledge resolution (KR), conflict detection (CD), and answer refusal (AR) subtasks. 
}
\vspace{-0.8em}
\end{table*}%

\subsection{Multi-Round Conversational Benchmark}

A number of dialogue benchmarks have been proposed in recent years that can be used to evaluate memory capabilities, such as LoCoMo~\cite{maharana2024evaluating}, LongMemEval~\cite{wu2025longmemeval}, and MemoryAgentBench~\cite{hu2025evaluating}. However, most of these benchmarks are text-only and do not provide an evaluation of multimodal capabilities. Although several localized multimodal dialogue benchmarks have been introduced recently, like MMDU~\cite{liu2024mmdu} and MMRC~\cite{xue2025mmrc}, they are single-session only and lack the multi-session conversation structure. Therefore, they are used to assess multi-round context understanding abilities rather than long-term memory.

\begin{figure}[t!]
\centering
\includegraphics[width=\linewidth]{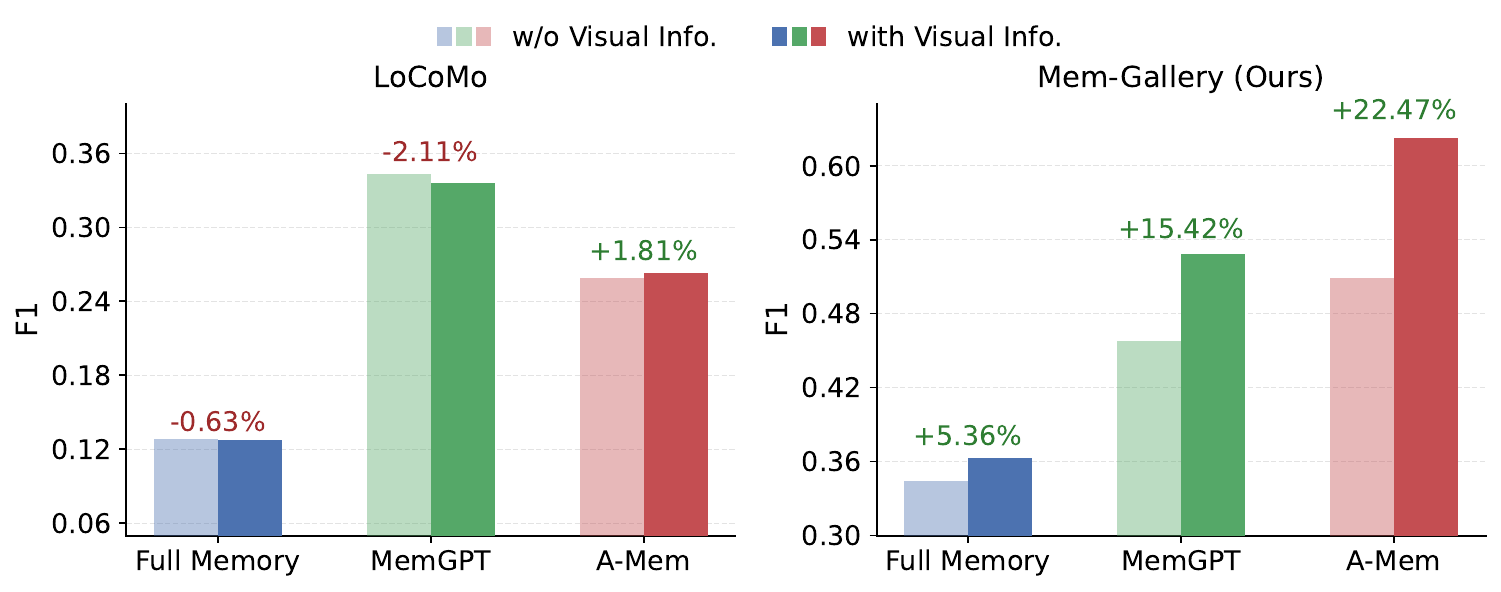}
\caption{Effectiveness analysis of visual information with representative memory models. Compared with LoCoMo~\cite{maharana2024evaluating}, visual information plays a much more critical role in \name.} 
\vspace{-0.8em}
\label{fig:comp_locomo}
\end{figure}

We compare \name with representative related works in Table~\ref{tab:bench_comparison}.
Overall, prior benchmarks exhibit a structural misalignment with multimodal long-term memory evaluation, \textit{either overlooking visual information, lacking multi-session structure, or failing to support the assessment of multimodal memory functionalities}.
Among them, LoCoMo is one of the few that incorporates visual information and a multi-session structure.
However, it supports a very restricted multimodal memory functionality evaluation, which is thus always used for text-only memory evaluation~\cite{xu2025mem,fang2025lightmem}.
Furthermore, Figure~\ref{fig:comp_locomo} shows that incorporating visual information in LoCoMo yields marginal or inconsistent gains, indicating that its evaluation tasks can largely be solved without visual clues and thus lack sufficient capacity to assess multimodal long-term memory. 
This highlights the need for a new benchmark that systematically integrates task-critical multimodal information into multi-session conversations, enabling a comprehensive evaluation of multimodal memory capabilities.

\subsection{Long-Term Agent Memory}
Real-world tasks typically require agents to interact with their environments in a multi-round and dynamic manner, e.g., multi-round dialogues in conversational agents, making long-term memory an important capability for agents~\cite{zhang2025survey,wei2025evo}. 
Prior works have explored different aspects of memory design for such agents. For example, Generative agents~\cite{park2023generative} introduced the concept of memory flow for social event simulation. A-Mem~\cite{xu2025mem} and MemoryOS~\cite{kang2025memory} designed agentic memory construction and maintenance mechanisms.
However, existing methods primarily focus on the textual modality, while real-world memory often requires the joint integration of multimodal information. Consequently, multimodal long-term memory has recently garnered growing interest, such as ViLoMem~\cite{bo2025agentic} and M3-Agent~\cite{long2025seeing}. 
Although recent M3-Bench~\cite{long2025seeing} has taken a step toward evaluating multimodal long-term memory with long-video QA, its setup differs fundamentally from conversational memory with multi-round interactions. Key challenges to long-term conversational memory, such as multimodal information being incrementally introduced, referenced, and regulated across sessions for personalized assistants, remain largely underexplored.

\begin{figure*}[t!]
\centering
\includegraphics[width=0.985\linewidth]{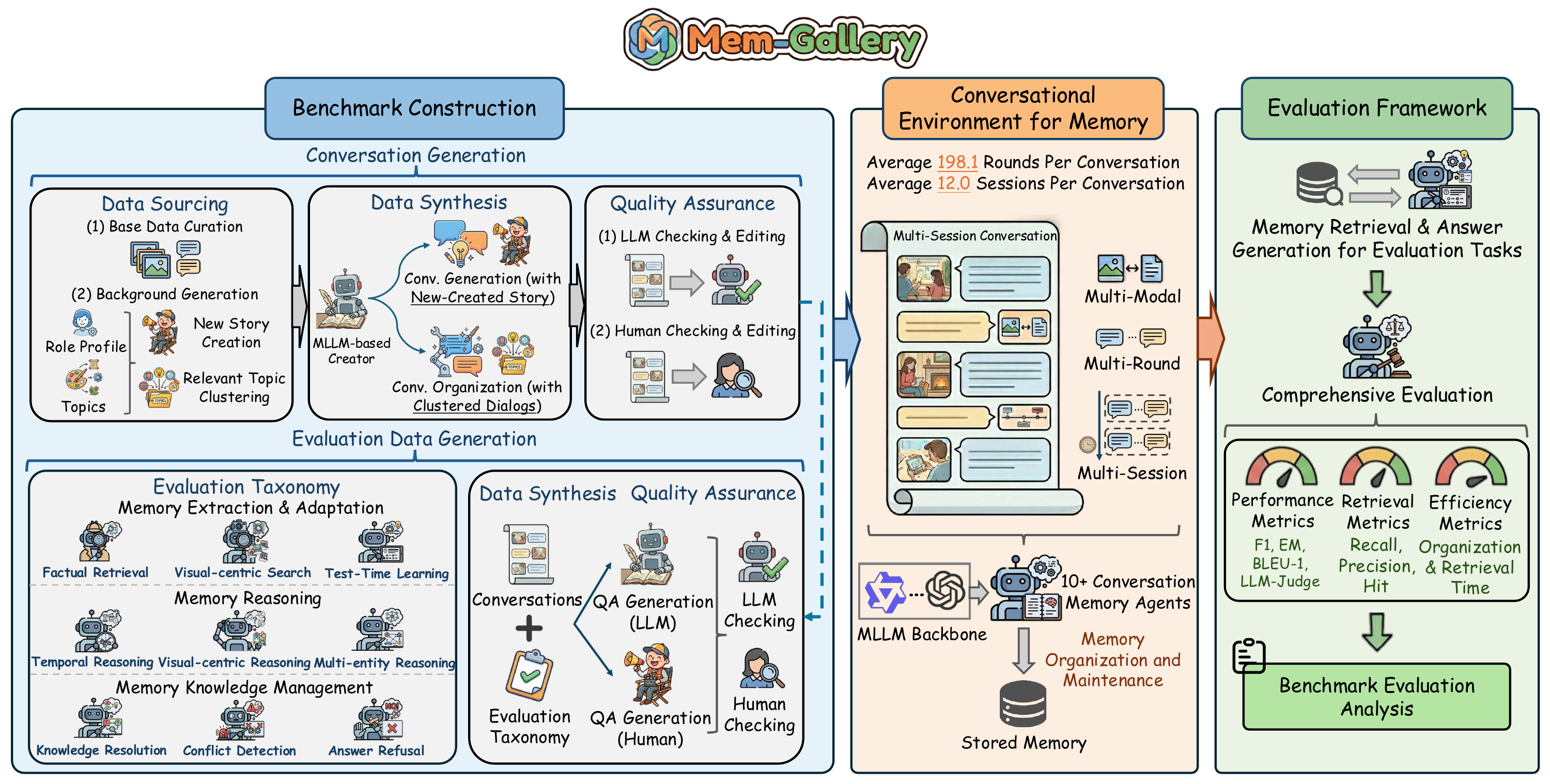}
\caption{The overall pipeline of our proposed \model.} 
\vspace{-0.7em}
\label{fig:main_fig}
\end{figure*} 

%% file: 3_method.tex
\section{\model Benchmark}
\label{sec:model}

As shown in Figure~\ref{fig:main_fig}, we describe \model from three aspects: (1) the benchmark construction, (2) the unified conversational environment, and (3) the evaluation framework and task taxonomies.

\subsection{Benchmark Construction}
The statistics of the new dataset can be found in Table~\ref{tab:base_stat}. Detailed dataset construction and statistics can be found in Appendix~\ref{sec:data_detail_appendix}.

\begin{table}[t]
\centering
\caption{Statistics of the \model dataset.}
\vspace{-0.5em}
\resizebox{\linewidth}{!}{%
\begin{tabular}{l|lc}
\toprule
\textbf{\model} & \textbf{Aspect} & \textbf{Statistics} \\
\midrule
\multirow{3}{*}{Conversation Data}
 & Sessions & 240 \\
 & Dialogue Rounds & 3{,}962 \\
 & Included Images & 1{,}003 \\
\midrule
\multirow{2}{*}{Evaluation Data}
 & QA Pairs with Annotated Clues  & 1{,}711 \\
 & Included Images & 487 \\
\bottomrule
\end{tabular}}\label{tab:base_stat}
\vspace{-0.6em}
\end{table}

\subsubsection{Conversation Data}
To support realistic long-term memory evaluation, the conversation data is organized as coherent multi-session interactions, with consistent user personas and tightly coupled visual \& textual content.

\textbf{Data Sourcing}. We first curate base materials from open-source resources, including images and parts of the textual content (details in Appendix~\ref{sec:data_source}). 
Specifically, we prioritize publicly available data that cover diverse everyday and domain-specific scenarios, with sufficient visual detail and semantic richness to support long-horizon multimodal grounding, rather than isolated or short-context understanding. On top of the curated base data, we further generate structured conversational backgrounds, such as user role profiles and conversation topics. These backgrounds serve as controllable anchors for conversation synthesis, ensuring topic diversity and long-range coherence.

\textbf{Data Synthesis}. Based on the sourced materials, we synthesize multi-session conversations through two complementary strategies, each producing valid multi-session conversations that are later unified into the final dataset.
On one hand, we adopt conversation generation with newly created stories as conversation backgrounds,
where human annotators design the story outline and inter-session transition logic. Advanced LLMs generate the text-part multi-session dialogues conditioned on these specifications. Annotators then insert appropriate images at suitable positions to ensure multimodal dependency.
On the other hand, we perform conversation organization via topic-based clustering, where existing single-session multimodal dialogues in MMRC~\cite{xue2025mmrc} are grouped into candidate multi-session conversations. Specifically, LLMs are used to extract representative topic keywords from single-session dialogues, followed by clustering under constraints on topical relevance, fluency, and length. The clustered conversations are then reordered and refined to form coherent long-term interaction sequences.
Details and the example can be found in Appendix~\ref{sec:conv_data_gen_appendix}.

\textbf{Quality Assurance}. We apply a two-stage quality assurance process. In the first stage, advanced LLMs automatically check and revise conversations for coherence, fluency, and factual consistency. In the second stage, human annotators carefully review each conversation and further refine the content through manual editing. This ensures the final conversation data maintains high multimodal quality and realistic conversation dynamics.

\subsubsection{Evaluation Data}
Based on the curated conversations, we systematically generate evaluation data. It consists of QA pairs based on a well-predefined evaluation taxonomy (details in Section~\ref{sec:eval_frame}) for each conversation.

\textbf{Data Synthesis}. Evaluation QA pairs are also constructed through two complementary ways, which jointly contribute to the final evaluation set.
On one hand, LLMs are prompted with selected conversation histories and task descriptions to generate QA candidates. On the other hand, human annotators construct QA pairs by reviewing each conversation and designing targeted questions. These two ways jointly ensure both coverage, difficulty, and diversity of evaluation instances.
In addition, for each QA pair, we explicitly annotate evidence clues for the correct answer that specify dialogue turns in which the relevant information is referenced for the answer. These annotated clues facilitate fine-grained analysis of memory behaviors, e.g., retrieval details (analyzed in Appendix~\ref{sec:retrieval_number}), beyond final task performance. Data synthesis details can be found in Appendix~\ref{sec:eval_data_gen_appendix}.

\textbf{Quality Assurance}. Generated QA pairs also undergo a two-stage verification. The same as conversation generation, LLMs are first used to check answer correctness and question clarity.
This verification step is followed by careful human revision.

\subsection{Conversational Environment for Memory}
Following previous works~\cite{maharana2024evaluating}, memory models are evaluated based on the input conversation.  
In our setting, memory agents must go beyond storing textual information in existing benchmarks and store and associate visual content with textual content. For textual memory, image captions are provided to preserve the visual content.

\textbf{Conversation Structure}. As shown in Figure~\ref{fig:conv_struct}, each conversation spans multiple sessions with temporal gaps. Within each session, agents engage in multi-round multimodal dialogues, while across sessions, information may be updated or contradicted. This design prevents reliance on short-term context and explicitly requires memory to persist and evolve across long-term conversational boundaries.
Visual content in the conversation is not limited to single-round perception but may be referenced and integrated with textual context across sessions. Consequently, agents must integrate multimodal clues distributed over extended timelines, rather than treating images as isolated observations.

\textbf{Benchmarking Memory Methods}.
We benchmark thirteen representative memory models under a unified setting. As conversations progress, agents accumulate an expanding memory. To ensure a fair and controlled comparison across diverse designs, all evaluated models follow a unified memory accumulation granularity and protocol~\cite{kang2025memory}, where information is incrementally stored along with the conversational timeline. Memory retrieval and answer generation are then performed based on the accumulated memory.
The problem definition can be found in Appendix~\ref{sec:pd}.

\begin{figure}[t!]
\centering
\includegraphics[width=0.935\linewidth]{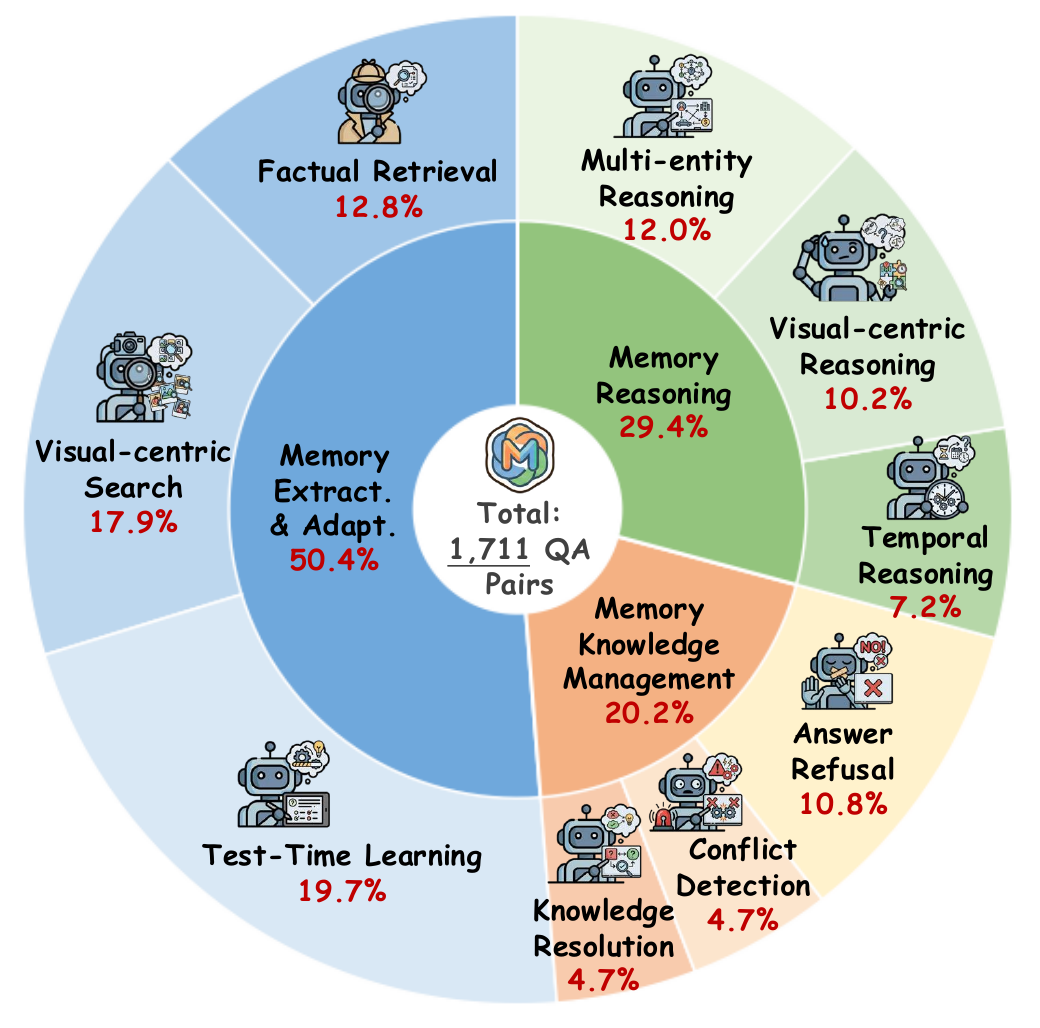}
\caption{Taxonomy and distribution of evaluation tasks.} 
\vspace{-0.8em}
\label{fig:evaluate_dist}
\end{figure}

\subsection{Evaluation Framework}\label{sec:eval_frame}
As shown in Figure~\ref{fig:evaluate_dist}, we design three task families to systematically evaluate multimodal long-term memory as an evolving capability in agentic systems over extended conversations. 
Specifically, memory extraction and adaptation assess whether agents can retrieve relevant multimodal information and adapt it as memory accumulates. Memory reasoning examines whether agents can integrate and reason over memory, accounting for temporal dependencies, visual evidence, and multiple entities. Memory knowledge management focuses on how agents regulate long-term memory under dynamic states, like handling inconsistencies, resolving conflicts, and refusing to answer when information is outdated, contradictory, or incomplete.

\subsubsection{Memory Extraction \& Adaptation}
At a fundamental level, memory agents must possess the capability of memory extraction and adaptation, ensuring that stored information can be effectively utilized~\cite{hu2025evaluating}.
Based on the properties, we design three subtask categories.
(1) \textit{\textbf{Factual Retrieval}}: evaluate the ability to accurately recall factual details (e.g., user preferences and past events) from multimodal interaction histories.
(2) \textit{\textbf{Visual-centric Search}}: evaluate whether the model can identify or retrieve specific visual instances from memory (e.g., shared images).
(3) \textit{\textbf{Test-Time Learning}}: measure the ability to adapt its memory to unseen multimodal examples at inference time.

\subsubsection{Memory Reasoning}
In addition to memory extraction, agents should be equipped with reasoning capabilities over the memory to address complex tasks~\cite{ke2025survey}.
We design three subtasks accordingly.
(1) \textit{\textbf{Temporal Reasoning}}: assess whether the model can synthesize and reason over temporally dependent questions from multimodal memory. (2) \textit{\textbf{Visual-centric Reasoning}}: test the model's capacity to retrieve and utilize visual information as cues for reasoning on multimodal memories. (3) \textit{\textbf{Multi-entity Reasoning}}: reasoning with multiple entities from memory, where each can be textual or visual.

\subsubsection{Memory Knowledge Management}
Unlike task-oriented scenarios such as web navigation, conversations are inherently more open-ended. Users may update previous information or provide incorrect details inadvertently during dialogues~\cite{hu2025evaluating}. To evaluate these challenges, we design the following subtasks.
(1) \textit{\textbf{Knowledge Resolution}}: Examine the ability to correctly update stored knowledge when new, contradictory information appears in the dialogue, maintaining consistency over time. (2) \textit{\textbf{Conflict Detection}}: Test whether the model can detect conflicts between newly observed information and existing memory. (3) \textit{\textbf{Answer Refusal}}: Assess the model's capability to abstain or refuse to answer when the requested information is unsupported by prior memory.

%% file: 4_experiment.tex
\section{Benchmarking Analysis}
\label{sec:experiment}
We conduct extensive experiments on our proposed \model for the following research questions.
\begin{itemize}[leftmargin=*, itemsep=0pt]
\vspace{-0.3em}
    \item \textbf{RQ1}: How effective are different multimodal storage designs?
    \item \textbf{RQ2}: How important are memory organization strategies in multimodal memory systems? 
    \item \textbf{RQ3}: What are the strengths and weaknesses of different memory models across task types?
    \item \textbf{RQ4}: What are the runtime efficiency characteristics of different memory approaches?
    \item \textbf{RQ5}: How does the number of retrieved memory entries affect the overall performance?
\end{itemize}

\subsection{Evaluation Setup}
\subsubsection{Model Implementation}
\paragraph{MLLM Backbones.} To ensure a broad coverage of model capacities, we use representative open-source MLLMs at two model scales, Qwen2.5-VL-3B-Instruct and Qwen2.5-VL-7B-Instruct~\cite{bai2025qwen2}, as well as representative closed-source MLLMs, namely GPT-4.1-Nano~\cite{achiam2023gpt} and Gemini-2.5-Flash-Lite~\cite{comanici2025gemini}, as backbone models. We adopt Qwen2.5-VL-7B-Instruct as our default MLLM backbone.

\paragraph{Memory Models.} We comprehensively include thirteen representative approaches, including eight textual memory methods and five multimodal memory methods. 
Specifically, the \textbf{textual memory} includes Full Memory (Text), First-in-first-out (FIFO), NaiveRAG, Generative Agents~\cite{park2023generative}, Reflexion~\cite{shinn2023reflexion}, MemGPT~\cite{packer2023memgpt}, A-Mem~\cite{xu2025mem}, and MemoryOS~\cite{kang2025memory}. 
The \textbf{multimodal memory} includes Full Memory (Multimodal), MuRAG~\cite{chen2022murag}, UniversalRAG~\cite{yeo2025universalrag}, NGM~\cite{fisher2025neural}, and AUGUSTUS~\cite{jain2025augustus}. 
Details of these models are provided in Appendix~\ref{sec:model_detail_appendix}. For methods that require a top-$K$ retriever, we adopt a default retrieval size of $K$=10.
To enable a fair comparison with multimodal memory systems, we provide textual memory with high-quality image captions generated by GPT-5.1. The detailed benchmark setup for fair comparison is shown in Appendix~\ref{sec:bench_eval_details}.

\begin{table*}[t!]
  \centering
  \caption{Main evaluation results on our \name based on Qwen-2.5-VL-7B. The best and second-performed memory model(s) are highlighted with \textcolor{orange}{orange} and \textcolor{blue}{blue} backgrounds. Results on Qwen-2.5-VL-3B, GPT-4.1-Nano, and Gemini-2.5-Flash-Lite can be found in Table~\ref{tab:result_qwen_vl_3b}, Table~\ref{tab:gpt_4_1_nano}, and Table~\ref{tab:gemini_flash_lite}, respectively.}
  \vspace{-0.5em}
  \resizebox{\linewidth}{!}{%
    \begin{tabular}{ccc|cccccccc|ccccc}
    \toprule
    \rowcolor{qwenbar} \multicolumn{3}{c|}{\qwenlogo\ Qwen-2.5-VL-7B} & Full (Text) & FIFO  & NaiveRAG & Gen. Agent & Reflexion & MemGPT & A-Mem & MemoryOS & Full (MM) & MuRAG & UniversalRAG & NGM   & AUGUSTUS \\
    \midrule
    \multirow{12}[6]{*}{\rotatebox{90}{Extract. \& Adapt.}} & \multirow{4}[2]{*}{FR} & F1    & 0.2376 & 0.1446 & 0.5852 & 0.2424 & 0.2391 & 0.5928 & 0.6072 & 0.6244 & 0.2150 & \cellcolor{TopOne} 0.6724 & \cellcolor{TopTwo} 0.6632 & 0.6364 & 0.6162 \\
          &       & BLEU-1 & 0.1865 & 0.1000 & 0.5045 & 0.1862 & 0.1903 & 0.5098 & 0.5138 & 0.5354 & 0.1626 & \cellcolor{TopOne} 0.5755 & \cellcolor{TopTwo} 0.5658 & 0.5606 & 0.5309 \\
          &       & EM    & 0.0913 & 0.0457 & 0.3059 & 0.0868 & 0.0913 & 0.3288 & 0.2922 & 0.3470 & 0.0685 & \cellcolor{TopTwo} 0.3607 & 0.3470 & \cellcolor{TopOne} 0.3744 & 0.3288 \\
          &       & LLM-Judge & 0.2626 & 0.1324 & 0.7763 & 0.2945 & 0.2626 & 0.8539 & 0.7808 & 0.8265 & 0.2260 & \cellcolor{TopOne} 0.8790 & \cellcolor{TopTwo} 0.8744 & 0.8082 & 0.8082 \\
\cmidrule{2-16}          & \multirow{4}[2]{*}{VS} & F1    & 0.1992 & 0.0612 & 0.7592 & 0.2970 & 0.1954 & 0.6239 & 0.7681 & 0.7853 & 0.1658 & \cellcolor{TopOne} 0.8818 & \cellcolor{TopTwo} 0.8708 & 0.8531 & 0.8499 \\
          &       & BLEU-1 & 0.1873 & 0.0549 & 0.7063 & 0.2616 & 0.1840 & 0.5834 & 0.6880 & 0.7181 & 0.1473 & \cellcolor{TopOne} 0.8442 & \cellcolor{TopTwo} 0.8343 & 0.8145 & 0.7942 \\
          &       & EM    & 0.1601 & 0.0392 & 0.5686 & 0.2124 & 0.1569 & 0.4118 & 0.5882 & 0.6046 & 0.1078 & \cellcolor{TopTwo} 0.6699 & \cellcolor{TopTwo} 0.6699 & \cellcolor{TopOne} 0.6863 & 0.6536 \\
          &       & LLM-Judge & 0.1961 & 0.0556 & 0.7402 & 0.2958 & 0.1895 & 0.5964 & 0.7369 & 0.7729 & 0.1683 & \cellcolor{TopOne} 0.8856 & \cellcolor{TopTwo} 0.8611 & 0.8480 & 0.8399 \\   
\cmidrule{2-16}    & \multirow{4}[2]{*}{TTL} & F1    & 0.4500 & 0.3351 & 0.6526 & 0.4851 & 0.4486 & 0.2924 & 0.6336 & 0.5484 & 0.4147 & \cellcolor{TopOne} 0.8177 & 0.7824 & 0.7817 & \cellcolor{TopTwo} 0.7913 \\
          &       & BLEU-1 & 0.3799 & 0.2692 & 0.5835 & 0.4150 & 0.3798 & 0.2295 & 0.5595 & 0.4697 & 0.3477 & \cellcolor{TopOne} 0.7449 & 0.7103 & 0.7168 & \cellcolor{TopTwo} 0.7206 \\
          &       & EM    & 0.2374 & 0.1365 & 0.4718 & 0.2967 & 0.2374 & 0.1009 & 0.4362 & 0.3561 & 0.2107 & \cellcolor{TopOne} 0.6172 & \cellcolor{TopTwo} 0.6142 & 0.6024 & 0.6113 \\
          &       & LLM-Judge & 0.7092 & 0.6677 & 0.8457 & 0.7582 & 0.7033 & 0.7092 & 0.7997 & 0.7715 & 0.7107 & \cellcolor{TopTwo} 0.9006 & 0.8501 & \cellcolor{TopOne} 0.9110 & 0.8932 \\
      
    \midrule
    \multirow{12}[6]{*}{\rotatebox{90}{Reasoning}} & \multirow{4}[2]{*}{TR} & F1    & 0.2545 & 0.1549 & 0.4887 & 0.2742 & 0.2553 & 0.5661 & 0.5604 & 0.5497 & 0.2294 & \cellcolor{TopOne} 0.5833 & 0.5460 & 0.5425 & \cellcolor{TopTwo} 0.5800 \\
          &       & BLEU-1 & 0.2363 & 0.1316 & 0.4587 & 0.2473 & 0.2363 & 0.5326 & 0.5361 & 0.5240 & 0.2065 & \cellcolor{TopOne} 0.5537 & 0.5137 & 0.5143 & \cellcolor{TopTwo} 0.5527 \\
          &       & EM    & 0.1545 & 0.0894 & 0.3496 & 0.1626 & 0.1545 & 0.3496 & 0.4065 & 0.3821 & 0.1382 & \cellcolor{TopTwo} 0.4309 & 0.4065 & 0.4146 & \cellcolor{TopOne} 0.4390 \\
          &       & LLM-Judge & 0.2805 & 0.1463 & 0.6382 & 0.3252 & 0.2764 & \cellcolor{TopOne} 0.8008 & 0.6951 & 0.7195 & 0.2480 & \cellcolor{TopTwo} 0.7724 & 0.7520 & 0.7195 & 0.7398 \\
\cmidrule{2-16}   & \multirow{4}[2]{*}{VR} & F1    & 0.2552 & 0.1207 & 0.3022 & 0.1955 & 0.2594 & 0.4593 & 0.4477 & 0.4280 & 0.2015 & \cellcolor{TopTwo} 0.4818 & \cellcolor{TopOne} 0.4879 & 0.4615 & 0.3866 \\
          &       & BLEU-1 & 0.2442 & 0.1005 & 0.2873 & 0.1815 & 0.2480 & 0.4459 & 0.4331 & 0.4126 & 0.1912 & \cellcolor{TopTwo} 0.4625 & \cellcolor{TopOne} 0.4682 & 0.4428 & 0.3726 \\
          &       & EM    & 0.2011 & 0.0690 & 0.1954 & 0.1207 & 0.2011 & \cellcolor{TopOne} 0.3851 & 0.3563 & 0.3391 & 0.1609 & \cellcolor{TopTwo} 0.3793 & \cellcolor{TopTwo} 0.3793 & 0.3678 & 0.2931 \\
          &       & LLM-Judge & 0.3046 & 0.1408 & 0.3793 & 0.2471 & 0.3046 & \cellcolor{TopOne} 0.6149 & 0.5747 & 0.5805 & 0.2586 & \cellcolor{TopTwo}  0.6092 & 0.5977 & 0.5460 & 0.4856 \\       
\cmidrule{2-16}   & \multirow{4}[2]{*}{MR} & F1    & 0.2411 & 0.1745 & 0.4640 & 0.2450 & 0.2428 & 0.4367 & 0.5000 & 0.4490 & 0.2101 & \cellcolor{TopTwo} 0.5007 & \cellcolor{TopOne} 0.5013 & 0.4746 & 0.4866 \\
          &       & BLEU-1 & 0.1739 & 0.1184 & 0.3543 & 0.1784 & 0.1770 & 0.3347 & \cellcolor{TopOne} 0.3908 & 0.3446 & 0.1429 & \cellcolor{TopTwo} 0.3903 & 0.3868 & 0.3635 & 0.3778 \\
          &       & EM    & 0.0340 & 0.0146 & 0.0874 & 0.0243 & 0.0340 & 0.0631 & 0.0728 & 0.0728 & 0.0194 & \cellcolor{TopTwo} 0.0874 & \cellcolor{TopOne} 0.0971 & 0.0728 & 0.0825 \\
          &       & LLM-Judge & 0.2985 & 0.1602 & 0.7791 & 0.3350 & 0.3058 & 0.8204 & 0.8083 & 0.8204 & 0.2791 & \cellcolor{TopOne} 0.8447 & 0.8422 & 0.7840 & 0.8228 \\    
    \midrule
    \multirow{12}[6]{*}{\rotatebox{90}{Knowledge Management}} & \multirow{4}[2]{*}{KR} & F1    & 0.2354 & 0.1697 & 0.3595 & 0.2674 & 0.2354 & 0.4292 & 0.4515 & \cellcolor{TopOne} 0.5031 & 0.2181 &  \cellcolor{TopTwo} 0.4601 & 0.4336 & 0.3944 & 0.3752 \\
          &       & BLEU-1 & 0.2005 & 0.1424 & 0.2986 & 0.2341 & 0.2000 & 0.3741 & 0.3992 & \cellcolor{TopOne} 0.4479 & 0.1883 & \cellcolor{TopTwo} 0.4073 & 0.3772 & 0.3444 & 0.3159 \\
          &       & EM    & 0.1235 & 0.0864 & 0.1605 & 0.1235 & 0.1235 & 0.2099 & 0.2469 & \cellcolor{TopOne} 0.2716 & 0.1235 & \cellcolor{TopTwo} 0.2593 & 0.2099 & 0.2099 & 0.1728 \\
          &       & LLM-Judge & 0.3395 & 0.2469 & 0.6358 & 0.3457 & 0.3395 & 0.7407 & 0.6667 & \cellcolor{TopTwo} 0.7593 & 0.2840 & \cellcolor{TopOne} 0.7840 & 0.7222 & 0.6728 & 0.6728 \\
\cmidrule{2-16}          & \multirow{4}[2]{*}{CD} & F1    & 0.3457 & \cellcolor{TopTwo} 0.3580 & 0.3457 & 0.3210 & 0.3333 & \cellcolor{TopTwo} 0.3580 & 0.3333 & 0.3333 & \cellcolor{TopTwo} 0.3580 & \cellcolor{TopOne} 0.3704 & 0.3457 & 0.3457 & 0.3210 \\
          &       & BLEU-1 & 0.3457 & \cellcolor{TopTwo} 0.3580 & 0.3457 & 0.3210 & 0.3333 & \cellcolor{TopTwo}  0.3580 & 0.3333 & 0.3333 & \cellcolor{TopTwo} 0.3580 & \cellcolor{TopOne} 0.3704 & 0.3457 & 0.3457 & 0.3210 \\
          &       & EM    & 0.3457 & \cellcolor{TopTwo} 0.3580 & 0.3457 & 0.3210 & 0.3333 & \cellcolor{TopTwo} 0.3580 & 0.3333 & 0.3333 & \cellcolor{TopTwo} 0.3580 & \cellcolor{TopOne} 0.3704 & 0.3457 & 0.3457 & 0.3210 \\
          &       & LLM-Judge & 0.3457 & \cellcolor{TopTwo} 0.3580 & 0.3457 & 0.3210 & 0.3333 & \cellcolor{TopTwo} 0.3580 & 0.3333 & 0.3333 & \cellcolor{TopTwo} 0.3580 & \cellcolor{TopOne} 0.3704 & 0.3457 & 0.3457 & 0.3210 \\
\cmidrule{2-16} & \multirow{4}[2]{*}{AR} & F1    & \cellcolor{TopTwo} 0.9958 & \cellcolor{TopOne} 1.0000 & 0.9581 & 0.9841 & \cellcolor{TopTwo} 0.9958 & 0.9849 & 0.9278 & 0.9845 & 0.9946 & 0.9418 & 0.9473 & 0.9580 & 0.9460 \\
          &       & BLEU-1 & \cellcolor{TopTwo} 0.9953 & \cellcolor{TopOne} 1.0000 & 0.9575 & 0.9839 & \cellcolor{TopTwo} 0.9953 & 0.9844 & 0.9257 & 0.9841 & 0.9946 & 0.9409 & 0.9466 & 0.9574 & 0.9459 \\
          &       & EM    & \cellcolor{TopTwo} 0.9946 & \cellcolor{TopOne} 1.0000 & 0.9565 & 0.9837 & \cellcolor{TopTwo} 0.9946 & 0.9837 & 0.9239 & 0.9837 & 0.9946 & 0.9402 & 0.9457 & 0.9565 & 0.9457 \\
          &       & LLM-Judge & \cellcolor{TopTwo} 0.9783 & \cellcolor{TopOne} 0.9837 & 0.9402 & 0.9674 &\cellcolor{TopTwo}   0.9783 & 0.9674 & 0.9375 & 0.9674 & \cellcolor{TopOne} 0.9837 & 0.9375 & 0.9429 & 0.9457 & 0.9429 \\         
    \midrule
    \multicolumn{2}{c}{\multirow{4}[2]{*}{Overall}} & F1    & 0.3625 & 0.2724 & 0.5974 & 0.3825 & 0.3619 & 0.5282 & 0.6228 & 0.6109 & 0.3354 & \cellcolor{TopOne} 0.6966 & \cellcolor{TopTwo} 0.6827 & 0.6691 & 0.6610 \\
    \multicolumn{2}{c}{} & BLEU-1 & 0.3279 & 0.2408 & 0.5441 & 0.3422 & 0.3279 & 0.4792 & 0.5629 & 0.5534 & 0.2999 & \cellcolor{TopOne} 0.6432 & \cellcolor{TopTwo} 0.6286 & 0.6200 & 0.6069 \\
    \multicolumn{2}{c}{} & EM    & 0.2519 & 0.1835 & 0.4161 & 0.2613 & 0.2507 & 0.3402 & 0.4296 & 0.4278 & 0.2279 & \cellcolor{TopOne} 0.4985 & 0.4927 & \cellcolor{TopTwo} 0.4944 & 0.4757 \\
    \multicolumn{2}{c}{} & LLM-Judge & 0.4331 & 0.3369 & 0.7241 & 0.4643 & 0.4307 & 0.7306 & 0.7431 & 0.7613 & 0.4129 & \cellcolor{TopOne} 0.8229 & \cellcolor{TopTwo} 0.8016 & 0.7861 & 0.7797 \\
    \bottomrule
    \end{tabular}%
    }
    \vspace{-0.8em}
  \label{tab:main_qwen_vl_7b}%
\end{table*}%

\subsubsection{Evaluation Metrics}
Following existing works, we evaluate memory performance using F1, BLEU-1, EM, and LLM-as-a-Judge metrics. For LLM-as-a-Judge, we adopt Qwen-2.5-72B-Instruct~\cite{bai2025qwen2} as the judging model. 
Note that the conflict detection task explicitly requires models to output either “Yes” or “No”. Since MLLMs possess instruction-following capability, the values of all metrics are identical for this task. 
For retrieval analysis, we adopt the widely used Recall, Precision, and Hit as the metrics. The details can be found in Appendix~\ref{sec:eval_metrics}.

\subsection{Main Results (RQ1-RQ3)}

Table~\ref{tab:main_qwen_vl_7b} illustrates the main benchmarking results. We can have the following observations.

\textbf{RQ1: Explicit multimodal memory preservation is critical, but increased architectural complexity does not necessarily bring better performance.}
Although with high-quality image captions, textual memory baselines still generally exhibit a performance gap compared to multimodal approaches, particularly in \textit{memory extraction and adaptation}. 
MuRAG, a simple multimodal method, can achieve the best overall performance among well-designed textual memory models. Specifically, it achieves 11.85\%, 7.69\%, 12.29\%, and 29.06\% F1 improvement over the best-performed textual memory on overall, FR, VS, and TTL tasks, respectively.
Furthermore, MuRAG and UniversalRAG, which simply preserve multimodal information without structured memory organizations, achieve stronger performance than complex multimodal memory systems. For example, MuRAG outperforms 4.11\% and 5.39\% overall F1 improvements over NGM and AUGUSTUS, respectively.
This contrast suggests that, beyond focusing on the memory architecture itself, how to effectively preserve multimodal information so as to maximize the utility of the memory architecture remains an open challenge.

\textbf{RQ2: Visual information faces harsher token consumption, making principled organization and maintenance critical.}
Without the proper organization, inputting all multimodal information into the context leads Full Memory (MM) to perform even worse than Full Memory (Text) in both memory extraction and reasoning tasks. Overall, Full Memory (MM) performs 8.08\% and 51.85\% worse F1 score than Full Memory (Text) and MuRAG.
As visual information is far more token-heavy than text~\cite{zhang2025llavamini}, under the token limit, naïvely accumulating multimodal content can introduce irrelevant visual noise, crowd out informative text, and thus hurt performance.
Besides, compared to MuRAG and UniversalRAG, although existing models like NGM and AUGUSTUS organize multimodal information in a structured manner, they still lack effective strategies for long-term conversational scenarios. Moreover, existing multimodal methods still lack agentic maintenance strategies like A-Mem and MemoryOS, which also constrain their performance.

\textbf{RQ3-1: Existing multimodal memory methods struggle on memory reasoning tasks.}
Although multimodal memory methods overall outperform textual memory approaches, we observe that their advantages are not pronounced in scenarios requiring reasoning. This is particularly evident in multi-entity reasoning and temporal reasoning tasks. Notably, even in the visual-centric reasoning task that explicitly relies on visual clues, MemGPT, a textual memory method, can achieve near-optimal performance. This suggests that existing multimodal memory approaches still primarily focus on the storage and retrieval of multimodal information, yet how to effectively reason over multimodal contents in memory remains largely unexplored.

\textbf{RQ3-2: Existing methods show limitations in scenarios involving information updates or conflicts.} 
From the memory knowledge management results, we derive two key observations.
First, there is a \textit{trade-off in refusal behavior}. Methods with weaker memory, e.g., FIFO, tend to exhibit stronger refusal performance, as they default to refusing when relevant information cannot be retrieved. In contrast, methods with stronger capabilities on memory extraction and reasoning show poorer refusal performance, like A-Mem and MuRAG, with 7.22\% and 5.82\% worse F1 score than FIFO. 
This suggests that future methods need to strike a balance between retrieving relevant information and distinguishing outdated and conflicting information to build safe and hallucination-free memory systems. 
Second, \textit{for both knowledge resolution and conflict detection tasks, neither multimodal nor textual memory methods achieve satisfactory performance}. This indicates that designing memory systems to support dynamic conflict detection and memory correction remains an important direction.

Full analysis under different MLLMs and case studies can be found in Appendix~\ref{sec:backbone} and~\ref{sec:case_study}.

\begin{figure}[tbp]
\centering
\includegraphics[width=\linewidth]{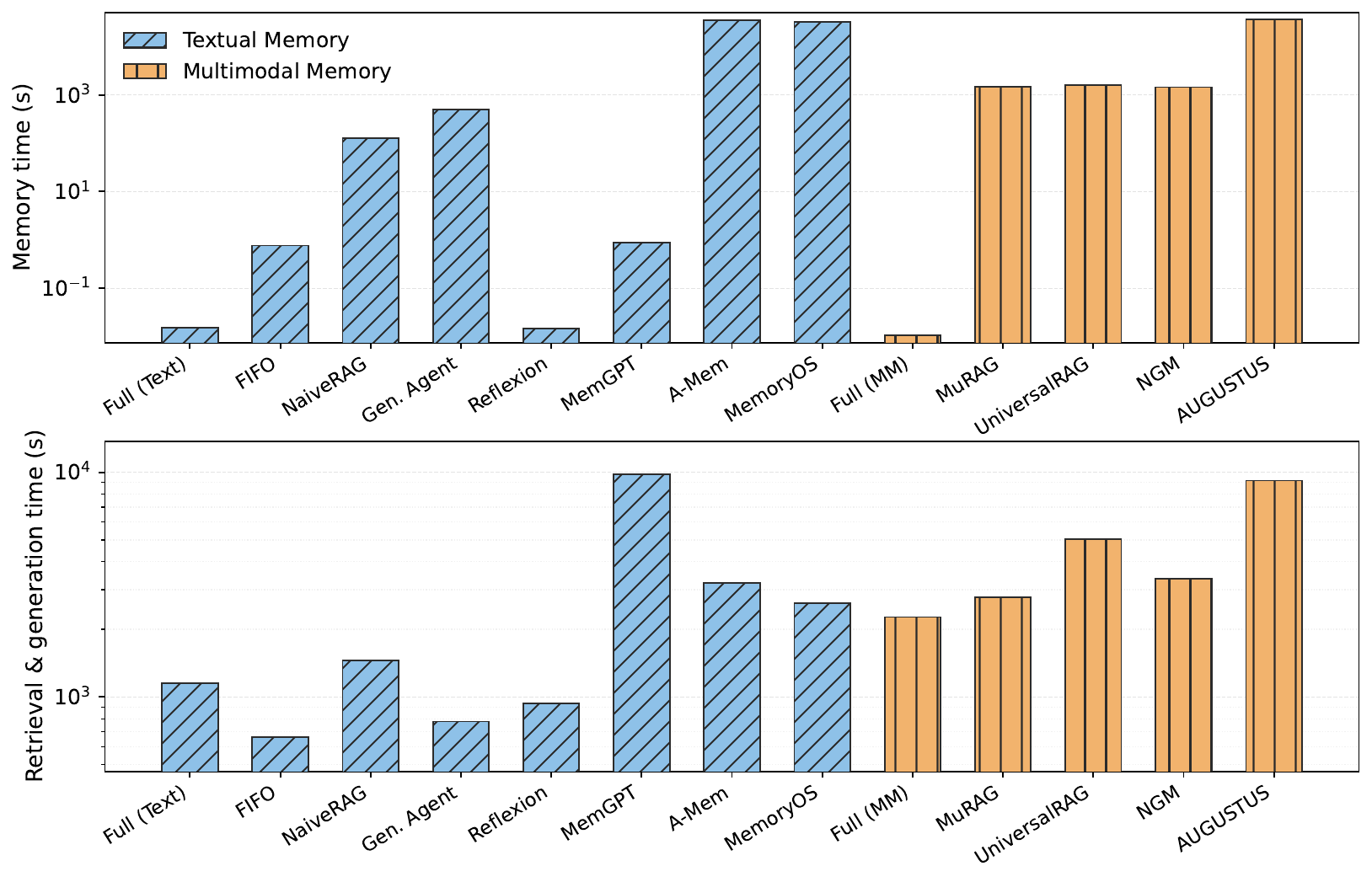}
\caption{Efficiency comparison results on information memorization time (top) and memory retrieval \& answer generation time (bottom) in seconds with the log scale.}
\label{fig:efficiency}
\vspace{-0.8em}
\end{figure}

\subsection{Efficiency Analysis (RQ4)}
Figure~\ref{fig:efficiency} presents an efficiency comparison of existing memory methods.
From the results, we can see that \textbf{multimodal memory incurs higher overhead than text-only memory in general}.  
Even a relatively simple multimodal memory like MuRAG can approach the computational cost of highly sophisticated textual memory systems, such as A-Mem and MemoryOS. This suggests that introducing multimodal signals fundamentally increases system complexity, regardless of whether advanced memory organization strategies are employed.
Thus, while multimodal memory can provide richer contextual information, it also introduces considerable inference-time overhead, which may limit its practicality in long-horizon or real-time agent scenarios.

\begin{table}[t]
  \centering
  \caption{Retrieval metrics on different size $K$.}
  \vspace{-0.6em}
  \resizebox{0.74\linewidth}{!}{%
    \begin{tabular}{l|cccc}
    \toprule
    Recall@$K$ & $K$=5     & $K$=10    & $K$=15    & $K$=20 \\
    \midrule
    Gen. Agent & 0.1707 & 0.2385 & 0.2854 & 0.3153 \\
    NaiveRAG & 0.5381 & 0.6723 & 0.7420 & 0.7877 \\
    MuRAG & \cellcolor{TopOne} 0.7506 & \cellcolor{TopOne} 0.8601 & \cellcolor{TopOne} 0.8990 & \cellcolor{TopOne} 0.9228 \\
    UniversalRAG & \cellcolor{TopTwo} 0.7311 &\cellcolor{TopTwo}  0.8411 & \cellcolor{TopTwo} 0.8781 & \cellcolor{TopTwo} 0.8998 \\
    NGM   & 0.6192 & 0.7475 & 0.7892 & 0.8065 \\
    AUGUSTUS & 0.6729 & 0.7529 & 0.7785 & 0.7860 \\
    \midrule
    Precision@$K$ & $K$=5     & $K$=10    & $K$=15    & $K$=20 \\
    \midrule
    Gen. Agent & 0.0955 & 0.0684 & 0.0544 & 0.0453 \\
    NaiveRAG & 0.2694 & 0.1757 & 0.1315 & 0.1047 \\
    MuRAG & \cellcolor{TopTwo} 0.3686 & 0.2220 & 0.1572 & 0.1220 \\
    UniversalRAG & \cellcolor{TopOne} 0.3691 & 0.2206 & 0.1555 & 0.1204 \\
    NGM   & 0.3564 & \cellcolor{TopOne} 0.3457 & \cellcolor{TopOne} 0.3450 & \cellcolor{TopOne} 0.3424 \\
    AUGUSTUS & 0.3341 & \cellcolor{TopTwo} 0.2488 & \cellcolor{TopTwo} 0.2213 & \cellcolor{TopTwo} 0.2124 \\
    \bottomrule
    \end{tabular}%
    }
  \label{tab:retrieval_num}%
  \vspace{-0.8em}
\end{table}%

\subsection{Retriever Analysis (RQ5)}

Table~\ref{tab:retrieval_num} shows multimodal retrievers like MuRAG and UniversalRAG achieve substantially higher clue recall. 
Though recall improves as $K$ increases, the \textbf{expanded retrieval coverage does not consistently translate into better QA performance}. 
Figure~\ref{fig:retrieval_num} shows multimodal memory models typically exhibit diminishing or saturated task gains beyond a moderate $K$, and may degrade at larger $K$. Table~\ref{tab:retrieval_num} explains this discrepancy. While the Recall of MuRAG and UniversalRAG increases as $K$ grows, its Precision drops sharply, indicating the introduction of substantial noise. 
NGM and AUGUSTUS exhibit more conservative recall growth with relatively stable Precision, suggesting structured organization or relevance filtering is an effective way to balance coverage and noise.
Full analysis and results are reported in Appendix~\ref{sec:retrieval_number}.

%% file: 5_conclusion.tex
\section{Conclusion}
\label{sec:conclusion}

In this paper, we present \model, a benchmark for evaluating multimodal long-term conversational memory. 
By grounding multi-session conversations in tightly coupled visual and textual contexts, \model enables systematic assessment beyond prior text-centric memory or localized multimodal context understanding benchmarks.
Extensive evaluation reveals several takeaways. 
These findings highlight the need for principled memory organization, selective retrieval, and robust maintenance for future multimodal memory design.

%% file: X_suppl.tex
\section{Appendix}
\label{sec:appendix}

\etocsettocdepth{3}
\localtableofcontents

\clearpage

\subsection{Problem Definition}\label{sec:pd}
\textbf{Notations}. We consider an MLLM-based agent interacting with an environment (user) over a sequence of time steps $t \in \{1, \dots, T\}$. Let $\mathcal{O} = \{o_1, o_2, \dots, o_T\}$ be a stream of multimodal observations, where each observation entry $o_t = \langle v_t, s_t \rangle$ may contain visual $v_t$ and textual $s_t$ modalities. The agentic memory system is defined by the tuple $\mathcal{S} = \langle \mathcal{M}, f_{\theta}, \Phi, \mathcal{R} \rangle$, where:

\begin{itemize}
    \item $\mathcal{M}_t = \{m_1, m_2, \dots, m_n\}$ is the external multimodal memory, an unbounded set of atomic units. Each unit $m_i$ encapsulates raw assets, cross-modal descriptions, and a joint latent embedding $\mathbf{e}_i$.
    \item $f_{\theta}$ is a multimodal encoder that projects heterogeneous data into a unified $d$-dimensional latent space $\mathbb{R}^d$.
    \item $\Phi$ is the update operator that governs the transition of information between new information in $\mathcal{O}$ and $\mathcal{M}$, including memory consolidation and eviction.
    \item $\mathcal{R}$ is the retrieval operator used to identify and surface relevant historical information from $\mathcal{M}$ based on current needs.
\end{itemize}

\paragraph{Memory Construction and Maintenance.}
The memory evolution occurs continuously as the agent ingests the observation stream $\mathcal{O}$ from the conversation. At each time step $t$, the update operator $\Phi$ consolidates new observations into the existing memory repository:
\begin{equation}
    \mathcal{M}_{t+1} = \Phi(\mathcal{M}_t, o_t, \pi_{evo}),
\end{equation}
where $\pi_{evo}$ represents an autonomous evolution policy, such as memory add, merge, or delete~\cite{yan2025memory,xiong2025memory}. 

\paragraph{Memory Retrieval.} Generation is invoked at a specific task timestamp $\tau$. Given a multimodal user query $q_\tau$, the retrieval operator $\mathcal{R}$ surfaces a relevant subset $\mathcal{M}_{ret}$ from the current state of the memory repository:
\begin{equation}
\resizebox{\linewidth}{!}{%
$
    \mathcal{M}_{ret} = \{ m_i \in \mathcal{M}_\tau \mid \text{rank}(\text{sim}(f_{\theta}(q_\tau), f_{\theta}(m_i))) \leq K \},
$
}
\end{equation}
where $\text{sim}(\cdot)$ is a similarity scoring function which computes the multimodal similarity and $K$ is the retrieval size.

\paragraph{Memory-Augmented Generation.}
The final action or response $y_\tau$ is generated by the MLLM agent by conditioning on the fusion of the working context $\mathbf{C}_\tau$, retrieved external knowledge $\mathcal{M}_{ret}$, and the specific task query $q_\tau$ at timestamp $\tau$:
\begin{equation}
    y_\tau = \text{MLLM}(\mathbf{C}_\tau \oplus \mathcal{M}_{ret} \oplus q_\tau),
\end{equation}
where $\oplus$ denotes the integration of heterogeneous tokens. This architecture ensures that the agent can maintain long-term coherence and persona consistency without being limited by the context window of the underlying model~\cite{zhang2025survey}.

\begin{table*}[t]
\centering
\caption{Detailed statistics of the new dataset in our \model. Each multi-session conversation scenario contains multiple related dialogue topics.}
 \resizebox{\linewidth}{!}{%
\label{tab:per_file_stats}
\begin{tabular}{l|ccc|cc}
\hline
\textbf{Scenarios} & \textbf{Sessions} & \textbf{Rounds} & \textbf{Images in Dialogs} & \textbf{QAs} & \textbf{Images in Questions} \\ \hline
AI, Robotics, Automation, Future Tech & 12 & 185 & 31 & 57 & 8 \\
Academic, Animal, Pet, Research, Life & 14 & 177 & 45 & 77 & 28 \\
Architecture, Art, Culture, Exhibition, Technology & 14 & 186 & 30 & 74 & 6 \\
Astronomy, Physics, Scientific Experiments, Cosmology & 11 & 203 & 30 & 54 & 2 \\
Baking, Dessert, Daily Life, Skill & 15 & 262 & 57 & 111 & 32 \\
Dog, Behavior, Research, Academic, Life & 11 & 180 & 38 & 63 & 21 \\
Education, Career, Research, Lifestyle & 12 & 227 & 56 & 106 & 21 \\
Entrepreneurship, Blockchain, Economics, Logistics, Nature & 10 & 144 & 36 & 90 & 25 \\
Fashion, Personal Care, Lifestyle, Shopping & 9 & 168 & 37 & 61 & 13 \\
Global Travel, Culture, Sightseeing & 10 & 176 & 66 & 87 & 29 \\
Global Travel, Sustainable Fashion, Design & 9 & 159 & 53 & 78 & 29 \\
Home, Health, Lifestyle, Product & 10 & 163 & 36 & 74 & 8 \\
Home, Repair, Maintenance, Cleaning & 18 & 270 & 43 & 68 & 0 \\
Landscape, Travel, Architecture, Nature & 11 & 173 & 64 & 92 & 38 \\
Music, Dance, Theater, Performance, Learning & 14 & 220 & 43 & 57 & 1 \\
Nature, Economics, Programming, Student Life & 10 & 201 & 78 & 136 & 67 \\
Parenting, Commuting, Hobbies, Travel, Gear & 16 & 221 & 48 & 82 & 14 \\
Python, Botany, AI, Student Life & 12 & 252 & 117 & 205 & 110 \\
Real Estate, Home Decor, DIY, Lifestyle & 11 & 196 & 53 & 75 & 11 \\
Technology, Ethics, Future Society & 11 & 199 & 42 & 64 & 24 \\
\hline
\textbf{Total Statistics} & \textbf{240} & \textbf{3,962} & \textbf{1,003} & \textbf{1,711} & \textbf{487} \\ \hline
\end{tabular}
}
\end{table*}

\subsection{Data Construction Details and Statistics}\label{sec:data_detail_appendix}
\subsubsection{Data Statistics}
\model is a new benchmark featuring: (1) scenario diversity, with long, multimodal, and multi-session conversations spanning twenty scenarios; (2) non-trivial multimodality, where visual and textual information are genuinely complementary and both important; and (3) multi-faceted capabilities, comprehensively assessing key memory abilities including extraction, test-time adaptation, reasoning, and knowledge management. The dataset in \model is constructed across 20 diverse conversation scenarios, comprising a total of 240 multi-session dialogues with 3,962 conversational rounds. It incorporates 1,003 input images, which are naturally grounded within the dialogue context. The details of each scenario can be found in Table~\ref{tab:per_file_stats}. Based on these dialogues, we further curate 1,711 human-annotated question–answer pairs to support systematic evaluation. Among them, 487 questions are explicitly associated with visual inputs, requiring models to reason over both textual and visual information. The details can be found in Table~\ref {tab:eval_task_stat}. This design ensures broad coverage of multimodal conversational settings while enabling fine-grained analysis of long-term multimodal memory capabilities.

\subsubsection{Data Source}\label{sec:data_source}
\textbf{Image Data Source}. To ensure sufficient evaluation difficulty and diversity, 
we collect fine-grained image categories as well as images that are closely related to the manually designed dialogue topics as multimodal conversation construction materials. All collected images are selected from sources with permissive knowledge-sharing licenses. These image resources include 
MMDU\footnote{\url{https://github.com/Liuziyu77/MMDU}}~\cite{liu2024mmdu}, CUB-200-2011\footnote{\url{https://www.vision.caltech.edu/datasets/cub_200_2011}}~\cite{wah2011caltech}, Stanford Dogs\footnote{\url{http://vision.stanford.edu/aditya86/ImageNetDogs}}~\cite{khosla2011novel}, Oxford flower\footnote{\url{https://www.robots.ox.ac.uk/~vgg/data/flowers/102}}~\cite{nilsback2008automated}, and DeepFashion\footnote{\url{https://mmlab.ie.cuhk.edu.hk/projects/DeepFashion.html}}~\cite{liu2016deepfashion}.
After collection, we manually select appropriate images and insert them at information-relevant positions within the generated textual conversations. The details can be found in Appendix~\ref{sec:conv_data_gen_appendix}.

\textbf{Single-Session Dialogue Data Source}. In addition to newly created materials, we reconstruct existing single-session dialogues with multimodal information, sourced from prior dataset MMRC\footnote{\url{https://github.com/haochen-MBZUAI/MMRC}}~\cite{xue2025mmrc}. Specifically, we first employ advanced LLMs, namely GPT-5.1 and Gemini-2.5-Pro, to extract representative keywords from each single-session dialogue, after which the higher-quality outputs are manually selected. Based on the extracted keywords, we further use LLMs to perform session-level clustering. Once candidate multi-session dialogues are formed, we conduct additional human filtering and enhancement to ensure the overall quality and coherence of the reconstructed multi-session conversations. The details can be found in Appendix~\ref{sec:conv_data_gen_appendix}.

\begin{figure*}[t!]
\centering
\includegraphics[width=\linewidth]{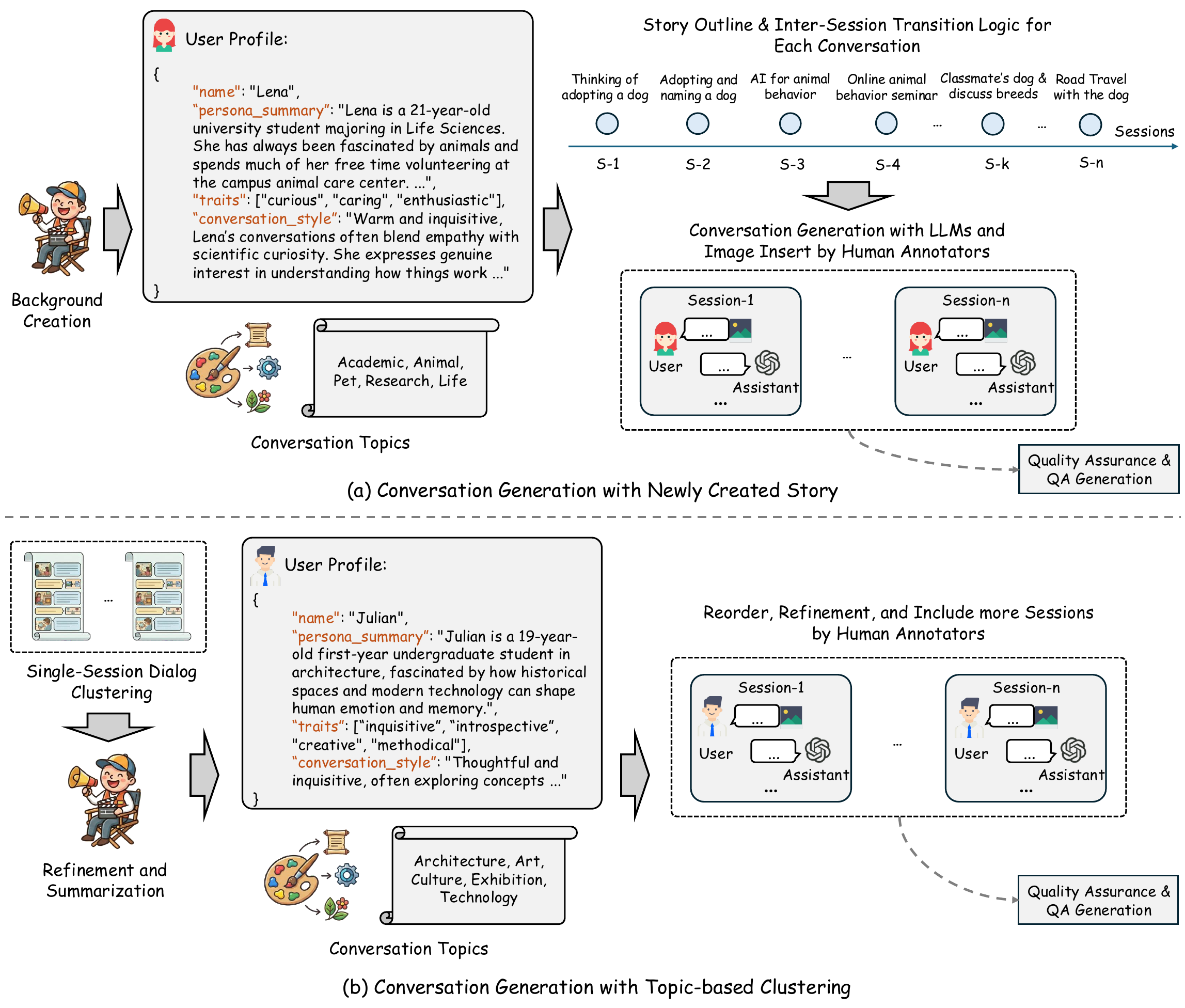}
\caption{Illustration of the generation for each conversation, including two ways.}
\label{fig:conv_gen}
\end{figure*}

\subsubsection{Conversation Data Synthesis}\label{sec:conv_data_gen_appendix}

As shown in Figure~\ref{fig:conv_gen}, our conversation data synthesis follows two main approaches.

First, we generate high-quality conversational stories with the assistance of human annotators. Annotators initially create the protagonist’s profile and overall conversation themes, and then specify the core content of each session through story outlines, as well as the transition logic between sessions. Based on this setup, dialogue generation proceeds in two stages. LLMs are first used to generate the textual dialogue content. Specifically, we employ GPT-5.1 and Gemini-2.5-Pro in parallel and manually select higher-quality outputs. The dialogues are then further refined, and appropriate images are inserted at suitable positions to establish multimodal dependencies. Each multimodal multi-session long conversation is constructed following this process, after which quality assurance and QA generation are conducted.

Second, we reconstruct existing single-session dialogue datasets via topic-based clustering. In this approach, LLMs are first used to extract session-level topic keywords and perform clustering. Human annotators then review and refine the clustered results, and summarize the corresponding user profiles and multi-session conversation topics. Since the resulting clustered long conversations often suffer from limited fluency, weak inter-session transition logic, and insufficient length, we further refine the dialogue content to ensure coherence and reasonable session transitions. To address insufficient conversation length, we additionally extend these clustered conversations by introducing more sessions, which are generated by LLMs and then manually verified and augmented with images, while preserving logical continuity across sessions.

\begin{figure}[t!]
\centering
\includegraphics[width=\linewidth]{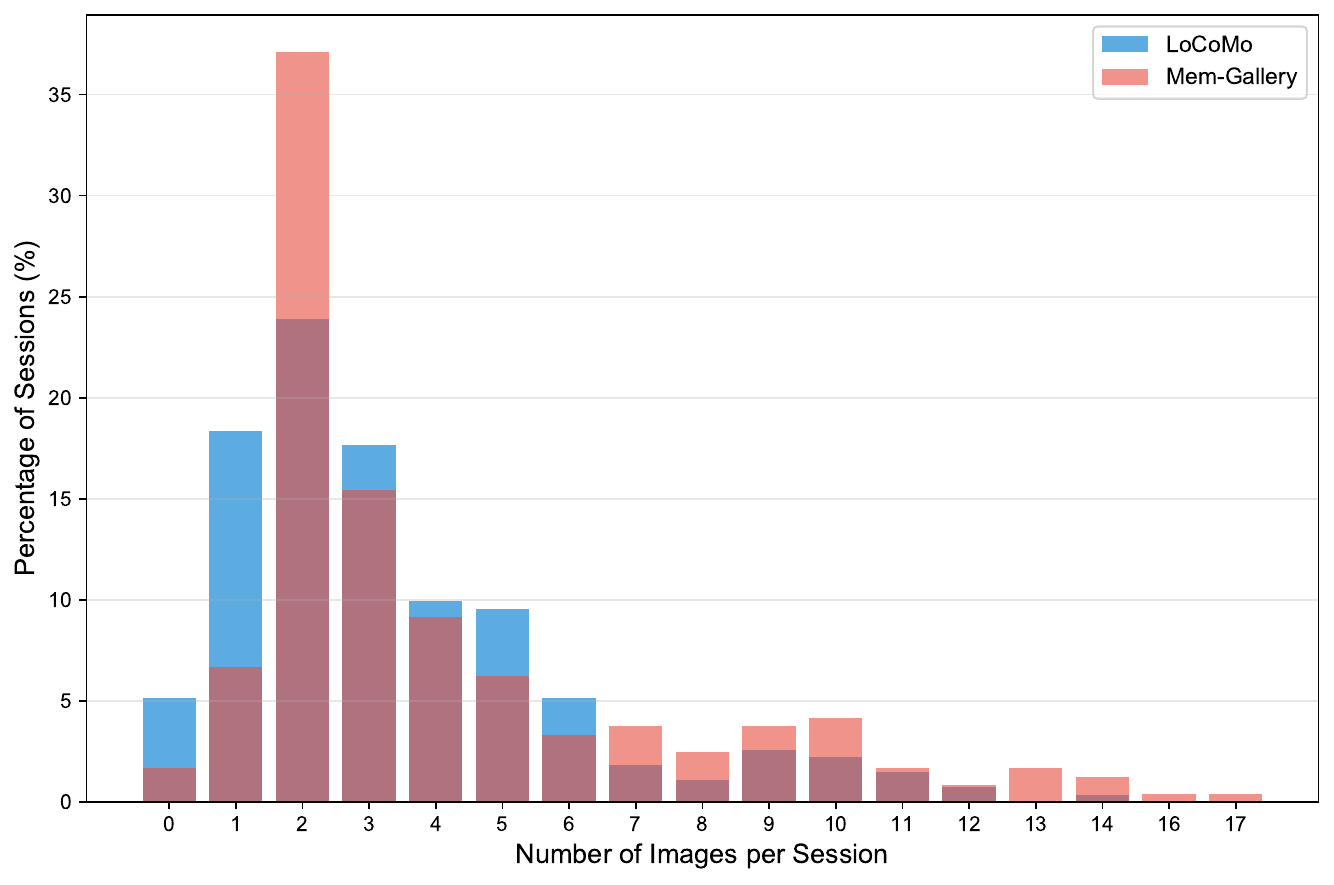}
\caption{Image number distribution of \model per conversation session compared with LoCoMo.}
\label{fig:image_dist}
\end{figure}

\begin{figure*}[t!]
\centering
\includegraphics[width=0.9\linewidth]{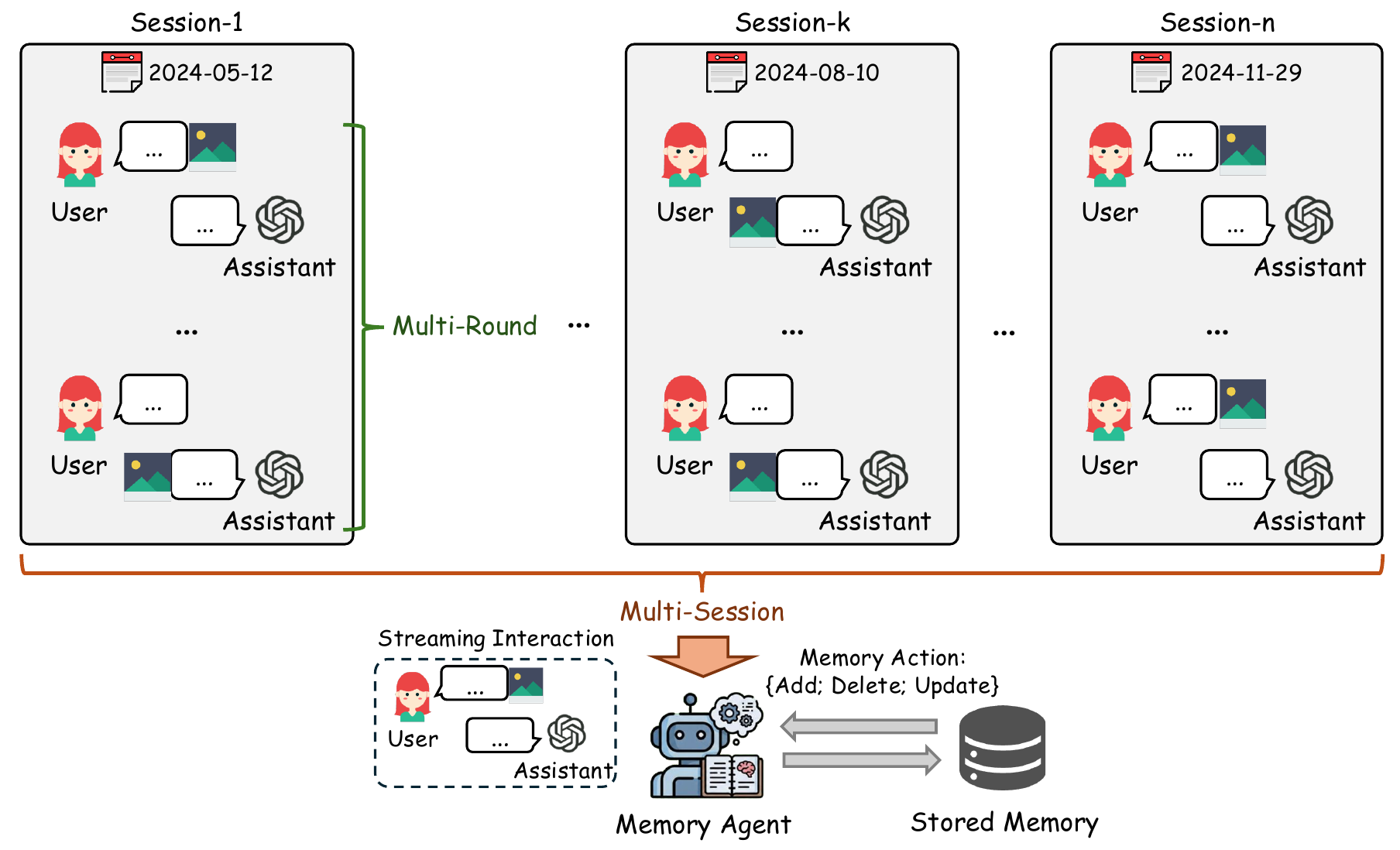}
\caption{Structure example of each conversation.}
\label{fig:conv_struct}
\end{figure*}

The distribution of image numbers per session in \model is described in Figure~\ref{fig:image_dist}. From the distribution, we observe that, beyond the importance and high quality of visual information in \model as compared in Figure~\ref{fig:comp_locomo}, \model significantly reduces the proportion of sessions with no images or only a single image compared to LoCoMo~\cite{maharana2024evaluating}. The overall distribution shifts rightward, with most sessions containing at least two images, and the number of sessions with multiple images (e.g., seven or more images) is substantially increased. In the most extreme cases, a single session contains up to 17 images. This design increases both the density of visual information in conversations and the overall difficulty of the dialogue tasks.
The conversation structure of \model and the information feed way for memory agents are shown in Figure~\ref{fig:conv_struct}.

\begin{table}[t!]
  \centering
  \caption{Evaluation task statistics on the QA number and average clues in each category. * The AR task is designed to evaluate refusal capability. The information in the question does not exist in the dialogue, and therefore clues are unsupported.}
  \resizebox{\linewidth}{!}{%
    \begin{tabular}{cc|cc}
    \toprule
    \multicolumn{2}{c|}{Evaluation Tasks} & Number of QAs & Avg. Clues \\
    \midrule
    \multirow{3}[2]{*}{Extract. \& Adapt.} & FR   & 219   & 2.90 \\
          & VS    & 306   & 1.91 \\
          & TTL  & 337   & 4.09 \\
    \midrule
    \multirow{3}[2]{*}{Reasoning} & TR    & 123   & 1.74 \\
          & VR    & 174   & 3.05 \\
          & MR    & 206   & 2.17 \\
        \midrule
    \multirow{3}[2]{*}{Knowledge Management} & KR    & 81    & 2.21 \\
          & CD    & 81    & 1.70 \\
          & AR*    & 184   &  --- \\
    \midrule
    \multicolumn{2}{c|}{Overall} & 1,711  & 2.69 \\
    \bottomrule
    \end{tabular}%
    }
  \label{tab:eval_task_stat}%
\end{table}%

\subsubsection{Evaluation Data Synthesis}\label{sec:eval_data_gen_appendix}
Evaluation QA pairs are constructed through two complementary and independent generation pipelines, which jointly contribute to the final evaluation set. These two pipelines are designed to balance coverage, difficulty, and diversity while ensuring that evaluation instances genuinely cover long-term and multimodal memory access.

In the first pipeline, advanced LLMs are prompted to automatically generate candidate QA pairs. The input to the LLM consists of selected multi-session conversation histories together with detailed task descriptions specifying the targeted evaluation capability (e.g., factual retrieval, temporal reasoning, or knowledge resolution). The prompts explicitly constrain questions to depend on information distributed across multiple dialogue turns or sessions, and, when applicable, to require visual evidence in addition to textual context. This pipeline enables scalable generation of QA candidates with broad task coverage.

In parallel, human annotators manually construct QA pairs by reviewing each entire multi-session conversation from start to end. Annotators are instructed to identify information dependencies that span long temporal gaps, involve updates or contradictions, or require integrating visual and textual cues. Based on these observations, annotators design targeted questions that cannot be answered from local context alone and instead require accessing accumulated multimodal memory. 
Notably, LLMs are unable to reliably generate QA pairs that explicitly depend on visual evidence. The resulting questions are often inaccurate or of low quality. Therefore, evaluation instances for tasks that require explicit visual cues are entirely constructed by human annotators.
This process emphasizes challenging cases that are difficult for automatic generation to reliably capture.

For each finalized QA pair, annotators explicitly annotate the evidence clues required to derive the correct answer with the assistance of LLMs. These clues specify the dialogue turns or visual content where the relevant information is introduced or referenced.

\subsection{Detailed Evaluation Task Descriptions}\label{sec:task_desc_appendix}

\name evaluates multimodal long-term conversational memory through three complementary functional dimensions, each consisting of three fine-grained task types. Together, these nine tasks provide a systematic and comprehensive assessment of how memory is extracted, adapted, reasoned over, and managed under long-horizon multimodal conversational settings:

\paragraph{Memory Extraction \& Adaptation.}
This dimension focuses on whether an agent can correctly extract, store, and adaptively utilize information accumulated throughout long-term multimodal interactions.

\begin{itemize}
    \item \textbf{Factual Retrieval (FR)}:  
    This task evaluates the model’s ability to accurately recall explicit factual information mentioned in previous multimodal dialogue turns, including user attributes, preferences, entities, events, and decisions. Questions may require retrieving information introduced long before the query, testing whether the memory system can preserve and access long-horizon facts rather than relying on short-term context.

    \item \textbf{Visual-centric Search (VS)}:  
    This task assesses whether the model can correctly and explicitly identify or retrieve relevant visual content from the conversation history. Queries typically require selecting the correct image or images associated with a specific entity, event, or user reference, testing the alignment between visual memory entries and textual cues in long-term memory.

    \item \textbf{Test-Time Learning (TTL)}:  
    This task measures the model’s ability to adapt its understanding at inference time based on newly provided multimodal examples. The agent must incorporate new multimodal instances into its existing memory and apply this updated knowledge to answer subsequent questions, without any parameter updates. This setting emphasizes online memory adaptation rather than static memorization.
\end{itemize}

\paragraph{Memory Reasoning.}
Beyond direct retrieval, real-world conversational agents must reason over stored memories. This dimension evaluates whether models can perform structured reasoning using long-term multimodal information accumulated over time.

\begin{itemize}
    \item \textbf{Temporal Reasoning (TR)}:  
    This task evaluates whether the model can reason over temporally ordered information in memory. Questions may involve comparing events across different sessions, identifying the order of occurrences, or determining when a particular event happened based on dispersed multimodal cues.

    \item \textbf{Visual-centric Reasoning (VR)}:  
    This task tests the model’s ability to use visual information as a key component of the reasoning process. Unlike visual-centric search, which focuses on retrieval, visual-centric reasoning requires integrating visual cues with textual context to infer properties, relationships, or similarities that are not explicitly stated.

    \item \textbf{Multi-entity Reasoning (MR)}:  
    This task assesses the model’s capacity to jointly reason over multiple entities stored in memory, where entities may be introduced at different times and may involve both textual and visual representations. Correct answers require synthesizing information across entities rather than recalling isolated facts.
\end{itemize}

\paragraph{Memory Knowledge Management.}
Conversational memory is inherently dynamic and imperfect. This dimension evaluates whether a memory system can properly manage evolving, conflicting, or missing information.

\begin{itemize}
    \item \textbf{Knowledge Resolution (KR)}:  
    This task examines whether the model can correctly update its memory when new information contradicts previously stored knowledge. The agent must discard or revise outdated beliefs and consistently rely on the most recent and correct information when answering questions.

    \item \textbf{Conflict Detection (CD)}:  
    This task evaluates the model’s ability to detect inconsistencies between newly observed information and existing memory. Rather than resolving the conflict, the agent is required to explicitly identify whether a contradiction exists, testing sensitivity to memory consistency.

    \item \textbf{Answer Refusal (AR)}:  
    This task assesses whether the model can appropriately refuse to answer when the requested information has never been mentioned or is unsupported by the conversation history. It tests whether the memory system avoids hallucination and recognizes the absence of valid evidence in long-term memory.
\end{itemize}

\subsection{Baseline Details}\label{sec:model_detail_appendix}

We include thirteen memory methods, with eight representative textual memory methods and five multimodal memory methods.

\subsubsection{Textual Memory with Visual Caption}
For the textual memory, we transfer the raw image into an image caption with GPT-5.1. The details of these models are illustrated as follows.

\begin{itemize}
    \item \textbf{Full Memory (Text)}: It includes all memory information in textual form as part of the context, and truncates it according to the context token limit.
    \item \textbf{FIFO} (First-in-first-out): It selects the most recent memory information as context according to the temporal order of the memory.
    \item \textbf{NaiveRAG}: It encodes the memory information into semantic vectors, then retrieves and returns the most similar memory entries based on vector similarity.
    \item \textbf{Generative Agent}~\cite{park2023generative}: It introduces a generative agent memory that stores experiences as language memories and retrieves relevant entries to condition agent behavior. It further synthesizes higher-level reflections from past experiences to support coherent long-term planning and interaction.
    \item \textbf{Reflexion}~\cite{shinn2023reflexion}: Reflexion proposes a verbal memory mechanism in which agents store self-generated reflections as episodic memories. After each trial, feedback is converted into language reflections that are appended to memory and reused as context to guide decisions, enabling trial-and-error learning via memory accumulation.
    \item \textbf{MemGPT}~\cite{packer2023memgpt}: MemGPT proposes an OS-inspired memory architecture that treats the LLM context window as limited working memory and manages long-term information via hierarchical external storage. Through self-directed function calls, the model dynamically pages relevant memories into context and evicts less useful ones, enabling effective long-term conversation beyond fixed context limits.
    \item \textbf{A-Mem}~\cite{xu2025mem}: A-MEM proposes an agentic memory system that organizes experiences as structured atomic notes with rich contextual descriptions, inspired by the Zettelkasten method. New memories autonomously establish links to related past memories and can trigger updates to existing ones, enabling a dynamically evolving memory network that supports long-term reasoning without predefined memory operations.
    \item \textbf{MemoryOS}~\cite{kang2025memory}: MemoryOS introduces an operating system–inspired memory framework that manages agent memory through hierarchical storage, dynamic updating, and semantic retrieval. It organizes memory into short-term, mid-term, and long-term persona layers, and applies OS-style mechanisms such as FIFO, segmented paging, and heat-based eviction to maintain coherent and personalized long-term conversations.
\end{itemize}

\subsubsection{Multimodal Memory}
For the multimodal memory, we include five representative models. The details of these models are illustrated as follows.

\begin{itemize}
    \item \textbf{Full Memory (Multimodal)}: It includes all multimodal memory information as context, estimates the token consumption of images using predefined token costs, and truncates the input according to the context token limit.
    \item \textbf{MuRAG}~\cite{chen2022murag}: MuRAG uses a dense multimodal retriever that encodes both queries and memory entries into a shared embedding space using a joint vision–language encoder. At inference time, it performs maximum inner product search over an external memory to retrieve the most relevant items, which are then used to augment generation. We use the retrieval paradigm in this paper for our benchmark implementation.
    \item \textbf{UniversalRAG}~\cite{yeo2025universalrag}: It introduces a modality- and granularity-aware RAG framework that dynamically routes each query to the most appropriate knowledge source before retrieval. Instead of retrieving from a unified corpus, it first predicts the required modality and granularity and then performs targeted retrieval within the selected corpus, reducing modality bias and retrieval noise.
    \item \textbf{NGM}~\cite{fisher2025neural}: It proposes Neural Graph Memory, a graph-structured multimodal memory that stores episodic experiences as nodes enriched with modality-specific embeddings and temporal metadata. Memories are retrieved via graph traversal, enabling associative, temporal, and cross-modal recall beyond flat vector similarity, and supporting long-horizon episodic reasoning.
    \item \textbf{AUGUSTUS}~\cite{jain2025augustus}: AUGUSTUS introduces a cognition-inspired multimodal memory that organizes long-term memory into recall memory for raw conversation history and a hierarchical contextual memory for semantic concepts linked to multimodal context. Abstracting user interactions into semantic tags and retrieving information through a concept-driven search enables efficient, personalized, and long-horizon multimodal memory access beyond flat vector databases.
\end{itemize}

The corresponding papers and implementation details of these methods are summarized in Table~\ref{tab:imple_source}. For methods without publicly available implementation code (marked as N/A), we re-implement them based on the methodological descriptions provided in the original papers if available.

\begin{table}[htbp]
  \centering
  \caption{The detailed resource list of models implemented in our benchmark.}
  \resizebox{\linewidth}{!}{%
    \begin{tabular}{l|ccc}
    \toprule
    Model & Venue & Paper & Implement. Source \\
    \midrule
    Full Memory-Text & N/A   & N/A   & \href{https://github.com/nuster1128/MemEngine/blob/master/memengine/memory/FUMemory.py}{Link} \\
    FIFO  & N/A   & N/A   & \href{https://github.com/nuster1128/MemEngine/blob/master/memengine/memory/STMemory.py}{Link} \\
    NaiveRAG & N/A   & N/A   & \href{https://github.com/nuster1128/MemEngine/blob/master/memengine/memory/LTMemory.py}{Link} \\
    Gen. Agent & UIST 2023 & \href{https://dl.acm.org/doi/pdf/10.1145/3586183.3606763}{Link} & \href{https://github.com/nuster1128/MemEngine/blob/master/memengine/memory/GAMemory.py}{Link} \\
    Reflexion & NeurIPS 2023 & \href{https://proceedings.neurips.cc/paper\_files/paper/2023/file/1b44b878bb782e6954cd888628510e90-Paper-Conference.pdf}{Link} & \href{https://github.com/nuster1128/MemEngine/blob/master/memengine/memory/RFMemory.py}{Link} \\
    MemGPT & Arxiv 2023 & \href{https://par.nsf.gov/servlets/purl/10524107}{Link} & \href{https://github.com/nuster1128/MemEngine/blob/master/memengine/memory/MGMemory.py}{Link} \\
    A-Mem & NeurIPS 2025 & \href{https://arxiv.org/pdf/2502.12110}{Link} & \href{https://github.com/WujiangXu/A-mem}{Link} \\
    MemoryOS & EMNLP 2025 & \href{https://aclanthology.org/2025.emnlp-main.1318.pdf}{Link} & \href{https://github.com/BAI-LAB/MemoryOS}{Link} \\
    Full Memory-Multimodal & N/A   & N/A   & N/A \\
    MuRAG & EMNLP 2022 & \href{https://aclanthology.org/2022.emnlp-main.375.pdf}{Link} & N/A \\
    UniversalRAG & Arxiv 2025 & \href{https://arxiv.org/pdf/2504.20734}{Link} & \href{https://github.com/wgcyeo/UniversalRAG}{Link} \\
    NGM   & Arxiv 2025 & \href{https://www.researchgate.net/profile/Matt-Fisher-7/publication/394440420_Neural_Graph_Memory_A_Structured_Approach_to_Long-Term_Memory_in_Multimodal_Agents/links/689ab8c337b271210509c20f/Neural-Graph-Memory-A-Structured-Approach-to-Long-Term-Memory-in-Multimodal-Agents.pdf}{Link} & \href{https://github.com/StuckInTheNet/Neural-Graph-Memory-NGM}{Link} \\
    AUGUSTUS & Arxiv 2025 & \href{https://arxiv.org/pdf/2510.15261}{Link} & N/A \\
    \bottomrule
    \end{tabular}%
    }
  \label{tab:imple_source}%
\end{table}%

\subsection{Benchmark Evaluation Details}\label{sec:bench_eval_details}
\subsubsection{Experimental Environments}
For open-sourced MLLM models, we obtain the official models from their corresponding Hugging Face repositories\footnote{\url{https://huggingface.co}}.
Then, we deploy them on A100 GPUs under the Linux system with the vLLM engine\footnote{\url{https://github.com/vllm-project/vllm}}. For closed-sourced MLLM models, we access them via official APIs. The memory model implementation and reproduction are built based on the open-sourced memory framework MemEngine\footnote{\url{https://github.com/nuster1128/MemEngine}}~\cite{zhang2025memengine}. 
The total API compute budget for open-sourced MLLM models was under one thousand US dollars. The detailed running time for each memory model is illustrated in Figure~\ref{fig:efficiency}.

\subsubsection{Benchmark Settings}\label{sec:bench_setup}
To eliminate the influence of backbone differences and minimize randomness, all experimental settings use the same MLLM backbone, with a fixed random seed and the temperature set to 0.
To further ensure a fair comparison, for memory systems that involve embedding-based storage and retrieval, we uniformly adopt the widely used GME-Qwen2-VL-2B-Instruct~\cite{zhang2024gme} as the embedding model.
For memory methods that require a retriever, we adopt a default retrieval size of $K$=10. For other hyperparameters of the models, we retain their optimal settings from the original codebase.

\subsubsection{Evaluation Pipeline}
The overall evaluation pipeline is based on the  MemEngine~\cite{zhang2025memengine} framework. The setup is illustrated in Appendix~\ref{sec:bench_setup}.
For baseline methods that involve calling an MLLM module, we consistently use the corresponding backbone MLLM of each method to ensure a fair comparison.
Both the conversational data and the evaluation data are stored in JSON format.
For conversational data, following existing works, dialogue information is streamed to each memory method at the granularity of dialogue rounds.
During question-answering evaluation, the accumulated memory contents are concatenated into the context, after which the corresponding MLLM is invoked to generate the final answer and save the output.
For tasks such as visual-centric search, conflict detection, and answer refusal, which require specific output formats, additional format-constrained prompts are appended after the question, as shown in Figure~\ref{fig:format_prompt}.
Finally, all generated answers are normalized in a unified manner before metric computation.
The prompt template used for LLM-as-a-Judge evaluation is illustrated in Figure~\ref{fig:llm_judge_prompt}.

\begin{figure*}[t] 
    \centering
\begin{tcolorbox}[
    enhanced,
    colframe=black,
    boxrule=0.5pt,
    title={\textcolor{white}{Format Restriction Prompts}},
    coltitle=white,
    fonttitle=\bfseries,
    attach boxed title to top left={xshift=2mm, yshift=-2mm},
    boxed title style={
        colback=black,
        sharp corners
    }
]
\vspace{2mm}
\textbf{Visual-centric Search:}

Return the image\_id of the image(s). If there are multiple images, sort them in ascending order and separate them by commas. Format example: “D2:IMG\_003, D2:IMG\_010, D10:IMG\_002” (for format reference only).

\vspace{2mm}
\textbf{Conflict Detection:}

Please check whether this information conflicts with the conversation, and reply strictly with either “Yes.” or “No.”

\vspace{2mm}
\textbf{Answer Refusal:}

Provide your answer based on the information in the conversation. Only if the information about the question is not present in the conversation, reply with: “Not mentioned.”

\end{tcolorbox}
\caption{Prompts for format restriction tasks.} 
    \label{fig:format_prompt}
\end{figure*}

\begin{figure*}[t!] 
    \centering
\begin{tcolorbox}[
    enhanced,
    colframe=black,
    boxrule=0.5pt,
    title={\textcolor{white}{LLM-as-a-Judge Prompt}},
    coltitle=white,
    fonttitle=\bfseries,
    attach boxed title to top left={xshift=2mm, yshift=-2mm},
    boxed title style={
        colback=black,
        sharp corners
    }
]
\vspace{2mm}

You are an impartial judge evaluating the memory capabilities of an AI assistant with the question-answering task.
Your task is to compare the Assistant's Answer against the Ground Truth and assign a score of 0, 0.25, 0.5, 0.75, or 1.

\vspace{2mm}
Scoring Rubric

\textbf{Score 0 (Incorrect / Miss):}

- The answer contradicts the Ground Truth.

- For Yes/No questions: The answer has the wrong polarity (e.g., says "Yes" when Ground Truth is "No").

- For Open-ended questions: The answer provides factually wrong information or hallucinations.

- The assistant fails to provide the required information.

\textbf{Score 0.25 (Poor / Tangential):}

- The answer touches on the topic but misses the core entity or key value required.

- The answer contains a mix of minor correct details and significant hallucinations or wrong associations.

- The answer is excessively vague to the point of being useless (e.g., answering "a dog" instead of "a golden retriever").

\textbf{Score 0.5 (Partial / Vague):}

- The answer is technically correct, but lacks confidence or is incomplete.

- The answer captures the main entity or concept correctly but misses a part of the required supporting details.

- For Yes/No questions: The polarity is correct, but the reasoning is flawed (if have), or the assistant is uncertain (e.g., "I think it might be Yes").

- For Open-ended questions: The answer is too general or misses key adjectives/details present in the Ground Truth.

\textbf{Score 0.75 (Good / Minor Imperfection):}

- The answer is largely accurate and captures the core information confidently.

- It misses only minor details (e.g., specific adjectives or secondary details) that do not alter the main truth.

- The answer contains all the correct information but includes unnecessary "fluff" or slight conversational filler that reduces precision.

\textbf{Score 1 (Correct / Exact):}

- The answer is accurate, precise, and confident.

- For Yes/No questions: The polarity matches the Ground Truth perfectly.

- For Open-ended questions: The answer contains all the core information and necessary details required by the Ground Truth without hallucinations.

\vspace{2mm}
Input Data

Question: \{{question}\}

Ground Truth: \{{ground\_truth}\}

Assistant Answer: \{{model\_output}\}

\vspace{2mm}
Output Format

Output strictly in the following JSON format:
{"score": <0, 0.25, 0.5, 0.75, or 1>, "reasoning": "<short explanation>"}

\end{tcolorbox}
\caption{The prompt of the LLM-as-a-Judge metric for Qwen-2.5-72B-Instruct.} 
    \label{fig:llm_judge_prompt}
\end{figure*}

\subsubsection{Evaluation Metrics}
\label{sec:eval_metrics}

We evaluate \model from three key perspectives: question-answering performance, retrieval effectiveness, and computational efficiency.
All metrics are computed at the instance level and then averaged over the QA set.

\paragraph{Question-Answering Metrics.}
For answer quality, we adopt four widely used metrics in conversational question answering.

\textbf{F1} measures token-level overlap between the predicted answer $A_p$ and the reference answer $A_r$.
Let $T_p$ and $T_r$ denote the multisets of tokens in $A_p$ and $A_r$, respectively.
Precision $P$ and recall $R$ are defined as:
\begin{equation}
P = \frac{|T_p \cap T_r|}{|T_p|}, \quad
R = \frac{|T_p \cap T_r|}{|T_r|}.
\end{equation}
The F1 score is then computed as:
\begin{equation}
\mathrm{F1} = \frac{2PR}{P + R}.
\end{equation}

\textbf{Exact Match (EM)} evaluates whether the predicted answer exactly matches the reference answer after normalization (e.g., lowercasing and punctuation removal):
\begin{equation}
\mathrm{EM} =
\begin{cases}
1, & \text{if } A_p = A_r, \\
0, & \text{otherwise}.
\end{cases}
\end{equation}

\textbf{BLEU-1} measures unigram-level precision between the predicted answer $A_p$ and the reference answer $A_r$.
Let $\mathrm{count}_{A_p}(w)$ denote the number of occurrences of unigram $w$ in $A_p$, and $\mathrm{count}_{A_r}(w)$ denote its occurrences in $A_r$.
The clipped count $c(w)$ is defined as:
\begin{equation}
c(w) = \min\bigl(\mathrm{count}_{A_p}(w), \mathrm{count}_{A_r}(w)\bigr).
\end{equation}
BLEU-1 is then computed as:
\begin{equation}
\mathrm{BLEU\text{-}1} =
\frac{\sum_{w \in A_p} c(w)}{\sum_{w \in A_p} \mathrm{count}_{A_p}(w)}.
\end{equation}

\textbf{LLM-as-a-Judge} evaluates semantic correctness using Qwen-2.5-72B-Instruct as an automatic evaluator.
Given the question, the ground-truth answer, and the model prediction, the judge assigns a discrete score from the set
$\{0, 0.25, 0.5, 0.75, 1\}$ according to a predefined rubric.
Specifically:
\begin{itemize}
    \item $0$: Incorrect or missing answer, including contradictions or hallucinations.
    \item $0.25$: Poor or tangential answer that touches on the topic but misses the core entity or value.
    \item $0.5$: Partially correct answer that captures the main concept but lacks completeness or key details.
    \item $0.75$: Largely correct answer with only minor omissions or imprecision.
    \item $1$: Correct and exact answer that fully matches the ground truth.
\end{itemize}
The final LLM-as-a-Judge score is obtained by averaging the assigned scores across evaluation instances.
This metric complements lexical overlap metrics by capturing semantic equivalence, paraphrasing, and partial correctness.

For the Conflict Detection (CD) task, the model is required to output a single binary token (\textit{Yes} or \textit{No}).
Since this is a binary classification task and all evaluated models reliably follow the instruction format, the values of F1, EM, BLEU-1, and LLM-as-a-Judge coincide for this task.

\paragraph{Retrieval Effectiveness Metrics.}
To assess memory access quality independently of answer generation, we adopt standard information retrieval metrics.
For each evaluation query, let $\mathcal{R}_K$ denote the set of top-$K$ retrieved memory entries, and $\mathcal{G}$ denote the set of ground-truth relevant memory entries annotated as evidence clues.

\textbf{Recall@K} is defined as:
\begin{equation}
\mathrm{Recall@}K = \frac{|\mathcal{R}_K \cap \mathcal{G}|}{|\mathcal{G}|}.
\end{equation}

\textbf{Precision@K} is defined as:
\begin{equation}
\mathrm{Precision@}K = \frac{|\mathcal{R}_K \cap \mathcal{G}|}{|\mathcal{R}_K|}.
\end{equation}

\textbf{Hit@K} measures whether at least one relevant memory entry is retrieved:
\begin{equation}
\mathrm{Hit@}K =
\begin{cases}
1, & \text{if } \mathcal{R}_K \cap \mathcal{G} \neq \emptyset, \\
0, & \text{otherwise}.
\end{cases}
\end{equation}

All retrieval metrics are first computed per query and then averaged across the evaluation set.

\paragraph{Efficiency Metrics.}
We evaluate efficiency from a system-level perspective.
For each model, we measure:
(1) \textbf{Information Memorization Time}, defined as the cumulative wall-clock time spent processing and storing conversational inputs during interaction.
(2) \textbf{Memory Retrieval and Answer Generation Time}, defined as the wall-clock time required to retrieve memory entries and generate the final response at evaluation time.
All efficiency metrics are reported in seconds.

\subsection{Additional Results and Analysis}
\subsubsection{MLLM Backbone Analysis}\label{sec:backbone}

\begin{figure}[t!]
\centering
\includegraphics[width=\linewidth]{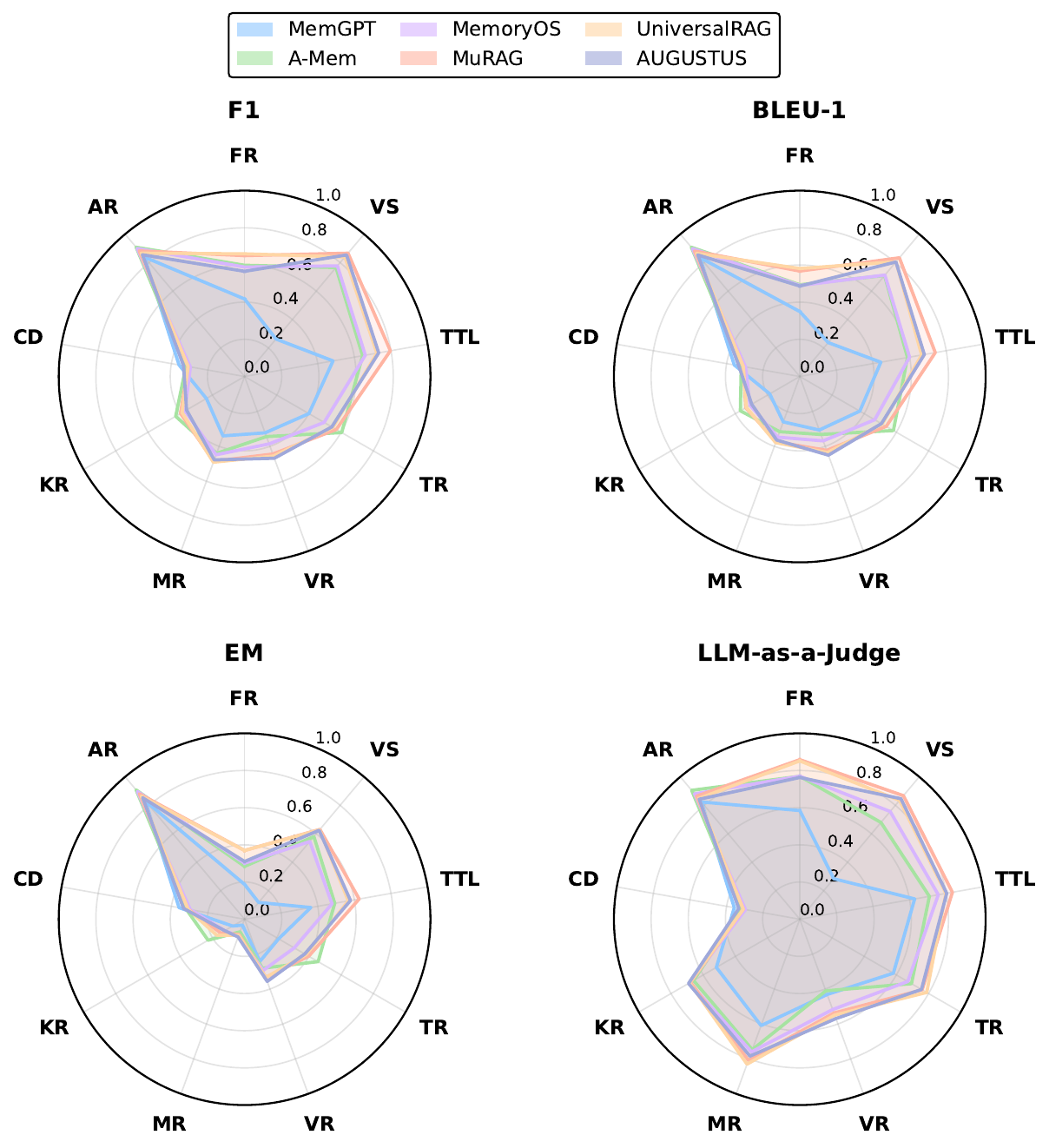}
\caption{Radar chart of subtask performance for representative memory models on Qwen-2.5-VL-3B.}
\label{fig:top_model_qwen_3b}
\end{figure}

\begin{figure}[t!]
\centering
\includegraphics[width=\linewidth]{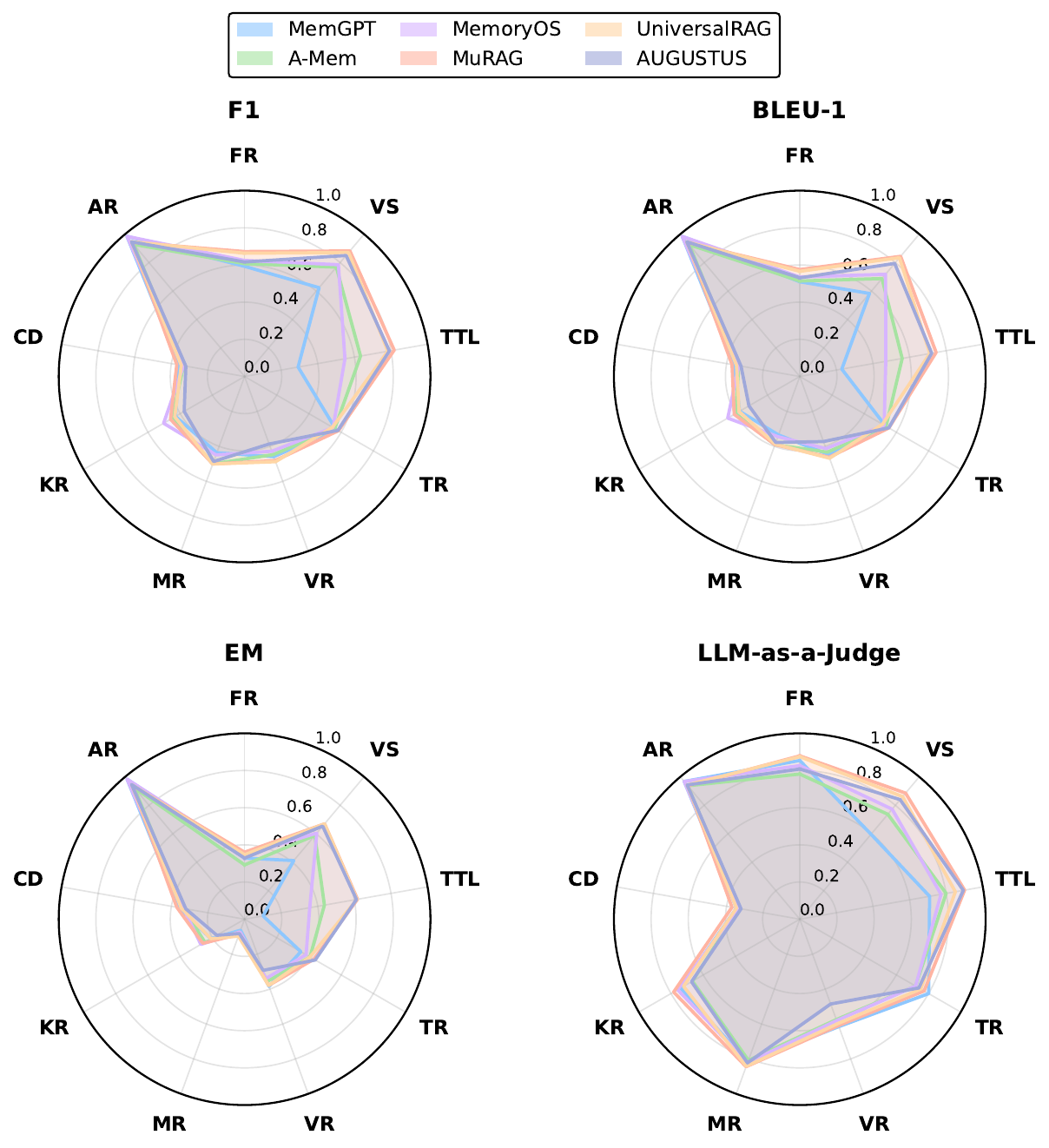}
\caption{Radar chart of subtask performance for representative memory models on Qwen-2.5-VL-7B.}
\label{fig:top_model_qwen_7b}
\end{figure}

\begin{figure}[t!]
\centering
\includegraphics[width=\linewidth]{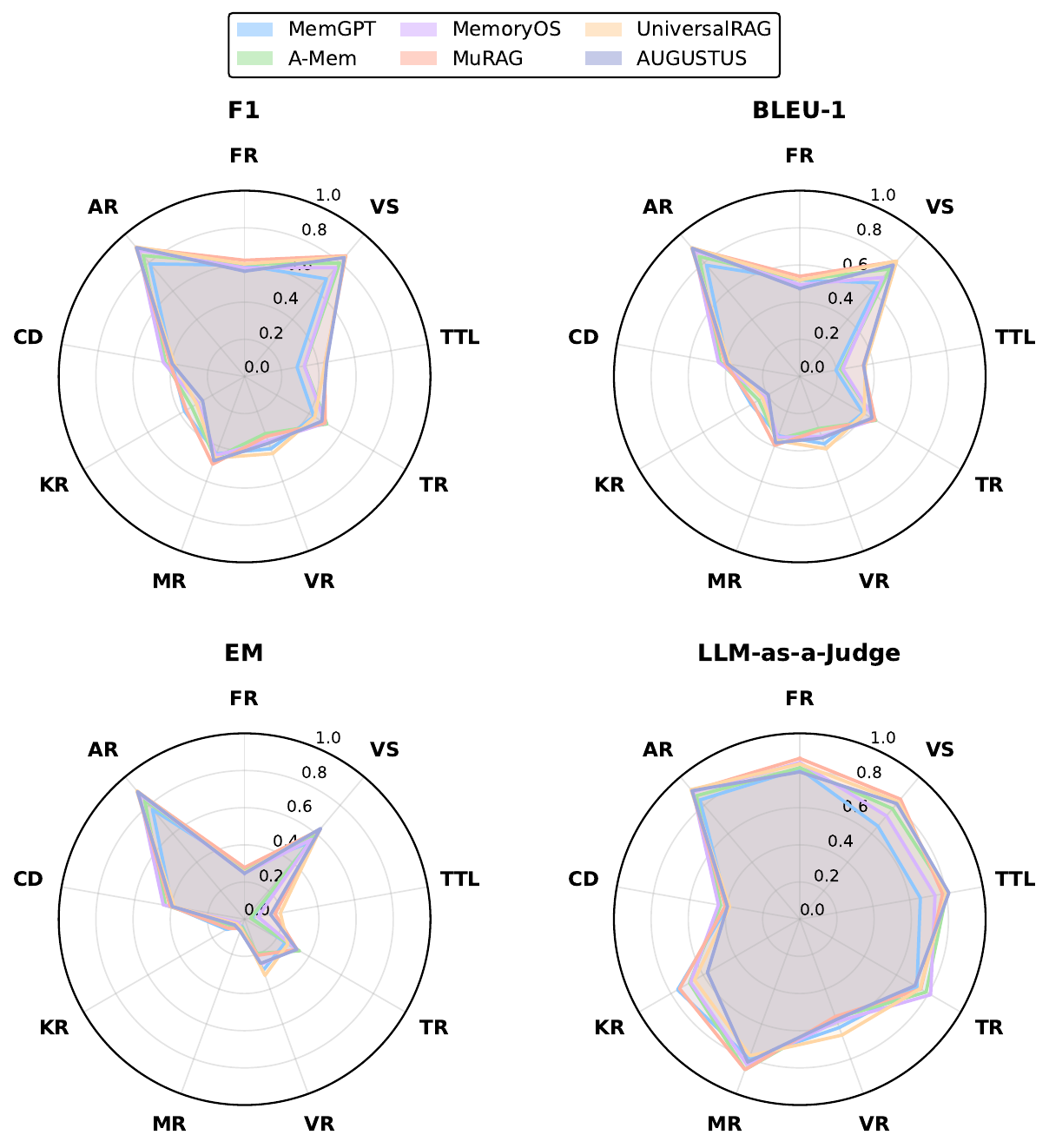}
\caption{Radar chart of subtask performance for representative memory models on GPT-4.1-Nano.}
\label{fig:top_model_gpt_nano}
\end{figure}

\begin{figure}[t!]
\centering
\includegraphics[width=\linewidth]{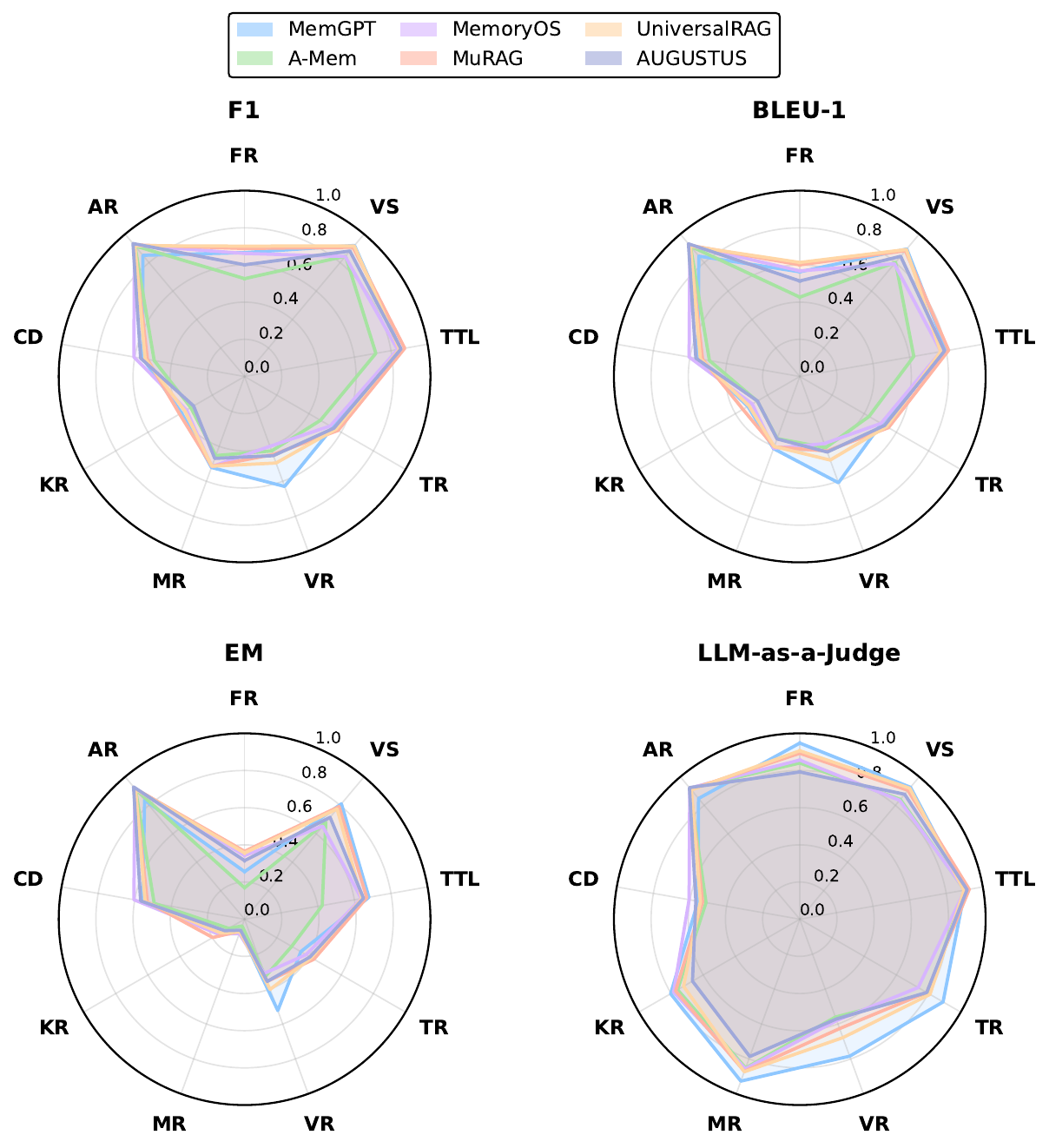}
\caption{Radar chart of subtask performance for representative memory models on Gemini-2.5-Flash-Lite.}
\label{fig:top_model_gemini}
\end{figure}

To analyze how the choice of MLLM backbone influences memory performance, we evaluate representative open-source and closed-source MLLMs under a unified memory framework. Specifically, Table~\ref{tab:result_qwen_vl_3b} and Table~\ref{tab:main_qwen_vl_7b} report results on the open-source Qwen-2.5-VL-3B and Qwen-2.5-VL-7B backbones, while Table~\ref{tab:gpt_4_1_nano} and Table~\ref{tab:gemini_flash_lite} report results on the closed-source GPT-4.1-Nano and Gemini-2.5-Flash-Lite backbones. 
Moreover, Figure~\ref{fig:top_model_qwen_3b}, Figure~\ref{fig:top_model_qwen_7b}, Figure~\ref{fig:top_model_gpt_nano}, and Figure~\ref{fig:top_model_gemini} illustrate the radar chart of three representative textual memory models (MemGPT, A-Mem, and MemoryOS) and three representative multimodal memory models (MuRAG, UniversalRAG, and AUGUSTUS) to clearly present the advantages and disadvantages of different subtasks.
Across all settings, the same set of textual and multimodal memory methods is applied, enabling a controlled comparison of backbone effects.

\textbf{Overall effect of backbone capability}. 
Across both open-source and closed-source backbones, stronger MLLMs consistently improve absolute memory performance across most tasks and methods. This trend is visible when scaling from Qwen-2.5-VL-3B to Qwen-2.5-VL-7B, and further from GPT-4.1-Nano to Gemini-2.5-Flash-Lite. The gains are systematic rather than isolated, indicating that increased backbone capacity enhances the model's ability to consume retrieved memory and generate coherent responses.
However, these improvements do not fundamentally alter which task categories remain difficult. Reasoning-intensive tasks and knowledge management tasks continue to exhibit lower performance relative to extraction-oriented tasks, even under the strongest backbones. This suggests that backbone scaling primarily improves memory utilization, rather than resolving intrinsic limitations in long-horizon reasoning and memory consistency of existing memory methods.

\textbf{Backbone scaling amplifies memory design quality rather than compensating for it}.
Despite large differences in backbone capacity and model family, the relative ranking of memory methods remains largely stable across all evaluated settings. Well-organized memory systems consistently outperform naïve or unstructured baselines under both weak and strong backbones, regardless of whether the memory is textual or multimodal. This stability indicates that backbone scaling does not compensate for poor memory organization. Instead, stronger backbones tend to amplify existing differences between memory designs. Multimodal memory benefits from stronger backbones only when cross-modal alignment and retrieval quality are sufficiently well controlled. Otherwise, strong textual memory can achieve comparable effectiveness with lower complexity, such as the equipment with Gemini-2.5-Flash-Lite.

\textbf{Reasoning remains a bottleneck despite stronger backbones}. Although stronger backbones lead to consistent gains on reasoning-oriented tasks, including temporal reasoning, visual-centric reasoning, and multi-entity reasoning, the magnitude of these gains is noticeably smaller than those observed in extraction-focused tasks. This pattern is consistent across both open-source and closed-source backbones. In several reasoning subtasks, the performance gap between textual and multimodal memory narrows as backbone capability increases, and well-designed textual memory methods remain competitive with multimodal approaches. These observations indicate that long-horizon reasoning over multimodal memory is not primarily constrained by backbone capacity, but rather by how multimodal information is structured, linked, and jointly reasoned over within the memory system. These results indicate that the organization and maintenance of multimodal memory mechanisms remain an open and promising direction for future research.

\textbf{Limited sensitivity in knowledge management tasks}.
For knowledge management tasks, including knowledge resolution, conflict detection, and answer refusal, improvements from stronger backbones are relatively limited across all evaluated MLLMs. Both open-source and closed-source backbones exhibit similar qualitative behavior, with only marginal performance differences attributable to backbone choice. This suggests diminishing returns from backbone scaling for safety- and consistency-oriented memory behaviors, which rely more heavily on explicit memory state tracking, conflict awareness, and decision policies than on raw generative or perceptual capacity.

\begin{table*}[t!]
  \centering
  \caption{Evaluation results on Qwen-2.5-VL-3B. The best and second-performed memory model(s) are highlighted with \textcolor{orange}{orange} and \textcolor{blue}{blue} backgrounds.}
  \resizebox{\linewidth}{!}{%
    \begin{tabular}{ccc|cccccccc|ccccc}
    \toprule
    \rowcolor{qwenbar} \multicolumn{3}{c|}{\qwenlogo\ Qwen-2.5-VL-3B} & Full (Text) & FIFO  & NaiveRAG & Gen. Agent & Reflexion & MemGPT & A-Mem & MemoryOS & Full (MM) & MuRAG & UniversalRAG & NGM   & AUGUSTUS \\
    \midrule
    \multirow{12}[6]{*}{\rotatebox{90}{Extract. \& Adapt.}} & \multirow{4}[2]{*}{FR} & F1    & 0.2501 & 0.1578 & 0.5664 & 0.2301 & 0.2467 & 0.4171 & 0.5985 & 0.5889 & 0.2282 & \cellcolor{TopTwo} 0.6504 & \cellcolor{TopOne} 0.6601 & 0.6045 & 0.5660 \\
          &       & BLEU-1 & 0.2000 & 0.1138 & 0.4862 & 0.1847 & 0.1976 & 0.3494 & 0.4937 & 0.4897 & 0.1795 & \cellcolor{TopTwo} 0.5673 & \cellcolor{TopOne} 0.5811 & 0.5290 & 0.4868 \\
          &       & EM    & 0.0959 & 0.0502 & 0.2740 & 0.0868 & 0.0959 & 0.1872 & 0.2843 & 0.3014 & 0.0868 & \cellcolor{TopOne} 0.3699 & \cellcolor{TopOne} 0.3699 & 0.3470 & 0.3105 \\
          &       & LLM-Judge & 0.3105 & 0.1918 & 0.7443 & 0.2922 & 0.3082 & 0.5845 & 0.7690 & 0.7694 & 0.2717 & \cellcolor{TopOne} 0.8584 & \cellcolor{TopTwo} 0.8516 & 0.7694 & 0.7626 \\
\cmidrule{2-16}     & \multirow{4}[2]{*}{VS} & F1    & 0.2687 & 0.1175 & 0.7424 & 0.3244 & 0.2691 & 0.2631 & 0.7653 & 0.7771 & 0.2154 & \cellcolor{TopOne} 0.8661 & 0.8395 & 0.8493 & \cellcolor{TopTwo} 0.8537 \\
          &       & BLEU-1 & 0.2466 & 0.1012 & 0.6934 & 0.2826 & 0.2471 & 0.2381 & 0.7096 & 0.7123 & 0.1959 & \cellcolor{TopOne} 0.8337 & 0.8056 & \cellcolor{TopTwo} 0.8066 & 0.8047 \\
          &       & EM    & 0.1569 & 0.0359 & 0.5033 & 0.1634 & 0.1601 & 0.1176 & 0.5802 & 0.5458 & 0.0980 & \cellcolor{TopTwo} 0.6307 & 0.6275 & \cellcolor{TopOne} 0.6569 & 0.6242 \\
          &       & LLM-Judge & 0.2337 & 0.0801 & 0.7092 & 0.2696 & 0.2337 & 0.2843 & 0.6794 & 0.7565 & 0.1797 & \cellcolor{TopOne} 0.8676 & 0.8350 & 0.8284 & \cellcolor{TopTwo} 0.8464 \\     
\cmidrule{2-16}      & \multirow{4}[2]{*}{TTL} & F1    & 0.4657 & 0.3640 & 0.6769 & 0.5454 & 0.4633 & 0.4833 & 0.6448 & 0.6595 & 0.4857 & \cellcolor{TopOne} 0.7966 & 0.7227 & \cellcolor{TopTwo} 0.7732 & 0.7328 \\
          &       & BLEU-1 & 0.4279 & 0.3161 & 0.6253 & 0.4845 & 0.4251 & 0.4427 & 0.5884 & 0.5997 & 0.4397 & \cellcolor{TopOne} 0.7402 & 0.6655 & \cellcolor{TopTwo} 0.7145 & 0.6803 \\
          &       & EM    & 0.3294 & 0.2344 & 0.5104 & 0.3769 & 0.3234 & 0.3620 & 0.4927 & 0.4777 & 0.3353 &\cellcolor{TopOne}  0.6261 & 0.5668 & \cellcolor{TopTwo} 0.6083 & 0.5786 \\
          &       & LLM-Judge & 0.6365 & 0.5415 & 0.7433 & 0.6706 & 0.6380 & 0.6276 & 0.7080 & 0.7567 & 0.6320 & \cellcolor{TopOne} 0.8338 & 0.7982 & \cellcolor{TopTwo} 0.8294 & 0.8027 \\    
    \midrule
    \multirow{12}[6]{*}{\rotatebox{90}{Reasoning}} & \multirow{4}[2]{*}{TR} & F1    & 0.2866 & 0.2398 & 0.4785 & 0.3288 & 0.2866 & 0.3996 & \cellcolor{TopOne} 0.6036 & 0.4942 & 0.2972 & \cellcolor{TopTwo} 0.5700 & 0.5484 & 0.5332 & 0.5422 \\
          &       & BLEU-1 & 0.2688 & 0.2226 & 0.4454 & 0.3036 & 0.2683 & 0.3728 & \cellcolor{TopOne} 0.5817 & 0.4654 & 0.2807 & \cellcolor{TopTwo} 0.5369 & 0.5119 & 0.4986 & 0.5072 \\
          &       & EM    & 0.1301 & 0.0976 & 0.3089 & 0.1626 & 0.1301 & 0.2114 & \cellcolor{TopOne} 0.4569 & 0.3089 & 0.1382 & \cellcolor{TopTwo} 0.3984 & 0.3821 & 0.3659 & 0.3740 \\
          &       & LLM-Judge & 0.3252 & 0.2480 & 0.6667 & 0.3821 & 0.3211 & 0.5813 & 0.6940 & 0.6707 & 0.3130 &\cellcolor{TopTwo} 0.7602 & \cellcolor{TopOne} 0.7886 & 0.6951 & 0.7561 \\
\cmidrule{2-16}          & \multirow{4}[2]{*}{VR} & F1    & 0.2545 & 0.1532 & 0.2966 & 0.1953 & 0.2552 & 0.3226 & 0.3441 & 0.3860 & 0.2285 & 0.4440 & \cellcolor{TopTwo} 0.4554 & 0.4260 & \cellcolor{TopOne} 0.4681 \\
          &       & BLEU-1 & 0.2398 & 0.1384 & 0.2778 & 0.1823 & 0.2395 & 0.3065 & 0.3319 & 0.3685 & 0.2176 & 0.4230 & \cellcolor{TopTwo} 0.4352 & 0.4100 & \cellcolor{TopOne} 0.4504 \\
          &       & EM    & 0.2011 & 0.1092 & 0.1897 & 0.1379 & 0.2011 & 0.2414 & 0.2806 & 0.2931 & 0.1839 & \cellcolor{TopTwo} 0.3333 & \cellcolor{TopTwo} 0.3333 & \cellcolor{TopTwo} 0.3333 & \cellcolor{TopOne} 0.3563 \\
          &       & LLM-Judge & 0.3161 & 0.1810 & 0.3822 & 0.2615 & 0.3190 & 0.4310 & 0.4101 & 0.5172 & 0.2615 & 0.5374 & \cellcolor{TopTwo} 0.5489 & 0.5201 & \cellcolor{TopOne} 0.5690 \\
\cmidrule{2-16}    & \multirow{4}[2]{*}{MR} & F1    & 0.2600 & 0.1934 & 0.4380 & 0.2683 & 0.2615 & 0.3400 & 0.4401 & 0.4483 & 0.2400 & \cellcolor{TopTwo} 0.4851 & \cellcolor{TopOne} 0.4908 & 0.4697 & 0.4773 \\
          &       & BLEU-1 & 0.1943 & 0.1364 & 0.3277 & 0.1997 & 0.1969 & 0.2599 & 0.3162 & 0.3474 & 0.1787 & \cellcolor{TopTwo} 0.3736 & \cellcolor{TopOne} 0.3788 & 0.3602 & 0.3639 \\
          &       & EM    & 0.0291 & 0.0243 & 0.0825 & 0.0388 & 0.0291 & 0.0340 & 0.0722 & \cellcolor{TopTwo} 0.0971 & 0.0194 & 0.0874 & 0.0922 & 0.0874 & \cellcolor{TopOne} 0.1019 \\
          &       & LLM-Judge & 0.4442 & 0.2985 & 0.7670 & 0.4345 & 0.4442 & 0.6092 & 0.7474 & 0.7597 & 0.4223 & \cellcolor{TopTwo} 0.8083 & \cellcolor{TopOne} 0.8301 & 0.7985 & 0.7840 \\      
    \midrule
    \multirow{12}[6]{*}{\rotatebox{90}{Knowledge Management}} & \multirow{4}[2]{*}{KR} & F1    & 0.2034 & 0.1563 & 0.3382 & 0.2613 & 0.2061 & 0.2372 & \cellcolor{TopOne} 0.4280 & 0.3706 & 0.2007 & \cellcolor{TopTwo} 0.3975 & 0.3883 & 0.3441 & 0.3604 \\
          &       & BLEU-1 & 0.1627 & 0.1118 & 0.2754 & 0.2085 & 0.1639 & 0.1869 & \cellcolor{TopOne} 0.3693 & 0.3094 & 0.1625 & \cellcolor{TopTwo} 0.3358 & 0.3323 & 0.2842 & 0.2987 \\
          &       & EM    & 0.0741 & 0.0370 & 0.1111 & 0.0864 & 0.0741 & 0.0741 & \cellcolor{TopOne} 0.2267 & 0.1235 & 0.0864 & 0.1481 & \cellcolor{TopTwo} 0.1728 & 0.1358 & 0.1235 \\
          &       & LLM-Judge & 0.4012 & 0.2963 & 0.5926 & 0.4012 & 0.3951 & 0.5185 & 0.6600 & \cellcolor{TopTwo} 0.6852 & 0.3704 & 0.6790 & \cellcolor{TopTwo} 0.6852 & 0.6420 & \cellcolor{TopOne} 0.6914 \\
\cmidrule{2-16}          & \multirow{4}[2]{*}{CD} & F1    & 0.3333 & \cellcolor{TopOne} 0.3580 & 0.3210 & 0.3210 & 0.3333 & \cellcolor{TopOne} 0.3580 & 0.3200 & 0.2963 & \cellcolor{TopTwo} 0.3457 & 0.3086 & 0.3086 & 0.3086 & 0.3333 \\
          &       & BLEU-1 & 0.3333 & \cellcolor{TopOne} 0.3580 & 0.3210 & 0.3210 & 0.3333 & \cellcolor{TopOne} 0.3580 & 0.3200 & 0.2963 & \cellcolor{TopTwo} 0.3457 & 0.3086 & 0.3086 & 0.3086 & 0.3333 \\
          &       & EM    & 0.3333 & \cellcolor{TopOne} 0.3580 & 0.3210 & 0.3210 & 0.3333 & \cellcolor{TopOne} 0.3580 & 0.3200 & 0.2963 & \cellcolor{TopTwo} 0.3457 & 0.3086 & 0.3086 & 0.3086 & 0.3333 \\
          &       & LLM-Judge & 0.3333 & \cellcolor{TopOne} 0.3580 & 0.3210 & 0.3210 & 0.3333 & \cellcolor{TopOne} 0.3580 & 0.3200 & 0.2963 & \cellcolor{TopTwo} 0.3457 & 0.3086 & 0.3086 & 0.3086 & 0.3333 \\
\cmidrule{2-16}   & \multirow{4}[2]{*}{AR} & F1    & \cellcolor{TopTwo} 0.9742 & \cellcolor{TopOne} 0.9783 & 0.8942 & 0.9578 & \cellcolor{TopTwo} 0.9742 & 0.8387 & 0.9080 & 0.8992 & 0.9647 & 0.8823 & 0.8661 & 0.8707 & 0.8553 \\
          &       & BLEU-1 & \cellcolor{TopTwo} 0.9730 & \cellcolor{TopOne} 0.9783 & 0.8924 & 0.9571 & \cellcolor{TopTwo} 0.9730 & 0.8355 & 0.9080 & 0.8982 & 0.9628 & 0.8811 & 0.8648 & 0.8702 & 0.8540 \\
          &       & EM    & \cellcolor{TopTwo} 0.9728 & \cellcolor{TopOne} 0.9783 & 0.8913 & 0.9565 & \cellcolor{TopTwo} 0.9728 & 0.8315 & 0.9080 & 0.8967 & 0.9620 & 0.8804 & 0.8641 & 0.8696 & 0.8533 \\
          &       & LLM-Judge & \cellcolor{TopOne} 0.9620 & \cellcolor{TopOne} 0.9620 & 0.8804 & 0.9429 & \cellcolor{TopOne} 0.9620 & 0.8234 & 0.9052 & 0.8804 & \cellcolor{TopTwo} 0.9484 & 0.8641 & 0.8478 & 0.8587 & 0.8424 \\       
    \midrule
    \multicolumn{2}{c}{\multirow{4}[2]{*}{Overall}} & F1    & 0.3798 & 0.2986 & 0.5832 & 0.4013 & 0.3793 & 0.4165 & 0.6059 & 0.6013 & 0.3665 & \cellcolor{TopOne} 0.6679 & \cellcolor{TopTwo} 0.6479 & 0.6443 & 0.6383 \\
    \multicolumn{2}{c}{} & BLEU-1 & 0.3492 & 0.2689 & 0.5333 & 0.3620 & 0.3487 & 0.3794 & 0.5505 & 0.5462 & 0.3360 & \cellcolor{TopOne} 0.6194 & \cellcolor{TopTwo} 0.5995 & 0.5953 & 0.5881 \\
    \multicolumn{2}{c}{} & EM    & 0.2624 & 0.2040 & 0.3933 & 0.2671 & 0.2618 & 0.2700 & 0.4303 & 0.4103 & 0.2496 & \cellcolor{TopOne} 0.4728 & 0.4594 & \cellcolor{TopTwo} 0.4670 & 0.4541 \\
    \multicolumn{2}{c}{} & LLM-Judge & 0.4541 & 0.3521 & 0.6856 & 0.4597 & 0.4538 & 0.5383 & 0.6886 & 0.7162 & 0.4272 & \cellcolor{TopOne} 0.7756 & \cellcolor{TopTwo} 0.7662 & 0.7463 & 0.7528 \\
    \bottomrule
    \end{tabular}%
    }
  \label{tab:result_qwen_vl_3b}%
\end{table*}%

\begin{table*}[t!]
  \centering
  \caption{Evaluation results on GPT-4.1-Nano. The best and second-performed memory model(s) are highlighted with \textcolor{orange}{orange} and \textcolor{blue}{blue} backgrounds.}
  \resizebox{\linewidth}{!}{%
    \begin{tabular}{ccc|cccccccc|ccccc}
    \toprule
    \rowcolor{gptbar} \multicolumn{3}{c|}{\openailogo\ GPT-4.1-Nano} & Full (Text) & FIFO  & NaiveRAG & Gen. Agent & Reflexion & MemGPT & A-Mem & MemoryOS & Full (MM) & MuRAG & UniversalRAG & NGM   & AUGUSTUS \\
    \midrule
    \multirow{12}[6]{*}{\rotatebox{90}{Extract. \& Adapt.}} & \multirow{4}[2]{*}{FR} & F1    & 0.2279 & 0.1379 & 0.5498 & 0.1672 & 0.2277 & 0.5983 & 0.6048 & 0.5848 & 0.1920 & \cellcolor{TopOne} 0.6256 & \cellcolor{TopTwo} 0.6060 & 0.5922 & 0.5666 \\
          &       & BLEU-1 & 0.1724 & 0.0907 & 0.4682 & 0.1164 & 0.1719 & 0.5193 & \cellcolor{TopTwo} 0.5194 & 0.4936 & 0.1413 & \cellcolor{TopOne} 0.5392 & 0.5189 & 0.5036 & 0.4734 \\
          &       & EM    & 0.0685 & 0.0274 & 0.2374 & 0.0228 & 0.0594 & 0.2740 & \cellcolor{TopTwo} 0.2648 & 0.2466 & 0.0548 & \cellcolor{TopOne} 0.2785 & 0.2511 & \cellcolor{TopOne} 0.2785 & 0.2466 \\
          &       & LLM-Judge & 0.2260 & 0.0936 & 0.7740 & 0.1918 & 0.2283 & 0.8082 & 0.8151 & \cellcolor{TopTwo} 0.8425 & 0.1872 & \cellcolor{TopOne} 0.8653 & 0.8356 & 0.7968 & 0.7922 \\
\cmidrule{2-16}          & \multirow{4}[2]{*}{VS} & F1    & 0.2330 & 0.0839 & 0.7457 & 0.1787 & 0.2356 & 0.6853 & 0.7965 & 0.7639 & 0.1318 & \cellcolor{TopOne} 0.8478 & \cellcolor{TopTwo} 0.8410 & 0.8329 & 0.8354 \\
          &       & BLEU-1 & 0.2175 & 0.0689 & 0.7021 & 0.1507 & 0.2210 & 0.6573 & 0.7553 & 0.6962 & 0.1155 & \cellcolor{TopTwo} 0.8092 & \cellcolor{TopOne} 0.8109 & 0.7906 & 0.7831 \\
          &       & EM    & 0.1732 & 0.0294 & 0.5294 & 0.0980 & 0.1732 & 0.5392 & 0.5882 & 0.5621 & 0.0588 & 0.6275 & \cellcolor{TopTwo} 0.6340 & \cellcolor{TopTwo} 0.6340 & \cellcolor{TopOne} 0.6373 \\
          &       & LLM-Judge & 0.2124 & 0.0507 & 0.6944 & 0.1585 & 0.2124 & 0.6536 & 0.7761 & 0.7255 & 0.0997 & \cellcolor{TopOne} 0.8431 & \cellcolor{TopTwo} 0.8252 & 0.8056 & 0.8121 \\
\cmidrule{2-16}    & \multirow{4}[2]{*}{TTL} & F1    & 0.2278 & 0.1473 & 0.3618 & 0.2021 & 0.2226 & 0.2866 & 0.3279 & 0.3263 & 0.1991 & 0.4395 & 0.4407 & \cellcolor{TopOne} 0.5273 & \cellcolor{TopTwo} 0.4488 \\
          &       & BLEU-1 & 0.1609 & 0.0974 & 0.2523 & 0.1414 & 0.1567 & 0.1969 & 0.2232 & 0.2346 & 0.1406 & 0.3468 & \cellcolor{TopTwo} 0.3581 & \cellcolor{TopOne} 0.4372 & 0.3491 \\
          &       & EM    & 0.0504 & 0.0148 & 0.0504 & 0.0297 & 0.0445 & 0.0386 & 0.0326 & 0.0742 & 0.0356 & 0.1662 & \cellcolor{TopTwo} 0.1958 & \cellcolor{TopOne} 0.2641 & 0.1424 \\
          &       & LLM-Judge & 0.5742 & 0.4585 & 0.7834 & 0.5801 & 0.5579 & 0.6588 & 0.8027 & 0.7389 & 0.5015 & 0.7804 & 0.8012 & \cellcolor{TopOne} 0.8709 & \cellcolor{TopTwo} 0.8145 \\      
    \midrule
    \multirow{12}[6]{*}{\rotatebox{90}{Reasoning}} & \multirow{4}[2]{*}{TR} & F1    & 0.2035 & 0.1356 & 0.4381 & 0.2028 & 0.1997 & 0.4306 & \cellcolor{TopOne} 0.5116 & \cellcolor{TopTwo} 0.5017 & 0.1916 & 0.5007 & 0.4366 & 0.4924 & 0.4808 \\
          &       & BLEU-1 & 0.1744 & 0.1143 & 0.4032 & 0.1746 & 0.1694 & 0.3932 & \cellcolor{TopOne} 0.4730 & 0.4649 & 0.1682 & \cellcolor{TopTwo} 0.4676 & 0.3997 & 0.4606 & 0.4484 \\
          &       & EM    & 0.0569 & 0.0488 & 0.2764 & 0.0894 & 0.0650 & 0.2520 & \cellcolor{TopTwo} 0.3415 & 0.3171 & 0.0650 & 0.3252 & 0.2683 & \cellcolor{TopOne} 0.3496 & 0.3252 \\
          &       & LLM-Judge & 0.2764 & 0.1382 & 0.7154 & 0.2886 & 0.2846 & 0.7276 & \cellcolor{TopTwo} 0.7846 & \cellcolor{TopOne} 0.8130 & 0.2398 & 0.7520 & 0.7480 & 0.7236 & 0.7154 \\
\cmidrule{2-16}          & \multirow{4}[2]{*}{VR} & F1    & 0.2011 & 0.0714 & 0.2041 & 0.0987 & 0.2000 & \cellcolor{TopTwo} 0.4140 & 0.3269 & 0.3697 & 0.0909 & 0.3426 & \cellcolor{TopOne} 0.4417 & 0.3043 & 0.3817 \\
          &       & BLEU-1 & 0.1796 & 0.0491 & 0.1791 & 0.0747 & 0.1772 & \cellcolor{TopTwo} 0.3855 & 0.2979 & 0.3431 & 0.0690 & 0.3077 & \cellcolor{TopOne} 0.4146 & 0.2738 & 0.3518 \\
          &       & EM    & 0.1379 & 0.0172 & 0.0862 & 0.0172 & 0.1322 & \cellcolor{TopTwo} 0.2931 & 0.1954 & 0.2586 & 0.0345 & 0.2069 & \cellcolor{TopOne} 0.3218 & 0.1724 & 0.2529 \\
          &       & LLM-Judge & 0.3276 & 0.1379 & 0.3736 & 0.1810 & 0.3305 & \cellcolor{TopTwo} 0.6207 & 0.5661 & 0.5862 & 0.2011 & 0.5575 & \cellcolor{TopOne} 0.6638 & 0.4713 & 0.5747 \\
\cmidrule{2-16}          & \multirow{4}[2]{*}{MR} & F1    & 0.2415 & 0.1814 & 0.4490 & 0.2354 & 0.2389 & 0.4428 & 0.4669 & 0.4523 & 0.2017 & \cellcolor{TopOne} 0.5024 & 0.4670 & 0.4570 & \cellcolor{TopTwo} 0.4829 \\
          &       & BLEU-1 & 0.1720 & 0.1174 & 0.3402 & 0.1704 & 0.1710 & 0.3388 & 0.3602 & 0.3479 & 0.1385 & \cellcolor{TopOne} 0.3946 & 0.3670 & 0.3506 & \cellcolor{TopTwo} 0.3798 \\
          &       & EM    & 0.0194 & 0.0097 & 0.0388 & 0.0097 & 0.0146 & 0.0388 & 0.0437 & \cellcolor{TopTwo} 0.0583 & 0.0097 & 0.0534 & 0.0534 & 0.0534 & \cellcolor{TopOne} 0.0680 \\
          &       & LLM-Judge & 0.2451 & 0.0922 & 0.7985 & 0.2597 & 0.2500 & 0.7985 & \cellcolor{TopOne} 0.8617 & \cellcolor{TopTwo} 0.8325 & 0.1942 & \cellcolor{TopOne} 0.8617 & 0.7816 & 0.8180 & 0.8180 \\
    \midrule
    \multirow{12}[6]{*}{\rotatebox{90}{Knowledge Management}} & \multirow{4}[2]{*}{KR} & F1    & 0.1454 & 0.1001 & 0.2852 & 0.1408 & 0.1411 & \cellcolor{TopOne} 0.3733 & 0.3283 & 0.2873 & 0.1293 & \cellcolor{TopTwo} 0.3626 & 0.2972 & 0.2878 & 0.2607 \\
          &       & BLEU-1 & 0.0965 & 0.0633 & 0.2174 & 0.0960 & 0.0934 & \cellcolor{TopOne} 0.3017 & 0.2556 & 0.2190 & 0.0902 & \cellcolor{TopTwo} 0.2897 & 0.2316 & 0.2252 & 0.1958 \\
          &       & EM    & 0.0123 & 0.0000 & 0.0494 & 0.0000 & 0.0123 & \cellcolor{TopOne} 0.1111 & 0.0741 & 0.0370 & 0.0123 & \cellcolor{TopTwo} 0.0988 & 0.0494 & 0.0864 & 0.0617 \\
          &       & LLM-Judge & 0.2840 & 0.1975 & 0.6049 & 0.2407 & 0.2963 & \cellcolor{TopOne} 0.7593 & 0.6852 & 0.6790 & 0.2531 & \cellcolor{TopTwo} 0.7469 & 0.6543 & 0.6173 & 0.5741 \\
\cmidrule{2-16}          & \multirow{4}[2]{*}{CD} & F1    & 0.3704 & 0.3704 & \cellcolor{TopTwo} 0.4321 & 0.4074 & 0.3704 & 0.3951 & \cellcolor{TopTwo} 0.4321 & \cellcolor{TopOne} 0.4444 & 0.3704 & 0.4074 & 0.3827 & \cellcolor{TopTwo} 0.4321 & 0.3951 \\
          &       & BLEU-1 & 0.3704 & 0.3704 & \cellcolor{TopTwo} 0.4321 & 0.4074 & 0.3704 & 0.3951 & \cellcolor{TopTwo} 0.4321 & \cellcolor{TopOne} 0.4444 & 0.3704 & 0.4074 & 0.3827 & \cellcolor{TopTwo} 0.4321 & 0.3951 \\
          &       & EM    & 0.3704 & 0.3704 & \cellcolor{TopTwo} 0.4321 & 0.4074 & 0.3704 & 0.3951 & \cellcolor{TopTwo} 0.4321 & \cellcolor{TopOne} 0.4444 & 0.3704 & 0.4074 & 0.3827 & \cellcolor{TopTwo} 0.4321 & 0.3951 \\
          &       & LLM-Judge & 0.3704 & 0.3704 & \cellcolor{TopTwo} 0.4321 & 0.4074 & 0.3704 & 0.3951 & \cellcolor{TopTwo} 0.4321 & \cellcolor{TopOne} 0.4444 & 0.3704 & 0.4074 & 0.3827 & \cellcolor{TopTwo} 0.4321 & 0.3951 \\
\cmidrule{2-16}   & \multirow{4}[2]{*}{AR} & F1    & 0.9682 & \cellcolor{TopOne} 0.9891 & 0.8406 & 0.9751 & 0.9737 & 0.7911 & 0.8496 & 0.8874 & \cellcolor{TopTwo} 0.9791 & 0.9039 & 0.9102 & 0.9382 & 0.9058 \\
          &       & BLEU-1 & 0.9679 & \cellcolor{TopOne} 0.9891 & 0.8323 & 0.9738 & 0.9733 & 0.7811 & 0.8430 & 0.8831 & \cellcolor{TopTwo} 0.9788 & 0.9003 & 0.9058 & 0.9335 & 0.9009 \\
          &       & EM    & 0.9674 & \cellcolor{TopOne} 0.9891 & 0.8261 & 0.9728 & 0.9728 & 0.7717 & 0.8370 & 0.8804 & \cellcolor{TopTwo} 0.9783 & 0.8967 & 0.9022 & 0.9293 & 0.8967 \\
          &       & LLM-Judge & 0.9565 & \cellcolor{TopOne} 0.9728 & 0.8587 & \cellcolor{TopTwo} 0.9647 & 0.9592 & 0.8370 & 0.8668 & 0.8940 & 0.9620 & 0.9022 & 0.9103 & 0.9321 & 0.9022 \\       
    \midrule
    \multicolumn{2}{c}{\multirow{4}[2]{*}{Overall}} & F1    & 0.3084 & 0.2291 & 0.5057 & 0.2769 & 0.3075 & 0.5034 & 0.5380 & 0.5339 & 0.2636 & \cellcolor{TopTwo} 0.5832 & 0.5774 & \cellcolor{TopOne} 0.5849 & 0.5703 \\
    \multicolumn{2}{c}{} & BLEU-1 & 0.2703 & 0.1974 & 0.4436 & 0.2389 & 0.2698 & 0.4480 & 0.4764 & 0.4705 & 0.2293 & \cellcolor{TopTwo} 0.5243 & 0.5236 & \cellcolor{TopOne} 0.5266 & 0.5080 \\
    \multicolumn{2}{c}{} & EM    & 0.1923 & 0.1420 & 0.2800 & 0.1596 & 0.1899 & 0.2987 & 0.3092 & 0.3203 & 0.1572 & 0.3518 & \cellcolor{TopTwo} 0.3600 & \cellcolor{TopOne} 0.3746 & 0.3489 \\
    \multicolumn{2}{c}{} & LLM-Judge & 0.3966 & 0.2779 & 0.7046 & 0.3720 & 0.3960 & 0.7063 & 0.7651 & 0.7507 & 0.3346 & \cellcolor{TopOne} 0.7814 & \cellcolor{TopTwo} 0.7747 & 0.7659 & 0.7583 \\
    \bottomrule
    \end{tabular}%
    }
  \label{tab:gpt_4_1_nano}%
\end{table*}%

\begin{table*}[t!]
  \centering
  \caption{Evaluation results on Gemini-2.5-Flash-Lite. The best and second-performed memory model(s) are highlighted with \textcolor{orange}{orange} and \textcolor{blue}{blue} backgrounds.}
  \resizebox{\linewidth}{!}{%
    \begin{tabular}{ccc|cccccccc|ccccc}
    \toprule
    \rowcolor{geminibar} \multicolumn{3}{c|}{\geminilogo\ Gemini-2.5-Flash-Lite} & Full (Text) & FIFO  & NaiveRAG & Gen. Agent & Reflexion & MemGPT & A-Mem & MemoryOS & Full (MM) & MuRAG & UniversalRAG & NGM   & AUGUSTUS \\
    \midrule
    \multirow{12}[6]{*}{\rotatebox{90}{Extract. \& Adapt.}} & \multirow{4}[2]{*}{FR} & F1    & 0.2014 & 0.1014 & 0.5905 & 0.2329 & 0.2014 & 0.6667 & 0.5266 & 0.6622 & 0.1534 & \cellcolor{TopTwo} 0.6888 & \cellcolor{TopOne} 0.7014 & 0.6341 & 0.6011 \\
          &       & BLEU-1 & 0.1652 & 0.0732 & 0.4999 & 0.1802 & 0.1652 & 0.5633 & 0.4270 & 0.5683 & 0.1214 & \cellcolor{TopTwo} 0.6009 & \cellcolor{TopOne} 0.6136 & 0.5510 & 0.5141 \\
          &       & EM    & 0.0822 & 0.0320 & 0.2648 & 0.0868 & 0.0822 & 0.2557 & 0.1690 & 0.3425 & 0.0594 & \cellcolor{TopOne} 0.3653 & \cellcolor{TopTwo} 0.3562 & 0.3333 & 0.3151 \\
          &       & LLM-Judge & 0.2237 & 0.0868 & 0.7922 & 0.2945 & 0.2215 & \cellcolor{TopOne} 0.9475 & 0.8402 & 0.8562 & 0.1507 & 0.8927 & \cellcolor{TopTwo} 0.9064 & 0.8082 & 0.7922 \\
\cmidrule{2-16}          & \multirow{4}[2]{*}{VS} & F1    & 0.2252 & 0.0598 & 0.7909 & 0.3450 & 0.2252 & \cellcolor{TopOne} 0.9204 & 0.8451 & 0.8427 & 0.1051 & 0.9077 & \cellcolor{TopTwo} 0.9175 & 0.8753 & 0.8811 \\
          &       & BLEU-1 & 0.2102 & 0.0553 & 0.7539 & 0.3156 & 0.2102 & \cellcolor{TopOne} 0.8957 & 0.8019 & 0.7918 & 0.0929 & 0.8837 & \cellcolor{TopTwo} 0.8919 & 0.8456 & 0.8443 \\
          &       & EM    & 0.1863 & 0.0327 & 0.6078 & 0.2451 & 0.1863 & \cellcolor{TopOne} 0.8105 & 0.6797 & 0.6536 & 0.0654 & \cellcolor{TopTwo} 0.7909 & 0.7745 & 0.7484 & 0.7157 \\
          &       & LLM-Judge & 0.2141 & 0.0458 & 0.7663 & 0.3170 & 0.2141 & \cellcolor{TopOne} 0.9265 & 0.8415 & 0.8350 & 0.0931 & 0.9052 & \cellcolor{TopTwo} 0.9199 & 0.8611 & 0.8775 \\
\cmidrule{2-16}     & \multirow{4}[2]{*}{TTL} & F1    & 0.4697 & 0.2002 & 0.7839 & 0.6149 & 0.4697 & 0.8517 & 0.7173 & 0.8360 & 0.3031 & \cellcolor{TopOne} 0.8750 & 0.8464 & 0.8477 & \cellcolor{TopTwo} 0.8549 \\
          &       & BLEU-1 & 0.4322 & 0.1260 & 0.7259 & 0.5552 & 0.4322 & \cellcolor{TopTwo} 0.8012 & 0.6229 & 0.7709 & 0.2654 & \cellcolor{TopOne} 0.8130 & 0.7689 & 0.7745 & 0.7908 \\
          &       & EM    & 0.3501 & 0.0712 & 0.6113 & 0.4481 & 0.3501 & \cellcolor{TopOne} 0.6795 & 0.4243 & 0.6380 & 0.1958 & \cellcolor{TopTwo} 0.6677 & 0.6410 & 0.6350 & 0.6499 \\
          &       & LLM-Judge & 0.5341 & 0.2507 & 0.8680 & 0.6869 & 0.5341 & 0.9021 & 0.9006 & 0.9021 & 0.3442 & \cellcolor{TopOne} 0.9258 & 0.8976 & \cellcolor{TopTwo} 0.9184 & 0.9139 \\     
    \midrule
    \multirow{12}[6]{*}{\rotatebox{90}{Reasoning}} & \multirow{4}[2]{*}{TR} & F1    & 0.2105 & 0.1447 & 0.4801 & 0.2201 & 0.2105 & 0.5475 & 0.4714 & 0.5325 & 0.1796 & \cellcolor{TopOne} 0.5806 & \cellcolor{TopTwo} 0.5675 & 0.5280 & 0.5555 \\
          &       & BLEU-1 & 0.1953 & 0.1345 & 0.4513 & 0.2046 & 0.1953 & 0.5071 & 0.4301 & 0.5047 & 0.1637 & \cellcolor{TopOne} 0.5507 & \cellcolor{TopTwo} 0.5392 & 0.5011 & 0.5267 \\
          &       & EM    & 0.0976 & 0.0407 & 0.3333 & 0.0732 & 0.0976 & 0.3496 & 0.2927 & 0.3821 & 0.0650 & \cellcolor{TopOne} 0.4309 & \cellcolor{TopTwo} 0.4146 & 0.3984 & 0.4065 \\
          &       & LLM-Judge & 0.2317 & 0.1423 & 0.6870 & 0.2846 & 0.2317 & \cellcolor{TopOne} 0.8902 & 0.7967 & 0.7358 & 0.1911 & 0.7886 & \cellcolor{TopTwo} 0.8089 & 0.6992 & 0.7886 \\
\cmidrule{2-16}          & \multirow{4}[2]{*}{VR} & F1    & 0.1909 & 0.0423 & 0.3135 & 0.1466 & 0.1909 & \cellcolor{TopOne} 0.6294 & 0.4262 & 0.3997 & 0.0798 & 0.4470 & \cellcolor{TopTwo} 0.4954 & 0.4422 & 0.4533 \\
          &       & BLEU-1 & 0.1819 & 0.0356 & 0.2985 & 0.1398 & 0.1819 & \cellcolor{TopOne} 0.6083 & 0.4075 & 0.3843 & 0.0707 & 0.4267 & \cellcolor{TopTwo} 0.4785 & 0.4200 & 0.4325 \\
          &       & EM    & 0.1494 & 0.0230 & 0.2241 & 0.1092 & 0.1494 & \cellcolor{TopOne} 0.5230 & 0.3276 & 0.3103 & 0.0517 & 0.3563 & \cellcolor{TopTwo} 0.4023 & 0.3563 & 0.3563 \\
          &       & LLM-Judge & 0.2443 & 0.0402 & 0.4138 & 0.2443 & 0.2443 & \cellcolor{TopOne} 0.7845 & 0.5603 & 0.5805 & 0.1063 & 0.6264 & \cellcolor{TopTwo} 0.6782 & 0.5920 & 0.5718 \\
\cmidrule{2-16}          & \multirow{4}[2]{*}{MR} & F1    & 0.1648 & 0.0834 & 0.4521 & 0.1765 & 0.1648 & \cellcolor{TopOne} 0.5210 & 0.4529 & \cellcolor{TopTwo} 0.5161 & 0.1268 & 0.5137 & 0.5137 & 0.4783 & 0.4691 \\
          &       & BLEU-1 & 0.1250 & 0.0594 & 0.3508 & 0.1265 & 0.1250 & \cellcolor{TopOne} 0.4130 & 0.3553 & \cellcolor{TopTwo} 0.4111 & 0.0987 & 0.4032 & 0.4011 & 0.3703 & 0.3575 \\
          &       & EM    & 0.0097 & 0.0097 & 0.0631 & 0.0097 & 0.0097 & 0.0680 & 0.0388 & \cellcolor{TopOne} 0.0825 & 0.0097 & 0.0680 & \cellcolor{TopTwo} 0.0728 & 0.0437 & 0.0631 \\
          &       & LLM-Judge & 0.2184 & 0.0922 & 0.7767 & 0.2743 & 0.2184 & \cellcolor{TopOne} 0.9272 & 0.8519 & 0.8568 & 0.1481 & 0.8714 & \cellcolor{TopTwo} 0.8738 & 0.8058 & 0.7864 \\
    \midrule
    \multirow{12}[6]{*}{\rotatebox{90}{Knowledge Management}} & \multirow{4}[2]{*}{KR} & F1    & 0.1157 & 0.0604 & 0.2813 & 0.1050 & 0.1157 & \cellcolor{TopTwo} 0.3915 & 0.3341 & 0.3609 & 0.0918 & \cellcolor{TopOne} 0.4037 & 0.3710 & 0.3111 & 0.3168 \\
          &       & BLEU-1 & 0.0848 & 0.0370 & 0.2177 & 0.0810 & 0.0848 & \cellcolor{TopTwo} 0.3225 & 0.2644 & 0.2956 & 0.0673 & \cellcolor{TopOne} 0.3534 & 0.3156 & 0.2607 & 0.2631 \\
          &       & EM    & 0.0247 & 0.0000 & 0.0864 & 0.0247 & 0.0247 & \cellcolor{TopTwo} 0.1605 & 0.0988 & \cellcolor{TopTwo} 0.1605 & 0.0247 & \cellcolor{TopOne} 0.1975 & 0.1481 & 0.1358 & 0.1235 \\
          &       & LLM-Judge & 0.2531 & 0.1296 & 0.6235 & 0.2037 & 0.2531 & \cellcolor{TopOne} 0.8025 & 0.7593 & \cellcolor{TopTwo} 0.7840 & 0.1728 & 0.7716 & 0.7222 & 0.5988 & 0.6667 \\
\cmidrule{2-16}          & \multirow{4}[2]{*}{CD} & F1    & 0.3951 & 0.3827 & 0.5185 & 0.4198 & 0.3951 & 0.5556 & 0.4938 & \cellcolor{TopTwo} 0.6049 & 0.3827 & 0.5309 & 0.5432 & \cellcolor{TopOne} 0.6296 & 0.5679 \\
          &       & BLEU-1 & 0.3951 & 0.3827 & 0.5185 & 0.4198 & 0.3951 & 0.5556 & 0.4938 & \cellcolor{TopTwo} 0.6049 & 0.3827 & 0.5309 & 0.5432 & \cellcolor{TopOne} 0.6296 & 0.5679 \\
          &       & EM    & 0.3951 & 0.3827 & 0.5185 & 0.4198 & 0.3951 & 0.5556 & 0.4938 & \cellcolor{TopTwo} 0.6049 & 0.3827 & 0.5309 & 0.5432 & \cellcolor{TopOne} 0.6296 & 0.5679 \\
          &       & LLM-Judge & 0.3951 & 0.3827 & 0.5185 & 0.4198 & 0.3951 & 0.5556 & 0.5123 & \cellcolor{TopTwo} 0.6049 & 0.3827 & 0.5309 & 0.5432 & \cellcolor{TopOne} 0.6296 & 0.5679 \\
\cmidrule{2-16}   & \multirow{4}[2]{*}{AR} & F1    & 0.9644 & \cellcolor{TopOne} 0.9946 & 0.9122 & 0.9688 & 0.9644 & 0.8512 & 0.9175 & 0.9220 & \cellcolor{TopTwo} 0.9844 & 0.9242 & 0.9186 & 0.9332 & 0.9339 \\
          &       & BLEU-1 & 0.9634 & \cellcolor{TopOne} 0.9946 & 0.9102 & 0.9682 & 0.9634 & 0.8454 & 0.9155 & 0.9205 & \cellcolor{TopTwo} 0.9841 & 0.9218 & 0.9162 & 0.9318 & 0.9321 \\
          &       & EM    & 0.9620 & \cellcolor{TopOne} 0.9946 & 0.9076 & 0.9674 & 0.9620 & 0.8370 & 0.9130 & 0.9185 & \cellcolor{TopTwo} 0.9837 & 0.9185 & 0.9130 & 0.9293 & 0.9293 \\
          &       & LLM-Judge & 0.9511 & \cellcolor{TopOne} 0.9783 & 0.8995 & 0.9592 & 0.9511 & 0.8478 & 0.9212 & 0.9212 & \cellcolor{TopTwo} 0.9674 & 0.9130 & 0.9076 & 0.9212 & 0.9239 \\       
    \midrule
    \multicolumn{2}{c}{\multirow{4}[2]{*}{Overall}} & F1    & 0.3408 & 0.2158 & 0.6282 & 0.3936 & 0.3408 & \cellcolor{TopOne} 0.7202 & 0.6294 & 0.6861 & 0.2627 & 0.7155 & \cellcolor{TopTwo} 0.7157 & 0.6901 & 0.6877 \\
    \multicolumn{2}{c}{} & BLEU-1 & 0.3178 & 0.1913 & 0.5796 & 0.3608 & 0.3178 & \cellcolor{TopOne} 0.6706 & 0.5702 & 0.6327 & 0.2424 & \cellcolor{TopTwo} 0.6676 & 0.6644 & 0.6400 & 0.6370 \\
    \multicolumn{2}{c}{} & EM    & 0.2595 & 0.1555 & 0.4436 & 0.2858 & 0.2595 & \cellcolor{TopTwo} 0.5219 & 0.4120 & 0.4904 & 0.1940 & \cellcolor{TopOne} 0.5283 & 0.5207 & 0.5079 & 0.5020 \\
    \multicolumn{2}{c}{} & LLM-Judge & 0.3729 & 0.2236 & 0.7452 & 0.4407 & 0.3726 & \cellcolor{TopOne} 0.8755 & 0.8115 & 0.8165 & 0.2764 & 0.8437 & \cellcolor{TopTwo} 0.8472 & 0.8030 & 0.8057 \\
    \bottomrule
    \end{tabular}%
    }
  \label{tab:gemini_flash_lite}%
\end{table*}%

\subsubsection{Memory Retrieval Analysis}\label{sec:retrieval_number}

\begin{figure}[tbp]
\centering
\includegraphics[width=\linewidth]{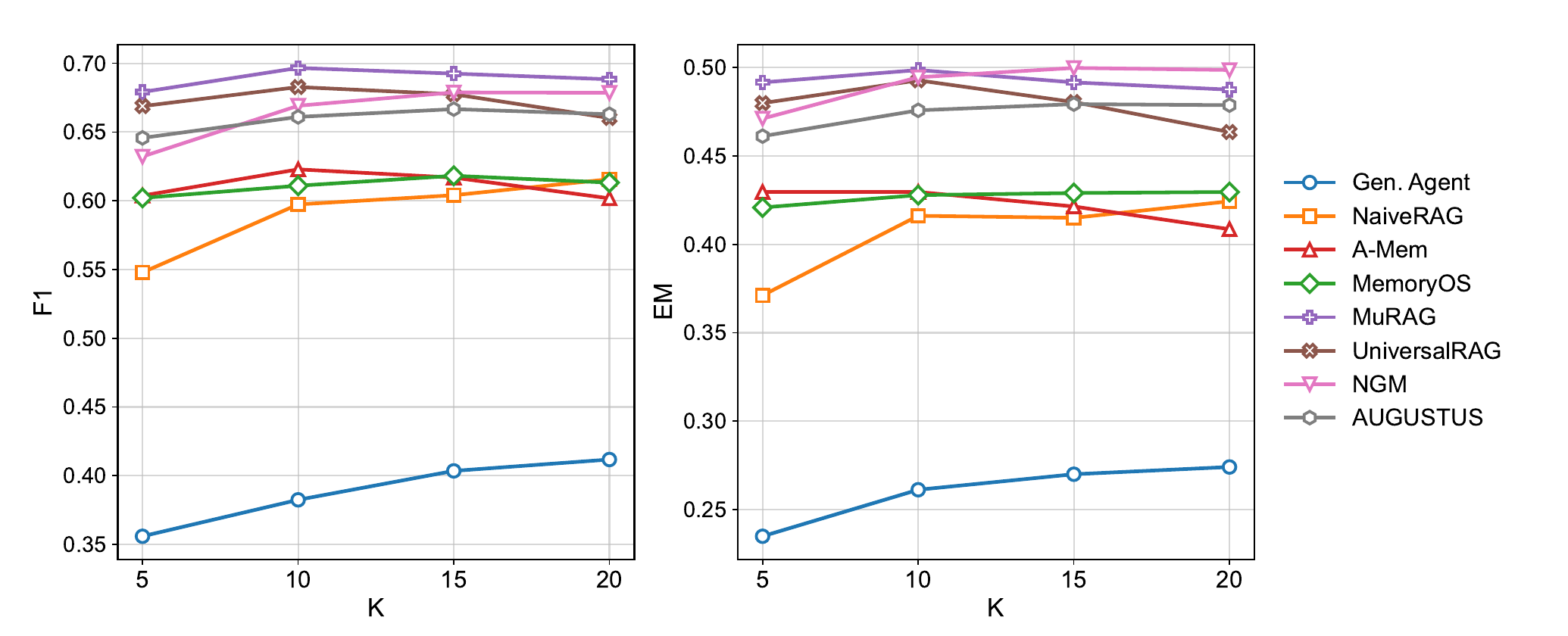}
\caption{Parameter study on the retrieval size $K$.}
\label{fig:retrieval_num}
\end{figure}

This section analyzes how the number of retrieved memory entries affects both retrieval effectiveness and downstream task performance. Table~\ref{tab:result_k_5}, Table~\ref{tab:main_qwen_vl_7b}, Table~\ref{tab:k_15}, and Table~\ref{tab:k_20} report task-level QA results under different retrieval sizes $K\in\{5,10,15,20\}$. Figure~\ref{fig:retrieval_num} shows the overall task performance trend under $K$ value changes.
Based on the fine-grained annotated clues for each evaluation data, Table~\ref{tab:full_retrieval} summarizes the corresponding retrieval metrics, including Recall@$K$, Precision@$K$, and Hit@$K$ for retrieval behavior analysis. All results are evaluated on the Qwen-2.5-VL-7B backbone.

\textbf{Impact of Retrieval Size on Downstream Task Performance}. The results in Table~\ref{tab:result_k_5}, Table~\ref{tab:main_qwen_vl_7b}, Table~\ref{tab:k_15}, and Table~\ref{tab:k_20} provide a continuous view of how retrieval quantity affects memory utilization and task performance. Across most memory models and task categories, we generally observe a non-monotonic relationship between retrieval size and downstream performance. Increasing the retrieval size from a small value ($K$=5) to a moderate range ($K$=10 or $K$=15) generally leads to consistent improvements, especially for memory extraction and adaptation tasks such as factual retrieval, visual-centric search, and test-time learning.

However, further increasing the retrieval size to $K$=20 does not consistently yield additional gains. As shown in Table~\ref{tab:k_20}, performance often saturates or fluctuates across tasks, and in some cases slightly degrades compared to the $K$=10 or $K$=15 settings. This pattern indicates that retrieving more memory entries beyond a moderate threshold does not necessarily translate into better task outcomes. This effect is particularly pronounced for multimodal memory methods. While larger retrieval sizes improve coverage of relevant multimodal entries, they also introduce additional redundant or weakly aligned visual–textual information. As a result, the benefits of higher recall are partially offset by increased noise in the retrieved memory, limiting the effectiveness of downstream reasoning and answer generation. In contrast, moderate retrieval sizes such as $K$=10, as reflected in Table~\ref{tab:main_qwen_vl_7b}, appear to offer a better balance between relevance and redundancy for multimodal memory utilization.

For reasoning-oriented tasks, including temporal reasoning, visual-centric reasoning, and multi-entity reasoning, the impact of retrieval size is even more constrained. Performance differences across $K$=5, $K$=10, $K$=15, and $K$=20 are relatively small, suggesting that reasoning performance is less sensitive to retrieval quantity and more dependent on how retrieved memory is structured and integrated. This observation holds consistently across both textual and multimodal memory methods.

Knowledge management tasks, such as knowledge resolution, conflict detection, and answer refusal, exhibit the weakest sensitivity to retrieval size. Across all values of $K$, task performance remains relatively stable, indicating that simply retrieving more memory entries is insufficient to improve consistency-oriented behaviors and knowledge management capability.

Overall, these results suggest that moderate retrieval sizes, exemplified by $K$=10, often provide the most effective trade-off between retrieval coverage and memory usability. Increasing retrieval size beyond this range yields diminishing returns, particularly for multimodal memory systems, highlighting the importance of selective and relevance-aware retrieval rather than indiscriminate scaling of retrieved memory entries.

\textbf{Retrieval Effectiveness versus Memory Utilization}.
Table~\ref{tab:full_retrieval} provides further insight into why increasing retrieval size does not consistently improve downstream task performance observed in Table~\ref{tab:main_qwen_vl_7b} and Table~\ref{tab:result_k_5}-\ref{tab:k_20}. As the retrieval size $K$ increases, retrieval effectiveness metrics exhibit a clear and expected trend: Recall@$K$ and Hit@$K$ consistently improve across all retrievers, indicating that more relevant memory entries are successfully retrieved. This trend holds for both textual and multimodal retrievers, and is especially pronounced for multimodal methods such as MuRAG and UniversalRAG, which achieve high recall even at moderate $K$ value.

However, this improvement in recall is accompanied by a systematic decline in Precision@$K$, particularly as $K$ increases beyond moderate values. While larger $K$ values ensure broader coverage of relevant memory entries, they also introduce a growing number of weakly relevant or redundant items. As a result, the retrieved memory becomes increasingly noisy, especially in multimodal settings where cross-modal alignment is imperfect. This divergence explains the observed decoupling between retrieval effectiveness and downstream task performance. Although Recall@$K$ continues to improve from $K$=10 to $K$=20, task performance in Table~\ref{tab:k_20} often saturates or fluctuates relative to Table~\ref{tab:main_qwen_vl_7b} and Table~\ref{tab:k_15}. In other words, higher recall does not guarantee better memory utilization. The additional retrieved entries do not consistently contribute useful evidence for reasoning or answer generation, and may instead distract the model during inference.

The effect is more evident for multimodal memory systems. Compared to textual memory, multimodal retrieval introduces heterogeneous visual–textual information, making it more sensitive to redundancy and misalignment. Consequently, multimodal methods benefit most from moderate retrieval sizes that balance recall and precision, as exemplified by the strong performance at $K$=10. Beyond this range, further recall gains are offset by increased retrieval noise, limiting improvements in downstream tasks.

Overall, these results indicate that retrieval quality, rather than retrieval quantity, is the dominant factor governing downstream memory performance. Simply scaling the retrieval size improves recall metrics but does not necessarily improve task outcomes. Effective multimodal memory systems therefore require selective, relevance-aware retrieval strategies that prioritize precision and alignment over exhaustive retrieval. Therefore, this highlights the need for future work to consider the principle that “less but high-quality is more” in multimodal memory design~\cite{chen2025less}.

\begin{table*}[t!]
  \centering
  \caption{Benchmark performance results on Qwen-2.5-VL-7B with the retrieval number $K$=5. The best and second-performed memory model(s) are highlighted with \textcolor{orange}{orange} and \textcolor{blue}{blue} backgrounds.}
  \resizebox{\linewidth}{!}{%
    \begin{tabular}{ccc|cccccccc|ccccc}
    \toprule
    \rowcolor{qwenbar} \multicolumn{3}{c|}{\qwenlogo\ Qwen-2.5-VL-7B (K=5)} & Full (Text) & FIFO  & NaiveRAG & Gen. Agent & Reflexion & MemGPT & A-Mem & MemoryOS & Full (MM) & MuRAG & UniversalRAG & NGM   & AUGUSTUS \\
    \midrule
    \multirow{12}[6]{*}{\rotatebox{90}{Extract. \& Adapt.}} & \multirow{4}[2]{*}{FR} & F1    & 0.2376 & 0.1075 & 0.5293 & 0.2216 & 0.2391 & 0.5928 & 0.5562 & 0.6045 & 0.2150 & \cellcolor{TopTwo} 0.6291 & \cellcolor{TopOne} 0.6321 & 0.5536 & 0.5745 \\
          &       & BLEU-1 & 0.1865 & 0.0683 & 0.4531 & 0.1684 & 0.1903 & 0.5098 & 0.4668 & 0.5229 & 0.1626 & \cellcolor{TopTwo} 0.5440 & \cellcolor{TopOne} 0.5464 & 0.4695 & 0.4870 \\
          &       & EM    & 0.0913 & 0.0228 & 0.2785 & 0.0685 & 0.0913 & 0.3288 & 0.2603 & 0.3288 & 0.0685 & \cellcolor{TopOne} 0.3516 & \cellcolor{TopTwo} 0.3379 & 0.3105 & 0.2968 \\
          &       & LLM-Judge & 0.2626 & 0.0662 & 0.7146 & 0.2580 & 0.2626 & \cellcolor{TopOne} 0.8539 & 0.7100 & 0.7877 & 0.2260 & \cellcolor{TopTwo} 0.8105 & 0.8059 & 0.7123 & 0.7489 \\
\cmidrule{2-16}          & \multirow{4}[2]{*}{VS} & F1    & 0.1992 & 0.0398 & 0.6864 & 0.2416 & 0.1954 & 0.6239 & 0.7366 & 0.7614 & 0.1658 & \cellcolor{TopTwo} 0.8664 & \cellcolor{TopOne} 0.8740 & 0.7847 & 0.8301 \\
          &       & BLEU-1 & 0.1873 & 0.0322 & 0.6307 & 0.2082 & 0.1840 & 0.5834 & 0.6596 & 0.7067 & 0.1473 & \cellcolor{TopTwo} 0.8301 & \cellcolor{TopOne} 0.8387 & 0.7417 & 0.7731 \\
          &       & EM    & 0.1601 & 0.0131 & 0.4967 & 0.1634 & 0.1569 & 0.4118 & 0.5719 & 0.5817 & 0.1078 & \cellcolor{TopTwo} 0.6928 & \cellcolor{TopOne} 0.6993 & 0.6340 & 0.6569 \\
          &       & LLM-Judge & 0.1961 & 0.0343 & 0.6650 & 0.2288 & 0.1895 & 0.5964 & 0.7059 & 0.7418 & 0.1683 & \cellcolor{TopOne} 0.8725 & \cellcolor{TopTwo} 0.8627 & 0.7778 & 0.8235 \\
\cmidrule{2-16}  & \multirow{4}[2]{*}{TTL} & F1    & 0.4500 & 0.3240 & 0.5938 & 0.4731 & 0.4486 & 0.2924 & 0.6882 & 0.5512 & 0.4147 & \cellcolor{TopOne} 0.8161 & \cellcolor{TopTwo} 0.7853 & 0.7706 & 0.7631 \\
          &       & BLEU-1 & 0.3799 & 0.2557 & 0.5170 & 0.3980 & 0.3798 & 0.2295 & 0.6261 & 0.4759 & 0.3477 & \cellcolor{TopOne} 0.7464 & \cellcolor{TopTwo} 0.7068 & 0.7027 & 0.6810 \\
          &       & EM    & 0.2374 & 0.0742 & 0.3858 & 0.2493 & 0.2374 & 0.1009 & 0.5134 & 0.3650 & 0.2107 & \cellcolor{TopTwo} 0.6024 & 0.5846 & \cellcolor{TopOne} 0.6053 & 0.5519 \\
          &       & LLM-Judge & 0.7092 & 0.6424 & 0.8353 & 0.7626 & 0.7033 & 0.7092 & 0.8220 & 0.7774 & 0.7107 & \cellcolor{TopOne} 0.9021 & 0.8620 & \cellcolor{TopTwo} 0.8887 & 0.8754 \\        
    \midrule
    \multirow{12}[6]{*}{\rotatebox{90}{Reasoning}}  & \multirow{4}[2]{*}{TR} & F1    & 0.2545 & 0.1484 & 0.4088 & 0.2017 & 0.2553 & \cellcolor{TopTwo} 0.5661 & 0.5235 & 0.5598 & 0.2294 & 0.5439 & 0.5233 & 0.5199 & \cellcolor{TopOne} 0.5779 \\
          &       & BLEU-1 & 0.2363 & 0.1305 & 0.3872 & 0.1836 & 0.2363 & \cellcolor{TopTwo} 0.5326 & 0.4959 & 0.5306 & 0.2065 & 0.5211 & 0.4955 & 0.4952 & \cellcolor{TopOne} 0.5475 \\
          &       & EM    & 0.1545 & 0.0732 & 0.3089 & 0.1138 & 0.1545 & 0.3496 & 0.4065 & 0.3902 & 0.1382 & 0.4146 & 0.3740 & \cellcolor{TopTwo} 0.3984 & \cellcolor{TopOne} 0.4309 \\
          &       & LLM-Judge & 0.2805 & 0.1341 & 0.5325 & 0.2276 & 0.2764 & \cellcolor{TopOne} 0.8008 & 0.6626 & 0.7276 & 0.2480 & 0.6707 & 0.6870 & 0.6463 & \cellcolor{TopTwo} 0.7317 \\
\cmidrule{2-16}          & \multirow{4}[2]{*}{VR} & F1    & 0.2552 & 0.0900 & 0.2711 & 0.1669 & 0.2594 & \cellcolor{TopOne} 0.4593 & 0.3963 & 0.4243 & 0.2015 & \cellcolor{TopTwo} 0.4390 & 0.4205 & 0.3951 & 0.3950 \\
          &       & BLEU-1 & 0.2442 & 0.0758 & 0.2548 & 0.1536 & 0.2480 & \cellcolor{TopOne} 0.4459 & 0.3797 & 0.4108 & 0.1912 & \cellcolor{TopTwo} 0.4175 & 0.3969 & 0.3767 & 0.3795 \\
          &       & EM    & 0.2011 & 0.0460 & 0.1552 & 0.0977 & 0.2011 & \cellcolor{TopOne} 0.3851 & 0.2989 & \cellcolor{TopTwo} 0.3391 & 0.1609 & 0.3276 & 0.3046 & 0.3103 & 0.3103 \\
          &       & LLM-Judge & 0.3046 & 0.1264 & 0.2931 & 0.1954 & 0.3046 & \cellcolor{TopOne} 0.6149 & 0.5000 & 0.5489 & 0.2586 & \cellcolor{TopTwo} 0.5632 & \cellcolor{TopTwo} 0.5632 & 0.4598 & 0.4885 \\
\cmidrule{2-16}          & \multirow{4}[2]{*}{MR} & F1    & 0.2411 & 0.1516 & 0.4056 & 0.2116 & 0.2428 & 0.4367 & 0.4503 & 0.4385 & 0.2101 & \cellcolor{TopOne} 0.4980 & \cellcolor{TopTwo} 0.4954 & 0.4343 & 0.4756 \\
          &       & BLEU-1 & 0.1739 & 0.0982 & 0.3012 & 0.1497 & 0.1770 & 0.3347 & 0.3476 & 0.3355 & 0.1429 & \cellcolor{TopOne} 0.3915 & \cellcolor{TopTwo} 0.3871 & 0.3295 & 0.3667 \\
          &       & EM    & 0.0340 & 0.0049 & 0.0631 & 0.0194 & 0.0340 & 0.0631 & 0.0777 & 0.0680 & 0.0194 & \cellcolor{TopOne} 0.0971 & \cellcolor{TopOne} 0.0971 & 0.0631 & \cellcolor{TopTwo} 0.0874 \\
          &       & LLM-Judge & 0.2985 & 0.1238 & 0.6917 & 0.2767 & 0.3058 & \cellcolor{TopTwo} 0.8204 & 0.7379 & 0.7961 & 0.2791 & \cellcolor{TopOne} 0.8228 & \cellcolor{TopTwo} 0.8204 & 0.7063 & 0.7888 \\
    \midrule
    \multirow{12}[6]{*}{\rotatebox{90}{Knowledge Management}} & \multirow{4}[2]{*}{KR} & F1    & 0.2354 & 0.1805 & 0.3331 & 0.2429 & 0.2354 & \cellcolor{TopTwo} 0.4292 & 0.3863 & \cellcolor{TopOne} 0.4686 & 0.2181 & 0.4176 & 0.4062 & 0.3835 & 0.3707 \\
          &       & BLEU-1 & 0.2005 & 0.1521 & 0.2798 & 0.2044 & 0.2000 & \cellcolor{TopTwo} 0.3741 & 0.3305 & \cellcolor{TopOne} 0.4087 & 0.1883 & 0.3660 & 0.3606 & 0.3346 & 0.3152 \\
          &       & EM    & 0.1235 & 0.0988 & 0.1358 & 0.0988 & 0.1235 & \cellcolor{TopTwo} 0.2099 & 0.1975 & \cellcolor{TopOne} 0.2222 & 0.1235 & \cellcolor{TopTwo} 0.2099 & 0.1975 & \cellcolor{TopTwo} 0.2099 & 0.1481 \\
          &       & LLM-Judge & 0.3395 & 0.2160 & 0.5988 & 0.3210 & 0.3395 & 0.7407 & 0.5988 & \cellcolor{TopOne} 0.7593 & 0.2840 & \cellcolor{TopTwo} 0.7469 & 0.6852 & 0.6358 & 0.6173 \\
\cmidrule{2-16}          & \multirow{4}[2]{*}{CD} & F1    & \cellcolor{TopTwo} 0.3457 & \cellcolor{TopOne} 0.3580 & 0.3333 & \cellcolor{TopTwo} 0.3457 & 0.3333 & \cellcolor{TopOne} 0.3580 & \cellcolor{TopTwo} 0.3457 & \cellcolor{TopTwo} 0.3457 & \cellcolor{TopOne} 0.3580 & \cellcolor{TopOne} 0.3580 & \cellcolor{TopTwo} 0.3457 & \cellcolor{TopTwo} 0.3457 & 0.3210 \\
          &       & BLEU-1 & \cellcolor{TopTwo} 0.3457 & \cellcolor{TopOne} 0.3580 & 0.3333 & \cellcolor{TopTwo} 0.3457 & 0.3333 & \cellcolor{TopOne} 0.3580 & \cellcolor{TopTwo} 0.3457 & \cellcolor{TopTwo} 0.3457 & \cellcolor{TopOne} 0.3580 & \cellcolor{TopOne} 0.3580 & \cellcolor{TopTwo} 0.3457 & \cellcolor{TopTwo} 0.3457 & 0.3210 \\
          &       & EM    & \cellcolor{TopTwo} 0.3457 & \cellcolor{TopOne} 0.3580 & 0.3333 & \cellcolor{TopTwo} 0.3457 & 0.3333 & \cellcolor{TopOne} 0.3580 & \cellcolor{TopTwo} 0.3457 & \cellcolor{TopTwo} 0.3457 & \cellcolor{TopOne} 0.3580 & \cellcolor{TopOne} 0.3580 & \cellcolor{TopTwo} 0.3457 & \cellcolor{TopTwo} 0.3457 & 0.3210 \\
          &       & LLM-Judge & \cellcolor{TopTwo} 0.3457 & \cellcolor{TopOne} 0.3580 & 0.3333 &  \cellcolor{TopTwo} 0.3457 & 0.3333 & \cellcolor{TopOne} 0.3580 & \cellcolor{TopTwo} 0.3457 & \cellcolor{TopTwo} 0.3457 &\cellcolor{TopOne} 0.3580 & \cellcolor{TopOne} 0.3580 & \cellcolor{TopTwo} 0.3457 & \cellcolor{TopTwo} 0.3457 & 0.3210 \\
\cmidrule{2-16}    & \multirow{4}[2]{*}{AR} & F1    & \cellcolor{TopTwo} 0.9958 & \cellcolor{TopOne} 1.0000 & 0.9581 & 0.9891 & \cellcolor{TopTwo} 0.9958 & 0.9849 & 0.9159 & 0.9785 & 0.9946 & 0.9532 & 0.9421 & 0.9749 & 0.9461 \\
          &       & BLEU-1 & \cellcolor{TopTwo} 0.9953 & \cellcolor{TopOne} 1.0000 & 0.9575 & 0.9891 & \cellcolor{TopTwo} 0.9953 & 0.9844 & 0.9141 & 0.9784 & 0.9946 & 0.9524 & 0.9413 & 0.9741 & 0.9459 \\
          &       & EM    & \cellcolor{TopTwo} 0.9946 & \cellcolor{TopOne} 1.0000 & 0.9565 & 0.9891 & \cellcolor{TopTwo} 0.9946 & 0.9837 & 0.9130 & 0.9783 & \cellcolor{TopTwo} 0.9946 & 0.9511 & 0.9402 & 0.9728 & 0.9457 \\
          &       & LLM-Judge & \cellcolor{TopTwo} 0.9783 & \cellcolor{TopOne} 0.9837 & 0.9402 & 0.9728 & \cellcolor{TopTwo} 0.9783 & 0.9674 & 0.9375 & 0.9620 & \cellcolor{TopOne} 0.9837 & 0.9429 & 0.9375 & 0.9565 & 0.9429 \\      
    \midrule
    \multicolumn{2}{c}{\multirow{4}[2]{*}{Overall}} & F1    & 0.3625 & 0.2558 & 0.5478 & 0.3560 & 0.3619 & 0.5282 & 0.6038 & 0.6021 & 0.3354 & \cellcolor{TopOne} 0.6791 & \cellcolor{TopTwo} 0.6688 & 0.6322 & 0.6457 \\
    \multicolumn{2}{c}{} & BLEU-1 & 0.3279 & 0.2255 & 0.4946 & 0.3164 & 0.3279 & 0.4792 & 0.5475 & 0.5483 & 0.2999 & \cellcolor{TopOne} 0.6288 & \cellcolor{TopTwo} 0.6164 & 0.5817 & 0.5887 \\
    \multicolumn{2}{c}{} & EM    & 0.2519 & 0.1596 & 0.3711 & 0.2350 & 0.2507 & 0.3402 & 0.4296 & 0.4208 & 0.2279 & \cellcolor{TopOne} 0.4915 & \cellcolor{TopTwo} 0.4798 & 0.4711 & 0.4611 \\
    \multicolumn{2}{c}{} & LLM-Judge & 0.4331 & 0.3115 & 0.6715 & 0.4299 & 0.4307 & 0.7306 & 0.7119 & 0.7463 & 0.4129 & \cellcolor{TopOne} 0.7957 & \cellcolor{TopTwo} 0.7823 & 0.7329 & 0.7586 \\
    \bottomrule
    \end{tabular}%
    }
  \label{tab:result_k_5}%
\end{table*}%

\begin{table*}[t!]
  \centering
  \caption{Benchmark performance results on Qwen-2.5-VL-7B with the retrieval number $K$=15. The best and second-performed memory model(s) are highlighted with \textcolor{orange}{orange} and \textcolor{blue}{blue} backgrounds.}
  \resizebox{\linewidth}{!}{%
    \begin{tabular}{ccc|cccccccc|ccccc}
    \toprule
    \rowcolor{qwenbar} \multicolumn{3}{c|}{\qwenlogo\ Qwen-2.5-VL-7B (K=15)} & Full (Text) & FIFO  & NaiveRAG & Gen. Agent & Reflexion & MemGPT & A-Mem & MemoryOS & Full (MM) & MuRAG & UniversalRAG & NGM   & AUGUSTUS \\
    \midrule
    \multirow{12}[6]{*}{\rotatebox{90}{Extract. \& Adapt.}} & \multirow{4}[2]{*}{FR} & F1    & 0.2376 & 0.1511 & 0.6123 & 0.2765 & 0.2391 & 0.5928 & 0.6258 & 0.6382 & 0.2150 & \cellcolor{TopOne} 0.6674 & \cellcolor{TopTwo} 0.6626 & 0.6482 & 0.6182 \\
          &       & BLEU-1 & 0.1865 & 0.1071 & 0.5273 & 0.2124 & 0.1903 & 0.5098 & 0.5321 & 0.5462 & 0.1626 & \cellcolor{TopOne} 0.5808 & 0.5678 & \cellcolor{TopTwo} 0.5698 & 0.5319 \\
          &       & EM    & 0.0913 & 0.0502 & 0.3151 & 0.0913 & 0.0913 & 0.3288 & 0.3105 & 0.3516 & 0.0685 & \cellcolor{TopOne} 0.3744 & 0.3379 & \cellcolor{TopTwo} 0.3699 & 0.3288 \\
          &       & LLM-Judge & 0.2626 & 0.1416 & 0.8082 & 0.3379 & 0.2626 & 0.8539 & 0.8037 & 0.8447 & 0.2260 & \cellcolor{TopOne} 0.8767 & \cellcolor{TopTwo} 0.8721 & 0.8151 & 0.8174 \\
\cmidrule{2-16}          & \multirow{4}[2]{*}{VS} & F1    & 0.1992 & 0.0794 & 0.7812 & 0.3425 & 0.1954 & 0.6239 & 0.7905 & 0.8022 & 0.1658 & \cellcolor{TopOne} 0.8725 & 0.8452 & \cellcolor{TopTwo} 0.8673 & 0.8607 \\
          &       & BLEU-1 & 0.1873 & 0.0680 & 0.7359 & 0.3032 & 0.1840 & 0.5834 & 0.7160 & 0.7337 & 0.1473 & \cellcolor{TopOne} 0.8362 & 0.8162 & \cellcolor{TopTwo} 0.8295 & 0.8084 \\
          &       & EM    & 0.1601 & 0.0425 & 0.5752 & 0.2320 & 0.1569 & 0.4118 & 0.5915 & 0.6111 & 0.1078 & 0.6536 & 0.6471 & \cellcolor{TopOne} 0.6961 & \cellcolor{TopTwo} 0.6569 \\
          &       & LLM-Judge & 0.1961 & 0.0719 & 0.7696 & 0.3366 & 0.1895 & 0.5964 & 0.7484 & 0.7958 & 0.1683 & \cellcolor{TopOne} 0.8742 & 0.8448 & \cellcolor{TopTwo} 0.8676 & 0.8497 \\
\cmidrule{2-16}       & \multirow{4}[2]{*}{TTL} & F1    & 0.4500 & 0.3410 & 0.6410 & 0.4981 & 0.4486 & 0.2924 & 0.5900 & 0.5302 & 0.4147 & \cellcolor{TopOne} 0.8071 & 0.7517 & 0.7782 & \cellcolor{TopTwo} 0.7975 \\
          &       & BLEU-1 & 0.3799 & 0.2838 & 0.5694 & 0.4283 & 0.3798 & 0.2295 & 0.5096 & 0.4502 & 0.3477 & \cellcolor{TopOne} 0.7307 & 0.6785 & 0.7128 & \cellcolor{TopTwo} 0.7291 \\
          &       & EM    & 0.2374 & 0.1662 & 0.4540 & 0.2908 & 0.2374 & 0.1009 & 0.3769 & 0.3264 & 0.2107 & 0.5964 & 0.5697 & \cellcolor{TopTwo} 0.5994 & \cellcolor{TopOne} 0.6172 \\
          &       & LLM-Judge & 0.7092 & 0.6810 & 0.8279 & 0.7611 & 0.7033 & 0.7092 & 0.7953 & 0.7582 & 0.7107 & \cellcolor{TopTwo} 0.9050 & 0.8591 & \cellcolor{TopTwo} 0.9050 & \cellcolor{TopOne} 0.9065 \\  
    \midrule
    \multirow{12}[6]{*}{\rotatebox{90}{Reasoning}}  & \multirow{4}[2]{*}{TR} & F1    & 0.2545 & 0.1930 & 0.4864 & 0.3335 & 0.2553 & 0.5661 & 0.5760 & 0.5614 & 0.2294 & \cellcolor{TopTwo} 0.5798 & 0.5759 & 0.5752 & \cellcolor{TopOne} 0.5882 \\
          &       & BLEU-1 & 0.2363 & 0.1684 & 0.4539 & 0.3122 & 0.2363 & 0.5326 & \cellcolor{TopTwo} 0.5485 & 0.5323 & 0.2065 & 0.5482 & 0.5413 & 0.5468 & \cellcolor{TopOne} 0.5602 \\
          &       & EM    & 0.1545 & 0.0976 & 0.3252 & 0.2276 & 0.1545 & 0.3496 & \cellcolor{TopTwo} 0.4309 & 0.3902 & 0.1382 & 0.4228 & 0.4065 & \cellcolor{TopOne} 0.4472 & \cellcolor{TopOne} 0.4472 \\
          &       & LLM-Judge & 0.2805 & 0.1870 & 0.6707 & 0.3943 & 0.2764 & 0.8008 & 0.7236 & 0.7358 & 0.2480 & \cellcolor{TopTwo} 0.7724 & \cellcolor{TopOne} 0.8008 & 0.7358 & 0.7561 \\
\cmidrule{2-16}          & \multirow{4}[2]{*}{VR} & F1    & 0.2552 & 0.1155 & 0.3303 & 0.2119 & 0.2594 & 0.4593 & 0.4458 & 0.4676 & 0.2015 & 0.4678 & \cellcolor{TopOne} 0.4913 & \cellcolor{TopTwo} 0.4697 & 0.3860 \\
          &       & BLEU-1 & 0.2442 & 0.0998 & 0.3144 & 0.2005 & 0.2480 & 0.4459 & 0.4335 & \cellcolor{TopTwo} 0.4533 & 0.1912 & 0.4474 & \cellcolor{TopOne} 0.4712 & 0.4494 & 0.3713 \\
          &       & EM    & 0.2011 & 0.0690 & 0.2299 & 0.1379 & 0.2011 & \cellcolor{TopOne} 0.3851 & 0.3563 & 0.3678 & 0.1609 & 0.3563 & \cellcolor{TopTwo} 0.3736 & 0.3621 & 0.2874 \\
          &       & LLM-Judge & 0.3046 & 0.1494 & 0.4368 & 0.2644 & 0.3046 & \cellcolor{TopOne} 0.6149 & 0.5632 & \cellcolor{TopTwo} 0.5862 & 0.2586 & 0.5690 & 0.5833 & 0.5632 & 0.4799 \\
\cmidrule{2-16}        & \multirow{4}[2]{*}{MR} & F1    & 0.2411 & 0.1747 & 0.4549 & 0.2471 & 0.2428 & 0.4367 & 0.4723 & 0.4547 & 0.2101 & \cellcolor{TopOne} 0.5042 & \cellcolor{TopTwo} 0.5004 & 0.4841 & 0.4839 \\
          &       & BLEU-1 & 0.1739 & 0.1225 & 0.3372 & 0.1813 & 0.1770 & 0.3347 & 0.3698 & 0.3519 & 0.1429 & \cellcolor{TopOne} 0.3910 & \cellcolor{TopTwo} 0.3840 & 0.3760 & 0.3748 \\
          &       & EM    & 0.0340 & 0.0146 & 0.0777 & 0.0243 & 0.0340 & 0.0631 & 0.0825 & 0.0825 & 0.0194 & \cellcolor{TopTwo} 0.0922 & \cellcolor{TopOne} 0.0971 & 0.0874 & 0.0874 \\
          &       & LLM-Judge & 0.2985 & 0.1893 & 0.8058 & 0.3811 & 0.3058 & 0.8204 & 0.7913 & 0.8301 & 0.2791 & \cellcolor{TopTwo} 0.8568 & \cellcolor{TopOne} 0.8665 & 0.8010 & 0.8252 \\  
    \midrule
    \multirow{12}[6]{*}{\rotatebox{90}{Knowledge Management}} & \multirow{4}[2]{*}{KR} & F1    & 0.2354 & 0.1685 & 0.3740 & 0.2752 & 0.2354 & 0.4292 & 0.4470 & \cellcolor{TopOne} 0.4888 & 0.2181 & \cellcolor{TopTwo} 0.4701 & 0.4629 & 0.4242 & 0.4175 \\
          &       & BLEU-1 & 0.2005 & 0.1290 & 0.3203 & 0.2338 & 0.2000 & 0.3741 & 0.3935 & \cellcolor{TopOne} 0.4345 & 0.1883 & \cellcolor{TopTwo} 0.4166 & 0.4052 & 0.3738 & 0.3508 \\
          &       & EM    & 0.1235 & 0.0617 & 0.1605 & 0.1235 & 0.1235 & 0.2099 & 0.2222 & \cellcolor{TopOne} 0.2593 & 0.1235 & \cellcolor{TopTwo} 0.2469 & 0.2222 & 0.2222 & 0.1975 \\
          &       & LLM-Judge & 0.3395 & 0.2469 & 0.6296 & 0.3704 & 0.3395 & 0.7407 & 0.6605 & 0.7654 & 0.2840 & \cellcolor{TopOne} 0.7963 & \cellcolor{TopTwo} 0.7716 & 0.7284 & 0.7099 \\
\cmidrule{2-16}          & \multirow{4}[2]{*}{CD} & F1    & 0.3457 & \cellcolor{TopTwo} 0.3580 & 0.3457 & 0.3086 & 0.3333 & \cellcolor{TopTwo} 0.3580 & 0.3210 & \cellcolor{TopTwo} 0.3580 & \cellcolor{TopTwo} 0.3580 & \cellcolor{TopOne} 0.3704 & \cellcolor{TopOne} 0.3704 & 0.3457 & 0.3210 \\
          &       & BLEU-1 & 0.3457 & \cellcolor{TopTwo} 0.3580 & 0.3457 & 0.3086 & 0.3333 & \cellcolor{TopTwo} 0.3580 & 0.3210 & \cellcolor{TopTwo} 0.3580 & \cellcolor{TopTwo} 0.3580 & \cellcolor{TopOne} 0.3704 & \cellcolor{TopOne} 0.3704 & 0.3457 & 0.3210 \\
          &       & EM    & 0.3457 & \cellcolor{TopTwo} 0.3580 & 0.3457 & 0.3086 & 0.3333 & \cellcolor{TopTwo} 0.3580 & 0.3210 & \cellcolor{TopTwo} 0.3580 & \cellcolor{TopTwo} 0.3580 & \cellcolor{TopOne} 0.3704 & \cellcolor{TopOne} 0.3704 & 0.3457 & 0.3210 \\
          &       & LLM-Judge & 0.3457 & \cellcolor{TopTwo} 0.3580 & 0.3457 & 0.3086 & 0.3333 & \cellcolor{TopTwo} 0.3580 & 0.3210 & \cellcolor{TopTwo} 0.3580 & \cellcolor{TopTwo} 0.3580 & \cellcolor{TopOne} 0.3704 & \cellcolor{TopOne} 0.3704 & 0.3457 & 0.3210 \\
\cmidrule{2-16}      & \multirow{4}[2]{*}{AR} & F1    & \cellcolor{TopOne} 0.9958 & \cellcolor{TopOne} 0.9958 & 0.9513 & 0.9843 & \cellcolor{TopOne} 0.9958 & 0.9849 & 0.9224 & 0.9842 & \cellcolor{TopTwo} 0.9946 & 0.9513 & 0.9523 & 0.9635 & 0.9464 \\
          &       & BLEU-1 & \cellcolor{TopOne} 0.9953 & \cellcolor{TopOne} 0.9953 & 0.9512 & 0.9840 & \cellcolor{TopOne} 0.9953 & 0.9844 & 0.9201 & 0.9840 & \cellcolor{TopTwo} 0.9946 & 0.9512 & 0.9518 & 0.9629 & 0.9461 \\
          &       & EM    & \cellcolor{TopOne} 0.9946 & \cellcolor{TopOne} 0.9946 & 0.9511 & \cellcolor{TopTwo} 0.9837 & \cellcolor{TopOne} 0.9946 & \cellcolor{TopTwo} 0.9837 & 0.9185 & \cellcolor{TopTwo} 0.9837 & \cellcolor{TopOne} 0.9946 & 0.9511 & 0.9511 & 0.9620 & 0.9457 \\
          &       & LLM-Judge & \cellcolor{TopTwo} 0.9783 & \cellcolor{TopTwo} 0.9783 & 0.9375 & 0.9674 & \cellcolor{TopTwo} 0.9783 & 0.9674 & 0.9429 & 0.9674 & \cellcolor{TopOne} 0.9837 & 0.9511 & 0.9348 & 0.9457 & 0.9375 \\    
    \midrule
    \multicolumn{2}{c}{\multirow{4}[2]{*}{Overall}} & F1    & 0.3625 & 0.2794 & 0.6040 & 0.4035 & 0.3619 & 0.5282 & 0.6168 & 0.6182 & 0.3354 & \cellcolor{TopOne} 0.6925 & 0.6775 & \cellcolor{TopTwo} 0.6788 & 0.6667 \\
    \multicolumn{2}{c}{} & BLEU-1 & 0.3279 & 0.2489 & 0.5503 & 0.3619 & 0.3279 & 0.4792 & 0.5574 & 0.5599 & 0.2999 & \cellcolor{TopOne} 0.6393 & 0.6244 & \cellcolor{TopTwo} 0.6296 & 0.6129 \\
    \multicolumn{2}{c}{} & EM    & 0.2519 & 0.1894 & 0.4150 & 0.2700 & 0.2507 & 0.3402 & 0.4214 & 0.4290 & 0.2279 & \cellcolor{TopTwo} 0.4915 & 0.4804 & \cellcolor{TopOne} 0.4997 & 0.4793 \\
    \multicolumn{2}{c}{} & LLM-Judge & 0.4331 & 0.3504 & 0.7408 & 0.4906 & 0.4307 & 0.7306 & 0.7458 & 0.7694 & 0.4129 & \cellcolor{TopOne} 0.8209 & \cellcolor{TopTwo} 0.8077 & 0.7969 & 0.7873 \\
    \bottomrule
    \end{tabular}%
    }
  \label{tab:k_15}%
\end{table*}%

\begin{table*}[t!]
  \centering
  \caption{Benchmark performance results on Qwen-2.5-VL-7B with the retrieval number $K$=20. The best and second-performed memory model(s) are highlighted with \textcolor{orange}{orange} and \textcolor{blue}{blue} backgrounds.}
  \resizebox{\linewidth}{!}{%
    \begin{tabular}{ccc|cccccccc|ccccc}
    \toprule
    \rowcolor{qwenbar} \multicolumn{3}{c|}{\qwenlogo\ Qwen-2.5-VL-7B (K=20)} & Full (Text) & FIFO  & NaiveRAG & Gen. Agent & Reflexion & MemGPT & A-Mem & MemoryOS & Full (MM) & MuRAG & UniversalRAG & NGM   & AUGUSTUS \\
    \midrule
    \multirow{12}[6]{*}{\rotatebox{90}{Extract. \& Adapt.}} & \multirow{4}[2]{*}{FR} & F1    & 0.2376 & 0.1743 & 0.6303 & 0.3020 & 0.2391 & 0.5928 & 0.6145 & 0.6265 & 0.2150 & \cellcolor{TopTwo} 0.6596 & \cellcolor{TopOne} 0.6609 & 0.6589 & 0.6216 \\
          &       & BLEU-1 & 0.1865 & 0.1295 & 0.5438 & 0.2366 & 0.1903 & 0.5098 & 0.5210 & 0.5396 & 0.1626 & \cellcolor{TopTwo} 0.5656 & 0.5639 & \cellcolor{TopOne} 0.5813 & 0.5335 \\
          &       & EM    & 0.0913 & 0.0548 & 0.3242 & 0.1142 & 0.0913 & 0.3288 & 0.2922 & 0.3379 & 0.0685 & \cellcolor{TopTwo} 0.3516 & 0.3425 & \cellcolor{TopOne} 0.3790 & 0.3379 \\
          &       & LLM-Judge & 0.2626 & 0.1598 & 0.8174 & 0.3744 & 0.2626 & 0.8539 & 0.8174 & 0.8288 & 0.2260 & \cellcolor{TopOne} 0.8950 & \cellcolor{TopTwo} 0.8790 & 0.8402 & 0.8128 \\
\cmidrule{2-16}          & \multirow{4}[2]{*}{VS} & F1    & 0.1992 & 0.0953 & 0.8144 & 0.3567 & 0.1954 & 0.6239 & 0.7782 & 0.7963 & 0.1658 & \cellcolor{TopOne} 0.8676 & 0.8444 & 0.8583 & \cellcolor{TopTwo} 0.8647 \\
          &       & BLEU-1 & 0.1873 & 0.0849 & 0.7704 & 0.3187 & 0.1840 & 0.5834 & 0.7026 & 0.7300 & 0.1473 & \cellcolor{TopOne}  0.8341 & 0.8160 & 0.8198 & \cellcolor{TopTwo} 0.8230 \\
          &       & EM    & 0.1601 & 0.0588 & 0.5980 & 0.2386 & 0.1569 & 0.4118 & 0.5882 & 0.6078 & 0.1078 & 0.6503 & 0.6503 & \cellcolor{TopOne} 0.6765 & \cellcolor{TopTwo} 0.6569 \\
          &       & LLM-Judge & 0.1961 & 0.0850 & 0.8023 & 0.3546 & 0.1895 & 0.5964 & 0.7337 & 0.7925 & 0.1683 & \cellcolor{TopOne} 0.8856 & 0.8611 & \cellcolor{TopTwo} 0.8660 & 0.8611 \\
\cmidrule{2-16}      & \multirow{4}[2]{*}{TTL} & F1    & 0.4500 & 0.3789 & 0.6104 & 0.4900 & 0.4486 & 0.2924 & 0.5542 & 0.5353 & 0.4147 & 0.7841 & 0.7211 & \cellcolor{TopOne} 0.7911 & \cellcolor{TopTwo} 0.7859 \\
          &       & BLEU-1 & 0.3799 & 0.3180 & 0.5355 & 0.4137 & 0.3798 & 0.2295 & 0.4735 & 0.4566 & 0.3477 & 0.7122 & 0.6478 & \cellcolor{TopOne} 0.7250 & \cellcolor{TopTwo} 0.7184 \\
          &       & EM    & 0.2374 & 0.1899 & 0.4243 & 0.2849 & 0.2374 & 0.1009 & 0.3561 & 0.3442 & 0.2107 & 0.5905 & 0.5312 & \cellcolor{TopTwo} 0.6142 & \cellcolor{TopOne} 0.6172 \\
          &       & LLM-Judge & 0.7092 & 0.6929 & 0.8131 & 0.7448 & 0.7033 & 0.7092 & 0.7611 & 0.7760 & 0.7107 & 0.8961 & 0.8694 & \cellcolor{TopTwo} 0.8976 & \cellcolor{TopOne} 0.9021 \\    
    \midrule
    \multirow{12}[6]{*}{\rotatebox{90}{Reasoning}} & \multirow{4}[2]{*}{TR} & F1    & 0.2545 & 0.1858 & 0.5154 & 0.3484 & 0.2553 & 0.5661 & 0.5502 & \cellcolor{TopOne} 0.5941 & 0.2294 & 0.5642 & 0.5613 & 0.5445 & \cellcolor{TopTwo} 0.5890 \\
          &       & BLEU-1 & 0.2363 & 0.1652 & 0.4848 & 0.3255 & 0.2363 & 0.5326 & 0.5220 & \cellcolor{TopOne} 0.5679 & 0.2065 & 0.5345 & 0.5291 & 0.5180 &  \cellcolor{TopTwo} 0.5615 \\
          &       & EM    & 0.1545 & 0.0976 & 0.3577 & 0.2358 & 0.1545 & 0.3496 & 0.3821 & \cellcolor{TopOne} 0.4472 & 0.1382 & 0.3984 & 0.3984 & 0.4146 & \cellcolor{TopTwo} 0.4390 \\
          &       & LLM-Judge & 0.2805 & 0.1870 & 0.7073 & 0.4268 & 0.2764 & \cellcolor{TopOne} 0.8008 & 0.7073 & 0.7398 & 0.2480 & \cellcolor{TopTwo} 0.7967 & 0.7846 & 0.7073 & 0.7642 \\
\cmidrule{2-16}          & \multirow{4}[2]{*}{VR} & F1    & 0.2552 & 0.1383 & 0.3654 & 0.2171 & 0.2594 & 0.4593 & 0.4216 & 0.4410 & 0.2015 & \cellcolor{TopOne} 0.5320 & 0.4344 &\cellcolor{TopTwo}  0.4772 & 0.3695 \\
          &       & BLEU-1 & 0.2442 & 0.1208 & 0.3507 & 0.2016 & 0.2480 & 0.4459 & 0.4096 & 0.4304 & 0.1912 & \cellcolor{TopOne} 0.5130 & 0.4192 & \cellcolor{TopTwo} 0.4563 & 0.3550 \\
          &       & EM    & 0.2011 & 0.0862 & 0.2644 & 0.1264 & 0.2011 & \cellcolor{TopTwo} 0.3851 & 0.3391 & 0.3621 & 0.1609 & \cellcolor{TopOne} 0.4138 & 0.3333 & 0.3678 & 0.2644 \\
          &       & LLM-Judge & 0.3046 & 0.1954 & 0.4770 & 0.2672 & 0.3046 & \cellcolor{TopTwo} 0.6149 & 0.5517 & 0.5891 & 0.2586 & \cellcolor{TopOne} 0.6322 & 0.5460 & 0.5718 & 0.4483 \\
\cmidrule{2-16}          & \multirow{4}[2]{*}{MR} & F1    & 0.2411 & 0.1837 & 0.4678 & 0.2616 & 0.2428 & 0.4367 & 0.4732 & 0.4554 & 0.2101 & \cellcolor{TopOne} 0.4988 & \cellcolor{TopTwo} 0.4955 & 0.4802 & 0.4742 \\
          &       & BLEU-1 & 0.1739 & 0.1271 & 0.3523 & 0.1932 & 0.1770 & 0.3347 & 0.3656 & 0.3550 & 0.1429 & \cellcolor{TopOne} 0.3863 & \cellcolor{TopTwo} 0.3757 & 0.3664 & 0.3683 \\
          &       & EM    & 0.0340 & 0.0194 & \cellcolor{TopTwo} 0.0922 & 0.0291 & 0.0340 & 0.0631 & 0.0874 & 0.0777 & 0.0194 & 0.0874 & 0.0874 & 0.0825 & \cellcolor{TopOne} 0.0971 \\
          &       & LLM-Judge & 0.2985 & 0.2039 & 0.8107 & 0.4029 & 0.3058 & 0.8204 & 0.7913 & 0.8374 & 0.2791 & \cellcolor{TopOne} 0.8786 & \cellcolor{TopTwo} 0.8689 & 0.8107 & 0.8155 \\
    \midrule
    \multirow{12}[6]{*}{\rotatebox{90}{Knowledge Management}}  & \multirow{4}[2]{*}{KR} & F1    & 0.2354 & 0.1924 & 0.4200 & 0.2909 & 0.2354 & 0.4292 & 0.4290 & 0.4597 & 0.2181 & \cellcolor{TopOne} 0.4648 & \cellcolor{TopTwo} 0.4604 & 0.4354 & 0.3967 \\
          &       & BLEU-1 & 0.2005 & 0.1546 & 0.3666 & 0.2530 & 0.2000 & 0.3741 & 0.3657 & \cellcolor{TopOne} 0.4083 & 0.1883 & \cellcolor{TopTwo} 0.4060 & 0.3969 & 0.3858 & 0.3343 \\
          &       & EM    & 0.1235 & 0.0741 & 0.2099 & 0.1481 & 0.1235 & 0.2099 & 0.1852 & 0.2222 & 0.1235 & \cellcolor{TopTwo} 0.2346 & 0.1975 & \cellcolor{TopOne} 0.2593 & 0.1728 \\
          &       & LLM-Judge & 0.3395 & 0.2593 & 0.7037 & 0.3951 & 0.3395 & 0.7407 & 0.7099 & 0.7531 & 0.2840 & \cellcolor{TopOne} 0.8025 & \cellcolor{TopTwo} 0.7778 & 0.7593 & 0.6975 \\
\cmidrule{2-16}          & \multirow{4}[2]{*}{CD} & F1    & 0.3457 & \cellcolor{TopTwo} 0.3580 & 0.3457 & 0.3086 & 0.3333 & \cellcolor{TopTwo} 0.3580 & 0.3333 & \cellcolor{TopTwo} 0.3580 & \cellcolor{TopTwo} 0.3580 & \cellcolor{TopOne} 0.3704 & 0.3457 & \cellcolor{TopTwo} 0.3580 & 0.3333 \\
          &       & BLEU-1 & 0.3457 & \cellcolor{TopTwo} 0.3580 & 0.3457 & 0.3086 & 0.3333 & \cellcolor{TopTwo} 0.3580 & 0.3333 & \cellcolor{TopTwo} 0.3580 & \cellcolor{TopTwo} 0.3580 & \cellcolor{TopOne} 0.3704 & 0.3457 & \cellcolor{TopTwo} 0.3580 & 0.3333 \\
          &       & EM    & 0.3457 & \cellcolor{TopTwo} 0.3580 & 0.3457 & 0.3086 & 0.3333 & \cellcolor{TopTwo} 0.3580 & 0.3333 & \cellcolor{TopTwo} 0.3580 & \cellcolor{TopTwo} 0.3580 & \cellcolor{TopOne} 0.3704 & 0.3457 & \cellcolor{TopTwo} 0.3580 & 0.3333 \\
          &       & LLM-Judge & 0.3457 & \cellcolor{TopTwo} 0.3580 & 0.3457 & 0.3086 & 0.3333 & \cellcolor{TopTwo} 0.3580 & 0.3333 & \cellcolor{TopTwo} 0.3580 & \cellcolor{TopTwo} 0.3580 & \cellcolor{TopOne} 0.3704 & 0.3457 & \cellcolor{TopTwo} 0.3580 & 0.3333 \\
\cmidrule{2-16}     & \multirow{4}[2]{*}{AR} & F1    & \cellcolor{TopOne} 0.9958 & \cellcolor{TopTwo} 0.9956 & 0.9514 & 0.9840 & \cellcolor{TopOne} 0.9958 & 0.9849 & 0.9233 & 0.9677 & 0.9946 & 0.9307 & 0.9322 & 0.9471 & 0.9519 \\
          &       & BLEU-1 & \cellcolor{TopOne} 0.9953 & \cellcolor{TopTwo} 0.9952 & 0.9512 & 0.9838 & \cellcolor{TopOne} 0.9953 & 0.9844 & 0.9210 & 0.9676 & 0.9946 & 0.9297 & 0.9306 & 0.9466 & 0.9515 \\
          &       & EM    & \cellcolor{TopOne} 0.9946 & \cellcolor{TopOne} 0.9946 & 0.9511 & \cellcolor{TopTwo} 0.9837 & \cellcolor{TopOne} 0.9946 & \cellcolor{TopTwo} 0.9837 & 0.9185 & 0.9674 & \cellcolor{TopOne} 0.9946 & 0.9293 & 0.9293 & 0.9457 & 0.9511 \\
          &       & LLM-Judge & \cellcolor{TopTwo} 0.9783 & \cellcolor{TopTwo} 0.9783 & 0.9348 & 0.9674 & \cellcolor{TopTwo} 0.9783 & 0.9674 & 0.9457 & 0.9565 & \cellcolor{TopOne} 0.9837 & 0.9266 & 0.9239 & 0.9348 & 0.9429 \\    
    \midrule
    \multicolumn{2}{c}{\multirow{4}[2]{*}{Overall}} & F1    & 0.3625 & 0.2966 & 0.6157 & 0.4118 & 0.3619 & 0.5282 & 0.6018 & 0.6132 & 0.3354 & \cellcolor{TopOne} 0.6884 & 0.6602 & \cellcolor{TopTwo} 0.6785 & 0.6629 \\
    \multicolumn{2}{c}{} & BLEU-1 & 0.3279 & 0.2651 & 0.5618 & 0.3683 & 0.3279 & 0.4792 & 0.5409 & 0.5572 & 0.2999 & \cellcolor{TopOne} 0.6357 & 0.6069 & \cellcolor{TopTwo} 0.6286 & 0.6117 \\
    \multicolumn{2}{c}{} & EM    & 0.2519 & 0.2005 & 0.4243 & 0.2741 & 0.2507 & 0.3402 & 0.4085 & 0.4296 & 0.2279 & \cellcolor{TopTwo} 0.4874 & 0.4635 & \cellcolor{TopOne} 0.4985 & 0.4787 \\
    \multicolumn{2}{c}{} & LLM-Judge & 0.4331 & 0.3644 & 0.7554 & 0.5018 & 0.4307 & 0.7306 & 0.7390 & 0.7700 & 0.4129 & \cellcolor{TopOne} 0.8320 & \cellcolor{TopTwo} 0.8068 & 0.7992 & 0.7846 \\
    \bottomrule
    \end{tabular}%
    }
  \label{tab:k_20}%
\end{table*}%

\begin{table*}[t!]
  \centering
  \caption{Retrieval evaluation results on Qwen-2.5-VL-7B with different retrieval number $K$. Since answer refusal (AR) questions intentionally introduce incorrect information and contain no supporting evidence in the conversation, evaluation metrics for this category are not supported.}
  \resizebox{0.95\linewidth}{!}{%
    \begin{tabular}{cl|cccc|cccc|cccc}
    \toprule
    \multicolumn{2}{c|}{\multirow{2}[2]{*}{\qwenlogo\ Qwen-2.5-VL-7B}} & \multicolumn{4}{c|}{Recall@$K$}   & \multicolumn{4}{c|}{Precision@$K$} & \multicolumn{4}{c}{Hit@$K$} \\
    \multicolumn{2}{c|}{} & 5     & 10    & 15    & 20    & 5     & 10    & 15    & 20    & 5     & 10    & 15    & 20 \\
    \midrule
    \multirow{7}[2]{*}{Overall} & FIFO  & 0.0116 & 0.0381 & 0.0685 & 0.0950 & 0.0043 & 0.0087 & 0.0123 & 0.0129 & 0.0196 & 0.0491 & 0.0792 & 0.1081 \\
          & Gen. Agent & 0.1707 & 0.2385 & 0.2854 & 0.3153 & 0.0955 & 0.0684 & 0.0544 & 0.0453 & 0.2718 & 0.3491 & 0.3962 & 0.4257 \\
          & NaiveRAG & 0.5381 & 0.6723 & 0.7420 & 0.7877 & 0.2694 & 0.1757 & 0.1315 & 0.1047 & 0.7236 & 0.8179 & 0.8605 & 0.8900 \\
          & MuRAG & 0.7506 & 0.8601 & 0.8990 & 0.9228 & 0.3686 & 0.2220 & 0.1572 & 0.1220 & 0.8861 & 0.9384 & 0.9548 & 0.9666 \\
          & UniversalRAG & 0.7311 & 0.8411 & 0.8781 & 0.8998 & 0.3691 & 0.2206 & 0.1555 & 0.1204 & 0.8697 & 0.9227 & 0.9404 & 0.9496 \\
          & NGM   & 0.6192 & 0.7475 & 0.7892 & 0.8065 & 0.3564 & 0.3457 & 0.3450 & 0.3424 & 0.7708 & 0.8612 & 0.8874 & 0.9037 \\
          & AUGUSTUS & 0.6729 & 0.7529 & 0.7785 & 0.7860 & 0.3341 & 0.2488 & 0.2213 & 0.2124 & 0.8553 & 0.8893 & 0.8959 & 0.8978 \\
    \midrule
    \multirow{7}[2]{*}{FR} & FIFO  & 0.0006 & 0.0354 & 0.0514 & 0.0793 & 0.0009 & 0.0059 & 0.0064 & 0.0075 & 0.0046 & 0.0365 & 0.0548 & 0.0822 \\
          & Gen. Agent & 0.1027 & 0.1363 & 0.1708 & 0.2062 & 0.0320 & 0.0237 & 0.0225 & 0.0215 & 0.1370 & 0.1872 & 0.2420 & 0.2922 \\
          & NaiveRAG & 0.4802 & 0.6297 & 0.7379 & 0.7811 & 0.2082 & 0.1511 & 0.1239 & 0.1014 & 0.6849 & 0.8265 & 0.8813 & 0.9041 \\
          & MuRAG & 0.6248 & 0.7712 & 0.8228 & 0.8647 & 0.2630 & 0.1840 & 0.1397 & 0.1146 & 0.7991 & 0.9041 & 0.9178 & 0.9406 \\
          & UniversalRAG & 0.6387 & 0.7940 & 0.8328 & 0.8768 & 0.2676 & 0.1913 & 0.1431 & 0.1180 & 0.8265 & 0.9178 & 0.9269 & 0.9498 \\
          & NGM   & 0.5025 & 0.6557 & 0.6999 & 0.7102 & 0.2348 & 0.2311 & 0.2329 & 0.2289 & 0.6712 & 0.8037 & 0.8265 & 0.8447 \\
          & AUGUSTUS & 0.5766 & 0.6626 & 0.6771 & 0.6862 & 0.2428 & 0.2025 & 0.1779 & 0.1748 & 0.7717 & 0.8174 & 0.8219 & 0.8402 \\
    \midrule
    \multirow{7}[2]{*}{VS} & FIFO  & 0.0172 & 0.0485 & 0.0678 & 0.0842 & 0.0059 & 0.0082 & 0.0085 & 0.0074 & 0.0261 & 0.0556 & 0.0719 & 0.0882 \\
          & Gen. Agent & 0.2173 & 0.2801 & 0.3309 & 0.3505 & 0.0706 & 0.0467 & 0.0379 & 0.0312 & 0.2843 & 0.3660 & 0.4183 & 0.4444 \\
          & NaiveRAG & 0.6723 & 0.7979 & 0.8300 & 0.8724 & 0.2366 & 0.1474 & 0.1037 & 0.0815 & 0.7712 & 0.8693 & 0.8856 & 0.9216 \\
          & MuRAG & 0.9080 & 0.9534 & 0.9660 & 0.9774 & 0.3379 & 0.1827 & 0.1237 & 0.0935 & 0.9444 & 0.9673 & 0.9739 & 0.9869 \\
          & UniversalRAG & 0.8906 & 0.9438 & 0.9570 & 0.9651 & 0.3301 & 0.1801 & 0.1218 & 0.0918 & 0.9444 & 0.9739 & 0.9837 & 0.9902 \\
          & NGM   & 0.7889 & 0.8799 & 0.8982 & 0.9217 & 0.3445 & 0.3204 & 0.3134 & 0.3058 & 0.8562 & 0.9183 & 0.9346 & 0.9542 \\
          & AUGUSTUS & 0.8363 & 0.8872 & 0.9056 & 0.9081 & 0.3124 & 0.2284 & 0.2062 & 0.1960 & 0.9248 & 0.9510 & 0.9542 & 0.9510 \\
    \midrule
        \multirow{7}[2]{*}{TTL} & FIFO  & 0.0026 & 0.0263 & 0.0938 & 0.1358 & 0.0042 & 0.0128 & 0.0265 & 0.0288 & 0.0208 & 0.0504 & 0.1246 & 0.1721 \\
          & Gen. Agent & 0.3152 & 0.4425 & 0.5036 & 0.5488 & 0.2570 & 0.1831 & 0.1375 & 0.1119 & 0.5460 & 0.6528 & 0.7003 & 0.7240 \\
          & NaiveRAG & 0.6837 & 0.8412 & 0.9017 & 0.9234 & 0.5501 & 0.3436 & 0.2457 & 0.1884 & 0.9496 & 0.9703 & 0.9881 & 0.9911 \\
          & MuRAG & 0.8523 & 0.9566 & 0.9760 & 0.9785 & 0.6754 & 0.3932 & 0.2671 & 0.2007 & 0.9822 & 0.9852 & 0.9941 & 0.9970 \\
          & UniversalRAG & 0.8617 & 0.9583 & 0.9682 & 0.9787 & 0.6967 & 0.3970 & 0.2669 & 0.2018 & 0.9763 & 0.9822 & 0.9881 & 0.9941 \\
          & NGM   & 0.7617 & 0.8919 & 0.9194 & 0.9203 & 0.7029 & 0.6919 & 0.6934 & 0.6930 & 0.9407 & 0.9763 & 0.9822 & 0.9822 \\
          & AUGUSTUS & 0.7231 & 0.8508 & 0.8903 & 0.9088 & 0.5569 & 0.3690 & 0.3094 & 0.2914 & 0.9733 & 0.9822 & 0.9881 & 0.9911 \\
    \midrule
        \multirow{7}[2]{*}{TR} & FIFO  & 0.0081 & 0.0122 & 0.0325 & 0.0569 & 0.0033 & 0.0024 & 0.0033 & 0.0037 & 0.0081 & 0.0163 & 0.0325 & 0.0569 \\
          & Gen. Agent & 0.0447 & 0.1043 & 0.1301 & 0.1558 & 0.0130 & 0.0154 & 0.0130 & 0.0122 & 0.0650 & 0.1463 & 0.1707 & 0.2033 \\
          & NaiveRAG & 0.3828 & 0.5000 & 0.5413 & 0.6009 & 0.1301 & 0.0870 & 0.0640 & 0.0524 & 0.5610 & 0.6585 & 0.6829 & 0.7317 \\
          & MuRAG & 0.6640 & 0.7724 & 0.8266 & 0.8557 & 0.2163 & 0.1285 & 0.0927 & 0.0732 & 0.8455 & 0.8943 & 0.9187 & 0.9187 \\
          & UniversalRAG & 0.6938 & 0.8089 & 0.8652 & 0.8733 & 0.2293 & 0.1341 & 0.0970 & 0.0736 & 0.8618 & 0.9106 & 0.9350 & 0.9431 \\
          & NGM   & 0.4898 & 0.5928 & 0.6179 & 0.6321 & 0.1846 & 0.1688 & 0.1626 & 0.1609 & 0.6504 & 0.7561 & 0.7805 & 0.8049 \\
          & AUGUSTUS & 0.6497 & 0.7243 & 0.7595 & 0.7683 & 0.2280 & 0.1743 & 0.1601 & 0.1509 & 0.8455 & 0.8780 & 0.8780 & 0.8780 \\
    \midrule
    \multirow{7}[2]{*}{VR} & FIFO  & 0.0000 & 0.0235 & 0.0551 & 0.0893 & 0.0000 & 0.0075 & 0.0126 & 0.0155 & 0.0000 & 0.0287 & 0.0632 & 0.1092 \\
          & Gen. Agent & 0.1439 & 0.1888 & 0.2521 & 0.2817 & 0.0885 & 0.0621 & 0.0544 & 0.0448 & 0.2874 & 0.3391 & 0.4138 & 0.4310 \\
          & NaiveRAG & 0.3093 & 0.4774 & 0.6010 & 0.6632 & 0.1540 & 0.1264 & 0.1115 & 0.0948 & 0.5057 & 0.6782 & 0.7759 & 0.8333 \\
          & MuRAG & 0.6158 & 0.7774 & 0.8551 & 0.8916 & 0.3299 & 0.2236 & 0.1701 & 0.1353 & 0.8276 & 0.9080 & 0.9483 & 0.9598 \\
          & UniversalRAG & 0.5497 & 0.6978 & 0.7560 & 0.7979 & 0.3356 & 0.2149 & 0.1586 & 0.1259 & 0.7759 & 0.8333 & 0.8621 & 0.8851 \\
          & NGM   & 0.4803 & 0.6351 & 0.7183 & 0.7541 & 0.2603 & 0.2707 & 0.2761 & 0.2762 & 0.6839 & 0.7989 & 0.8506 & 0.8793 \\
          & AUGUSTUS & 0.5110 & 0.5722 & 0.5796 & 0.5924 & 0.3430 & 0.3139 & 0.3032 & 0.3045 & 0.7644 & 0.8103 & 0.8103 & 0.8218 \\
    \midrule
    \multirow{7}[2]{*}{MR} & FIFO  & 0.0178 & 0.0591 & 0.0736 & 0.0914 & 0.0058 & 0.0117 & 0.0097 & 0.0085 & 0.0243 & 0.0728 & 0.0777 & 0.0971 \\
          & Gen. Agent & 0.1219 & 0.1752 & 0.2122 & 0.2376 & 0.0505 & 0.0359 & 0.0298 & 0.0252 & 0.2039 & 0.2621 & 0.2913 & 0.3204 \\
          & NaiveRAG & 0.5659 & 0.6600 & 0.7306 & 0.7793 & 0.2010 & 0.1257 & 0.0958 & 0.0784 & 0.7379 & 0.8252 & 0.8738 & 0.8981 \\
          & MuRAG & 0.7626 & 0.8527 & 0.9034 & 0.9237 & 0.2641 & 0.1597 & 0.1175 & 0.0925 & 0.9175 & 0.9612 & 0.9806 & 0.9806 \\
          & UniversalRAG & 0.7506 & 0.8590 & 0.9108 & 0.9364 & 0.2612 & 0.1631 & 0.1204 & 0.0949 & 0.8981 & 0.9563 & 0.9806 & 0.9854 \\
          & NGM   & 0.6023 & 0.7464 & 0.7927 & 0.8133 & 0.2551 & 0.2415 & 0.2401 & 0.2394 & 0.7573 & 0.8835 & 0.9223 & 0.9369 \\
          & AUGUSTUS & 0.7162 & 0.7748 & 0.8048 & 0.8014 & 0.2596 & 0.1964 & 0.1783 & 0.1684 & 0.8786 & 0.9175 & 0.9320 & 0.9223 \\
    \midrule
    \multirow{7}[2]{*}{KR} & FIFO  & 0.0412 & 0.0607 & 0.0885 & 0.1091 & 0.0099 & 0.0086 & 0.0099 & 0.0099 & 0.0494 & 0.0864 & 0.1111 & 0.1235 \\
          & Gen. Agent & 0.0782 & 0.1091 & 0.1214 & 0.1389 & 0.0247 & 0.0185 & 0.0148 & 0.0136 & 0.1111 & 0.1728 & 0.1852 & 0.2346 \\
          & NaiveRAG & 0.3899 & 0.5300 & 0.5808 & 0.6529 & 0.1605 & 0.1111 & 0.0823 & 0.0685 & 0.5679 & 0.7160 & 0.7531 & 0.8025 \\
          & MuRAG & 0.5752 & 0.7495 & 0.8015 & 0.8424 & 0.2272 & 0.1506 & 0.1119 & 0.0895 & 0.7531 & 0.8889 & 0.9012 & 0.9259 \\
          & UniversalRAG & 0.4750 & 0.6453 & 0.7433 & 0.7841 & 0.1852 & 0.1259 & 0.1029 & 0.0827 & 0.6790 & 0.8272 & 0.8889 & 0.8889 \\
          & NGM   & 0.4141 & 0.5599 & 0.6626 & 0.6863 & 0.1934 & 0.1841 & 0.1914 & 0.1907 & 0.6049 & 0.7037 & 0.7778 & 0.8025 \\
          & AUGUSTUS & 0.4483 & 0.5523 & 0.5860 & 0.5970 & 0.1895 & 0.1480 & 0.1324 & 0.1287 & 0.6420 & 0.7160 & 0.7284 & 0.7284 \\
    \midrule
    \multirow{7}[2]{*}{CD} & FIFO  & 0.0432 & 0.0494 & 0.0617 & 0.0741 & 0.0099 & 0.0062 & 0.0049 & 0.0043 & 0.0494 & 0.0494 & 0.0617 & 0.0741 \\
          & Gen. Agent & 0.0432 & 0.1091 & 0.1728 & 0.1934 & 0.0148 & 0.0198 & 0.0230 & 0.0204 & 0.0617 & 0.1852 & 0.2469 & 0.2593 \\
          & NaiveRAG & 0.3868 & 0.4650 & 0.5539 & 0.6284 & 0.1333 & 0.0827 & 0.0667 & 0.0574 & 0.5432 & 0.5926 & 0.7037 & 0.7407 \\
          & MuRAG & 0.6383 & 0.7868 & 0.8218 & 0.8881 & 0.2148 & 0.1296 & 0.0897 & 0.0759 & 0.7407 & 0.8519 & 0.8765 & 0.9259 \\
          & UniversalRAG & 0.4877 & 0.6008 & 0.6605 & 0.6687 & 0.1704 & 0.1037 & 0.0765 & 0.0586 & 0.5926 & 0.7160 & 0.7407 & 0.7284 \\
          & NGM   & 0.4428 & 0.5621 & 0.6074 & 0.6383 & 0.1774 & 0.1673 & 0.1642 & 0.1596 & 0.5802 & 0.7160 & 0.7407 & 0.7654 \\
          & AUGUSTUS & 0.6049 & 0.6584 & 0.6893 & 0.6770 & 0.2123 & 0.1578 & 0.1446 & 0.1385 & 0.6914 & 0.7531 & 0.7778 & 0.7654 \\
    \bottomrule
    \end{tabular}%
    }
  \label{tab:full_retrieval}%
\end{table*}%

\subsection{Case Study}\label{sec:case_study}
\begin{figure*}[t!]
\centering
\includegraphics[width=\linewidth]{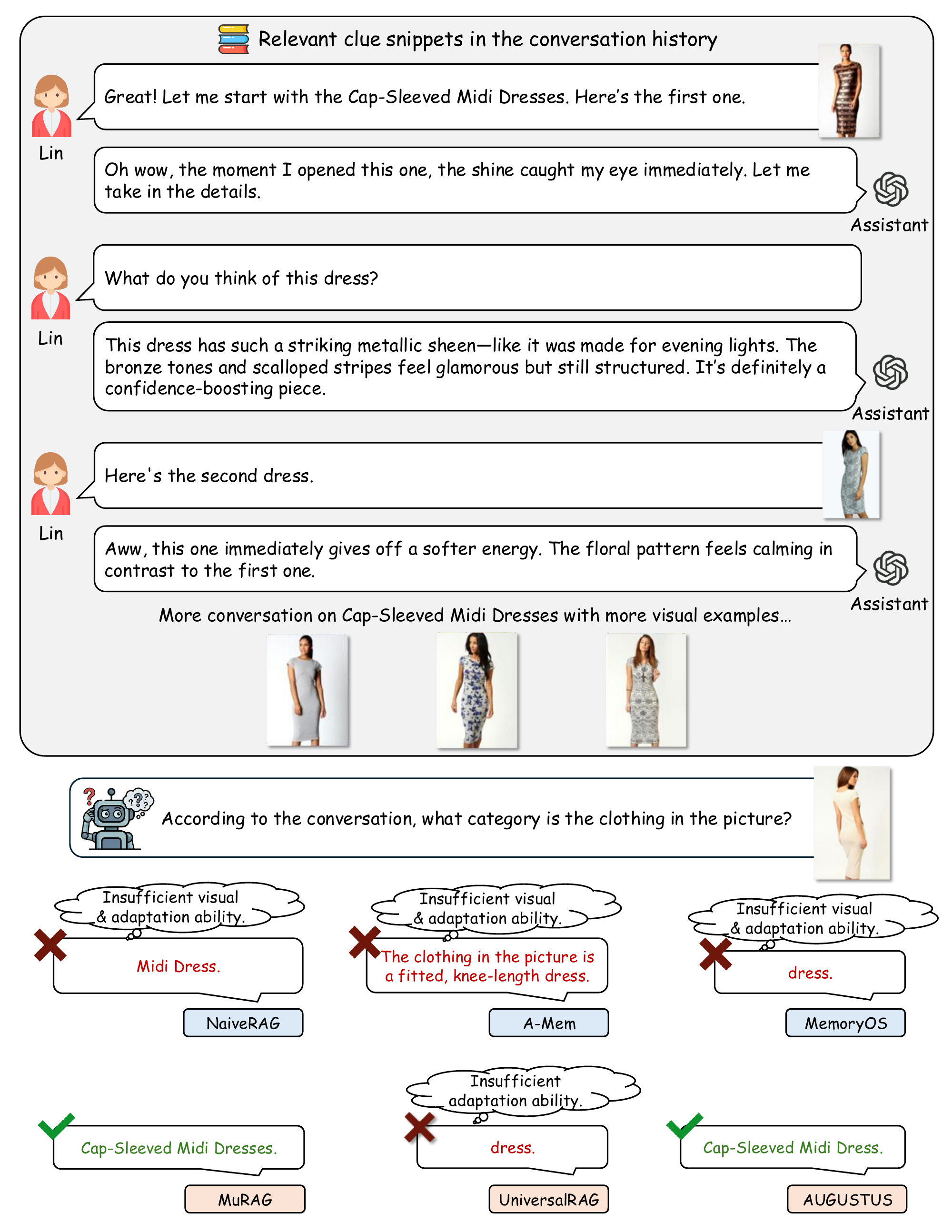}
\caption{Case study with the test-time learning example.}
\label{fig:ttl_case}
\end{figure*}

\begin{figure*}[t!]
\centering
\includegraphics[width=\linewidth]{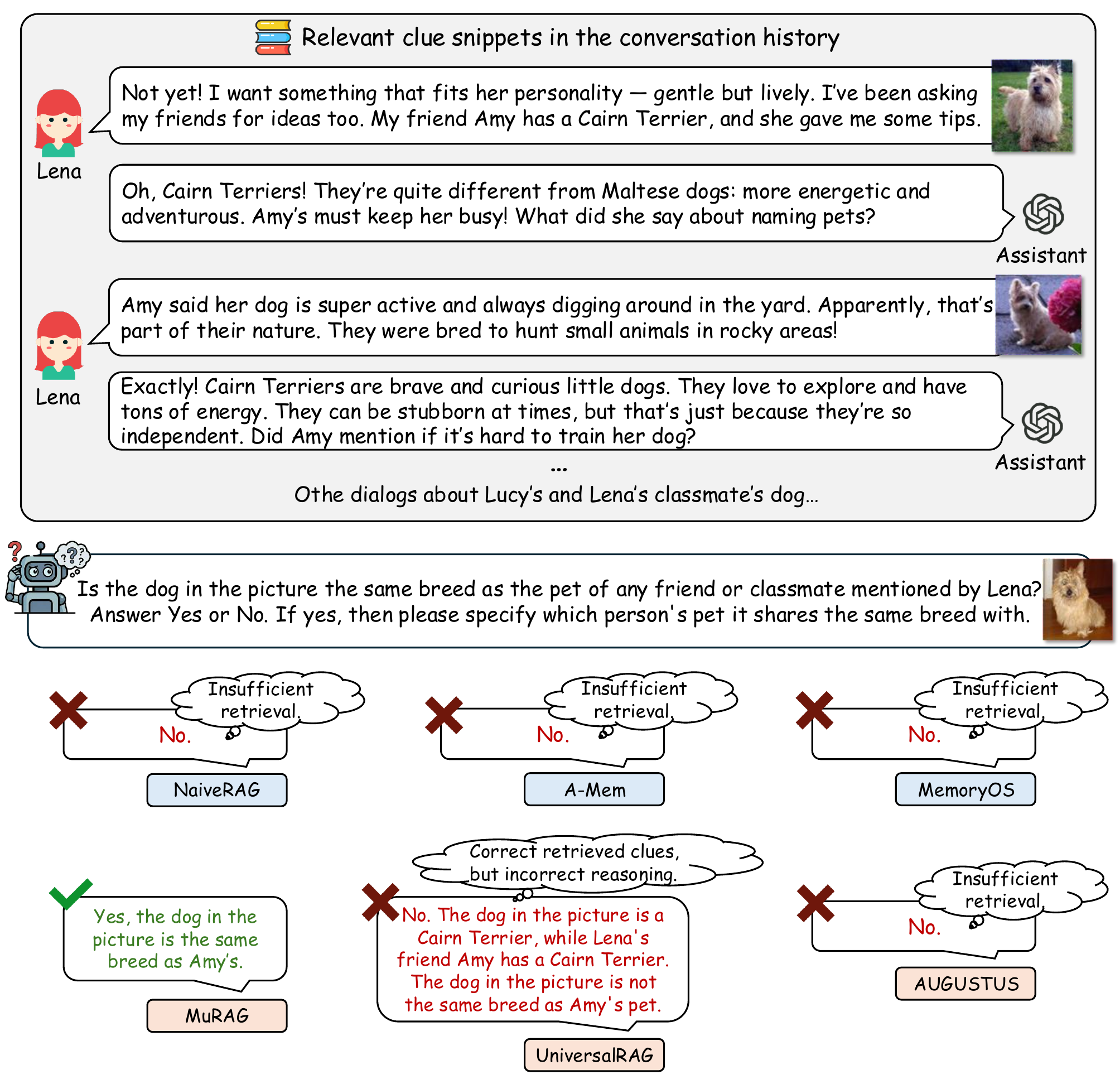}
\caption{Case study with the visual-centric reasoning example.}
\label{fig:vr_case}
\end{figure*}

\begin{figure*}[t!]
\centering
\includegraphics[width=\linewidth]{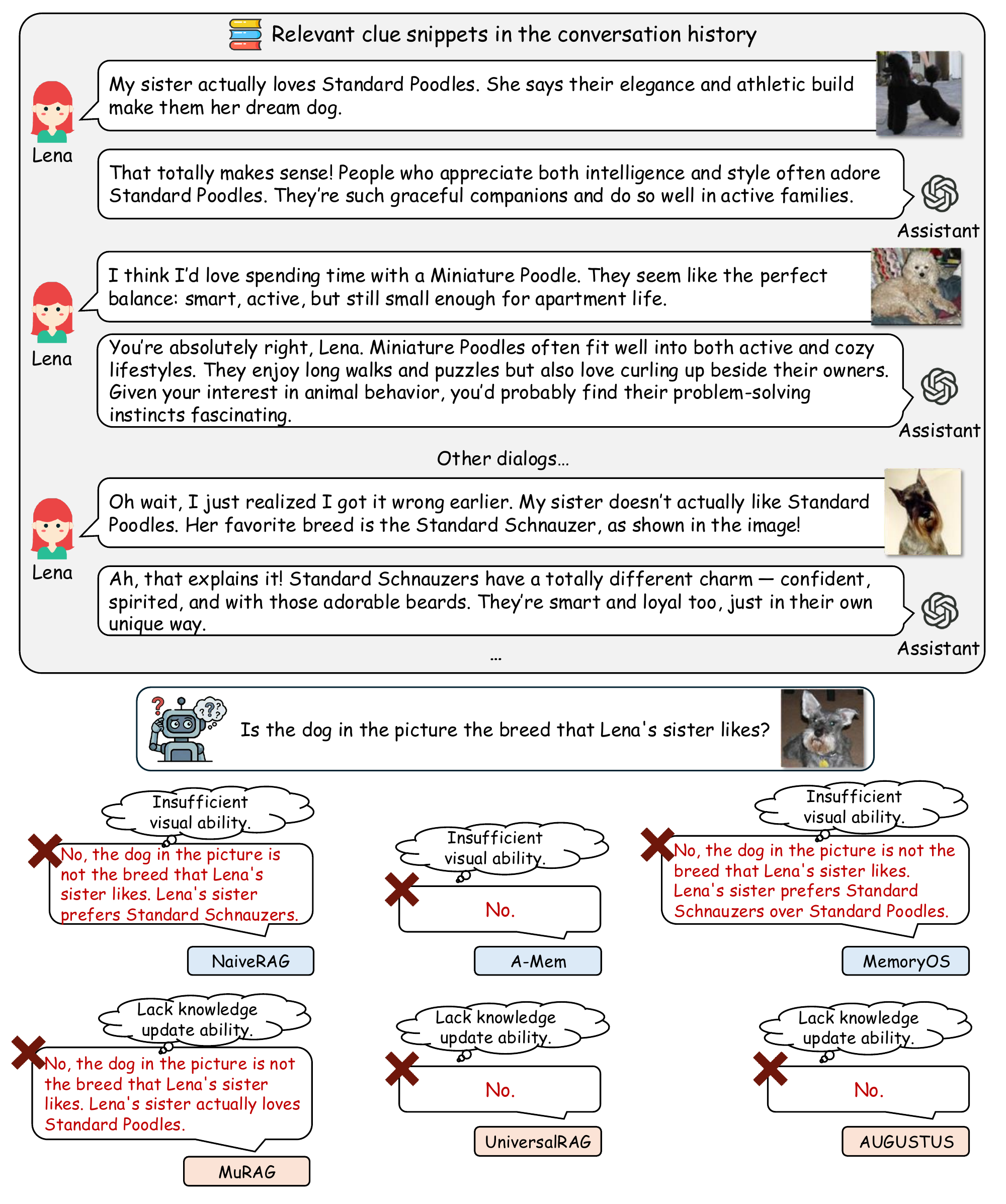}
\caption{Case study with the knowledge resolution example.}
\label{fig:kr_case}
\end{figure*}

We further present a qualitative case study to illustrate how representative memory systems behave under different multimodal long-term memory challenges in \name.
We select three textual memory models (NaiveRAG, A-Mem, MemoryOS) and three multimodal memory models (MuRAG, UniversalRAG, AUGUSTUS) and analyze their responses on three representative subtasks. These cases correspond to test-time learning of the memory extraction \& adaptation task, visual-centric reasoning of the memory reasoning task, and knowledge resolution of the memory knowledge management task, as shown in Figure~\ref{fig:ttl_case}, Figure~\ref{fig:vr_case}, and Figure~\ref{fig:kr_case}, respectively. From the case studies, we can have the following observations.

Figure~\ref{fig:ttl_case} presents a test-time learning scenario in which the agent must adapt to newly introduced visual exemplars during the conversation and generalize the learned concept at inference time. The example requires identifying the clothing category based on a sequence of visual demonstrations and prior conversational cues.
Although textual memory baselines are augmented with high-quality visual captions, textual memory methods still exhibit clear limitations. 
Although NaiveRAG and A-Mem retrieve relevant textual clues, they lack the ability to integrate newly observed visual patterns, resulting in generic or incorrect predictions. MemoryOS exhibits a similar limitation, indicating that strong textual organization alone is insufficient when the task requires visual abstraction at inference time.
In contrast, multimodal memory methods that explicitly preserve visual representations demonstrate stronger adaptability. MuRAG and AUGUSTUS successfully align the new image with previously observed visual examples and correctly infer the category. This case highlights the limitation of caption-based textual memory. While high-quality captions can convey semantic details, they remain a lossy and static representation of visual information, making it difficult for models to perform visual pattern induction and concept generalization at test time. Effective test-time learning therefore requires memory systems that can retain and compare visual evidence directly, rather than relying solely on textualized visual descriptions.

Figure~\ref{fig:vr_case} illustrates a visual-centric reasoning task, where the model must determine whether the dog in a given image matches the breed of a pet mentioned earlier in the conversation. This case reveals a more subtle failure mode. Several models retrieve the correct conversational evidence, such as the mention that Amy owns a Cairn Terrier, yet still produce incorrect final answers. In particular, UniversalRAG demonstrates correct retrieval but flawed reasoning, incorrectly concluding that the dog in the image is not a Cairn Terrier despite visual similarity. This indicates that retrieval success does not guarantee correct multimodal reasoning. Textual memory models, on the other hand, often fail earlier due to missing or incomplete visual grounding. Only MuRAG correctly integrates both the retrieved textual clue and the visual evidence to reach the correct conclusion. This case exposes a critical gap in existing memory systems, namely that multimodal reasoning requires not only retrieving relevant information but also jointly reasoning over visual and textual cues in a coherent manner.

Figure~\ref{fig:kr_case} focuses on knowledge resolution under evolving conversational states. In this example, the user explicitly corrects earlier information about a preferred dog breed. The model must update its memory and discard outdated knowledge. Most evaluated methods struggle with this scenario. This example reveals two distinct failure modes. First, some memory systems fail to update their knowledge, continuing to rely on outdated beliefs despite the explicit correction in the dialogue. This behavior is particularly evident in some multimodal memory methods, indicating that preserving multimodal information alone does not guarantee effective knowledge revision. Second, even when some textual memory models successfully incorporate the updated preference at the textual level, they still fail to produce the correct answer due to insufficient visual grounding. Lacking the ability to accurately recognize or verify the dog breed shown in the image, these models are unable to align the updated textual knowledge with the visual evidence, leading to incorrect final predictions. This case demonstrates that effective knowledge resolution requires both explicit mechanisms for memory revision and robust visual understanding to ground updated knowledge in perceptual evidence, which remains largely underexplored in current multimodal designs.

\subsection{Ethics Statement}

\subsubsection{Data Privacy}
To ensure data privacy, the benchmark does not include any personally identifiable information from real user conversations. All conversations were either generated by LLMs or reconstructed from open-sourced paper repositories, and were subsequently verified and edited under strong human supervision by the author team. The majority of images and portions of single-session dialogue data were collected from open-source repositories. For newly collected images sourced from the web, only materials released under permissive or Creative Commons licenses were used. All conversations and question–answer pairs were reviewed by annotators to remove inaccurate, misleading, or inappropriate content and to avoid the inclusion of sensitive material. The benchmark construction process does not involve real user interactions or private user data. Thus, the dataset does not involve data privacy violations and is intended solely for research and benchmarking purposes.

\subsubsection{Annotation} 
All annotation and verification work in this benchmark was conducted exclusively for data quality control and validation purposes. The annotation tasks were limited to checking the correctness, consistency, and completeness of conversational data and model-generated outputs, including verifying dialogue coherence across sessions, validating multimodal grounding, and reviewing question–answer pairs for clarity and correctness. Annotators were treated solely as an interface for data verification rather than as research subjects, and no analysis of annotator behavior, agreement patterns, or individual decision strategies was performed.
No personally identifiable information or demographic attributes were collected or recorded during the annotation process. The annotation workflow did not involve tracking individual annotators, comparing annotator performance, or modeling human judgment patterns. All annotation results were used only to improve data quality and benchmark reliability, with the research focus remaining only on the properties of the data, models, and evaluation framework rather than on any human behavior.
All annotators were fully informed that the annotated data would be released publicly for research purposes. Since all annotation and verification work was conducted by the author team as part of the research process, no external recruitment or monetary compensation was involved. The benchmark study does not involve any privacy data of real users or human-subject research.

\subsubsection{Scientific Artifacts}
All data sources and models underlying the benchmark are used with explicit references and official links. Detailed information can be found in Appendix~\ref{sec:data_source},~\ref{sec:model_detail_appendix}, and~\ref{sec:bench_eval_details}. 
The benchmark primarily targets English-language multimodal conversations across diverse daily-life and domain-specific scenarios, as illustrated in Appendix~\ref{sec:data_detail_appendix}. No demographic attributes of real individuals are represented or analyzed. The benchmark and dataset will be released publicly with clear documentation, permitting use for research purposes.

\subsection{Potential Risks}
This benchmark is designed for research and evaluation purposes only. While it does not involve real user data or personal information, potential risks may arise from unintended misuse, such as over-interpreting benchmark results as indicators of real-world deployment readiness. In addition, models evaluated on this benchmark may inherit biases present in underlying pretrained language model backbones~\cite{schramowski2022large}. We mitigate these risks by clearly scoping the benchmark to controlled memory research settings and by providing transparent documentation of data sources, evaluation protocols, and limitations.

\subsection{Use of AI Assistants} 
LLMs are used in this work strictly as auxiliary tools for data construction, an evaluation metric, and limited language polishing. During dataset construction, LLMs assist in drafting candidate multi-session conversational texts under predefined structural constraints and scenario specifications. These drafts are subsequently reviewed, edited, and refined by annotators, who also insert appropriate images and ensure that multimodal dependencies are genuine, necessary, and non-trivial for downstream tasks.
LLMs are further used to generate part of the candidate question–answer pairs and to perform preliminary checks for coherence, clarity, and factual consistency. All QA pairs are then manually verified and revised by annotators. In addition, LLMs are used in a limited manner for language polishing of the manuscript, including improving fluency and presentation. 
In our evaluation, we also adopt LLM-as-a-Judge as one of the evaluation metrics, following common practice in prior work~\cite{li2025generation}.
All technical content, experimental analysis, and scientific claims are authored and finalized by the authors.